%% file: main.tex
\newif\ifisKDD

\isKDDfalse

\ifisKDD
\documentclass[sigconf]{acmart}
\usepackage[utf8]{inputenc}
\usepackage[T1]{fontenc}
\usepackage{amsfonts}
\usepackage{nicefrac}
\usepackage{xcolor}
\usepackage{tikz}
\usepackage{amsmath,multicol,bm,graphics}

\else
\documentclass{article} 
\usepackage{fullpage}

\fi

\usepackage{tcolorbox}
\usepackage{microtype}
\usepackage{graphicx}
\usepackage{subcaption}
\usepackage{booktabs}

\usepackage{hyperref}
\usepackage{url}

\ifisKDD

\AtBeginDocument{
  \providecommand\BibTeX{{
    \normalfont B\kern-0.5em{\scshape i\kern-0.25em b}\kern-0.8em\TeX}}}

\copyrightyear{2023}
\acmYear{2023}
\setcopyright{rightsretained}
\acmConference[KDD '23]{Proceedings of the 29th ACM SIGKDD Conference on Knowledge Discovery and Data Mining}{August 6--10, 2023}{Long Beach, CA, USA}
\acmBooktitle{Proceedings of the 29th ACM SIGKDD Conference on Knowledge Discovery and Data Mining (KDD '23), August 6--10, 2023, Long Beach, CA, USA}
\acmDOI{10.1145/3580305.3599518}
\acmISBN{979-8-4007-0103-0/23/08}

\fi

\input{sections/Macro}

\begin{document}

\ifisKDD
\else

\title{Evaluating natural language processing models with generalization metrics that do not need access to any training or testing data}
\date{}

\author{%
  Yaoqing Yang$^{1}$, 
  Ryan Theisen$^{2}$, 
  Liam Hodgkinson$^{3}$, 
  Joseph E. Gonzalez$^{2}$,  \\
  Kannan Ramchandran$^{2}$, 
  Charles H. Martin$^{4}$, 
  Michael W. Mahoney$^{2,5,6}$ \\
  $^1$ Dartmouth College\\
  $^2$ University of California, Berkeley\\
  $^3$ University of Melbourne\\
  $^4$ Calculation Consulting\\
  $^5$ International Computer Science Institute\\
  $^6$ Lawrence Berkeley National Laboratory\\
}

\maketitle

\fi

\input{sections/Abstract}

\input{sections/Introduction_KDD}
\input{sections/Preliminaries}
\input{sections/HTtheory}
\input{sections/Experiments}
\input{sections/Experiments_Huggingface}
\input{sections/Conclusion}

\vspace{5mm}
\noindent\textbf{Acknowledgements.}
MM would like to acknowledge the IARPA (contract W911NF20C0035), NSF, and ONR for providing partial support of this work.
KR would like to acknowledge support from NSF CIF-1937357, NSF CIF-2007669 and ARO fund 051242-002.
JG would like to acknowledge supports from NSF CISE Expeditions Award CCF-1730628, NSF CAREER Award and gifts from Alibaba Group, Amazon Web Services, Ant Group, Ericsson, Facebook, Futurewei, Google, Intel, Microsoft, Nvidia, Scotiabank, Splunk and VMware.
WeightWatcher is a publicly-available tool distributed under Apache License 2.0 with copyright held by Calculation Consulting.
Our conclusions do not necessarily reflect the position or the policy of our sponsors, and no official endorsement should be~inferred.

\bibliographystyle{ACM-Reference-Format}
\bibliography{bib}

\ifisKDD
\else
\appendix
\input{sections/Metrics}
\input{sections/Additional_setup}
\input{sections/Margin_issue}
\input{sections/Experiments_ODR}

\input{sections/Experiments_Additional}
\fi

\end{document}

%% file: sections/Macro.tex
\usepackage[utf8]{inputenc}
\usepackage{graphicx}

\ifisKDD
\else
\fi

\usepackage{enumitem}
\usepackage{amsmath}
\usepackage{amsfonts}
\usepackage{dcolumn}
\usepackage{tabu}
\usepackage{array}
\usepackage{xspace}
\usepackage{makecell}
\usepackage{amsthm}

\theoremstyle{plain}
\newtheorem{theorem}{Theorem}[section]

\theoremstyle{definition}

\theoremstyle{remark}
\newtheorem{remark}[theorem]{Remark}

\usepackage{colortbl}
\usepackage{booktabs, multirow} 
\usepackage{soul}
\usepackage{changepage,threeparttable}

\PassOptionsToPackage{compress,sort}{natbib}
\usepackage{natbib}

\newcommand{\xb}{\mathbf{x}}

\newcommand{\Ib}{\mathbf{I}}

\newcommand{\Ub}{\mathbf{U}}
\newcommand{\Wb}{\mathbf{W}}
\newcommand{\Xb}{\mathbf{X}}

\newcommand{\vect}{\text{vect}}
\newcommand{\sumid}{\sum_{i=1}^d}
\newcommand{\aveid}{\frac{1}{d}\sum_{i=1}^d}

\newcommand\gb{ \rowcolor{gray!15}}

\newcommand{\LTWODIST}{\texttt{l2\_dist}\xspace}
\newcommand{\LTWO}{\texttt{l2}\xspace}
\newcommand{\PARAMNORM}{\texttt{param\_norm}\xspace}
\newcommand{\FRODIST}{\texttt{fro\_dist}\xspace}
\newcommand{\LOGNORM}{\texttt{log\_norm}\xspace}
\newcommand{\LOGSUMOFFRO}{\texttt{log\_sum\_of\_fro}\xspace}
\newcommand{\LOGSPECTRALNORM}{\texttt{log\_spectral\_norm}\xspace}
\newcommand{\DISTSPECINIT}{\texttt{dist\_spec\_int}\xspace}
\newcommand{\LOGPRODOFFRO}{\texttt{log\_prod\_of\_fro}\xspace}
\newcommand{\LOGSUMOFSPEC}{\texttt{log\_sum\_of\_spec}\xspace}
\newcommand{\LOGPRODOFSPEC}{\texttt{log\_prod\_of\_spec}\xspace}
\newcommand{\PATHNORM}{\texttt{path\_norm}\xspace}
\newcommand{\ALPHA}{\texttt{PL\_alpha}\xspace}

\newcommand{\MPSOFTRANK}{\texttt{mp\_softrank}\xspace}
\newcommand{\STABLERANK}{\texttt{stable\_rank}\xspace}
\newcommand{\EXPONENT}{\texttt{E\_TPL\_lambda}\xspace}
\newcommand{\ETPLBETA}{\texttt{E\_TPL\_beta}\xspace}
\newcommand{\EXPDISTEXPONENT}{\texttt{EXP\_lambda}\xspace}

\newcommand{\KSDISTANCE}{\texttt{ks\_distance}\xspace}
\newcommand{\PLKSDISTANCE}{\texttt{PL\_ks\_distance}\xspace}
\newcommand{\ETPLKSDISTANCE}{\texttt{E\_TPL\_ks\_distance}\xspace}

\newcommand{\ALPHAWEIGHTED}{\texttt{alpha\_weighted}\xspace}
\newcommand{\LOGALPHANORM}{\texttt{log\_alpha\_norm}\xspace}
\newcommand{\INVERSEMARGIN}{\texttt{inverse\_margin}\xspace}
\newcommand{\LOGPRODOFSPECOVERMARGIN}{\texttt{log\_prod\_of\_spec\_over\_margin}\xspace}
\newcommand{\LOGSUMOFSPECOVERMARGIN}{\texttt{log\_sum\_of\_spec\_over\_margin}\xspace}
\newcommand{\LOGPRODOFFROOVERMARGIN}{\texttt{log\_prod\_of\_fro\_over\_margin}\xspace}
\newcommand{\LOGSUMOFFROOVERMARGIN}{\texttt{log\_sum\_of\_fro\_over\_margin}\xspace}
\newcommand{\PATHNORMOVERMARGIN}{\texttt{path\_norm\_over\_margin}\xspace}
\newcommand{\PACBAYESINIT}{\texttt{pacbayes\_init}\xspace}
\newcommand{\PACBAYESORIG}{\texttt{pacbayes\_orig}\xspace}
\newcommand{\PACBAYESFLATNESS}{\texttt{pacbayes\_flatness}\xspace}
\newcommand{\PACBAYESMAGINIT}{\texttt{pacbayes\_mag\_init}\xspace}
\newcommand{\PACBAYESMAGORIG}{\texttt{pacbayes\_mag\_orig}\xspace}
\newcommand{\PACBAYESMAGFLATNESS}{\texttt{pacbayes\_mag\_flatness}\xspace}

\newcommand{\numberofmetrics}{28\xspace}
\newcommand{\numberofmodels}{360\xspace}
\newcommand{\numberofmodelsWMT}{200\xspace}

%% file: sections/Abstract.tex
\begin{abstract}
Selecting suitable architecture parameters and training hyperparameters is essential for enhancing machine learning (ML) model performance.
Several recent empirical studies conduct large-scale correlational analysis on neural networks (NNs) to search for effective \emph{generalization metrics} that can guide this type of model selection. Effective metrics are typically expected to correlate strongly with test performance.
In this paper, we expand on prior analyses by examining generalization-metric-based model selection with the following objectives:
(i) focusing on natural language processing (NLP) tasks, as prior work primarily concentrates on computer vision (CV) tasks; 
(ii) considering metrics that directly predict \emph{test error} instead of the \emph{generalization gap};
(iii) exploring metrics that do not need access to data to compute.
From these objectives, we are able to provide the first model selection results on large pretrained Transformers from Huggingface using generalization metrics. 
Our analyses consider (I) hundreds of Transformers trained in different settings, in which we systematically vary the amount of data, the model size and the optimization hyperparameters, (II) a total of 51 pretrained Transformers from eight families of Huggingface NLP models, including GPT2, BERT, etc., and (III) a total of \numberofmetrics existing and novel generalization metrics.
Despite their niche status, we find that metrics derived from the heavy-tail (HT) perspective are particularly useful in NLP tasks, exhibiting stronger correlations than other, more popular metrics. To further examine these metrics, we extend prior formulations relying on power law (PL) spectral distributions to exponential (EXP) and exponentially-truncated power law (E-TPL) families.%
\ifisKDD
\else
\footnote{This is the technical report version of a paper that appeared in conference version as ``Test accuracy vs. generalization gap: model selection in NLP without accessing training or testing data''~\citep{yang2023test}; the title is different due to the conference submission policy, and there is an additional detailed empirical analysis that covers more datasets and evaluation methods.}
\fi
\end{abstract}

%% file: sections/Introduction_KDD.tex
\section{Introduction}

Selecting the optimal hyperparameters, such as those for training or model size, is a critical phase in the ML pipeline.
Motivated by the importance of model selection, recent years have seen a wide array of large-scale empirical studies on the various metrics used to predict the test-time performance of ML models \citep{jiang2019fantastic,dziugaite2020search,martin2020predicting_NatComm,MM21a_simpsons_TR}.
These \emph{generalization metrics} have been applied in a wide variety of data science tasks, including predicting the quality of pretrained learning models \citep{martin2020predicting_NatComm,martin2019traditional}, designing effective training procedures \citep{foret2020sharpness,izmailov2018averaging}, improving network efficiency \citep{chen2021neural,dong2019hawq}, quantifying model robustness \citep{yang2020boundary,tanay2016boundary}, improving ensemble learning techniques \citep{garipov2018loss,fort2019deep}, analyzing and improving large-scale machine learning contests \citep{MM21a_simpsons_TR}, and so on.
They are typically studied using correlational analysis, measuring how strongly each metric correlates with (and therefore, can predict) model performance. 
In this regard, several recent works point out the deficiencies of existing generalization metrics, including a lack of ``robustness'' to the changes of environmental hyperparameters \citep{jiang2019fantastic,dziugaite2020search} (such as data, neural network architecture and training schemes), or the \emph{Simpson's paradox} that generalization metrics perform differently (i.e., predict opposite trends) when applied to each sub-part of a collection of learning models or to the holistic study \citep{MM21a_simpsons_TR}.
Another drawback is the over-reliance on CV models, which are relatively well-explored, and are not always representative of other types of tasks.
With few exceptions \citep{nakkiran2019deep,martin2020predicting_NatComm,yang2021taxonomizing}, systematic studies in other fields, such as NLP, are largely missing.

\noindent{\bf Generalization metrics for model selection in NLP.}
The objective of this work is to provide a systematic study of generalization metrics in NLP, addressing several deficiencies in prior studies \citep{jiang2019fantastic,dziugaite2020search,martin2020predicting_NatComm}.
Compared to CV, model selection in NLP has several important differences that require careful consideration.
For example, the training data from standard CV benchmarks can often be easily obtained, while large language model datasets are typically web-scale and are challenging to access.
Therefore, generalization metrics that can assess the quality of learning models \emph{without access to data} are ideal for NLP.
In this paper, we focus on generalization metrics that do \emph{not} need access to data, which is useful for evaluating pretrained NLP models \citep{wolf-etal-2020-transformers}.
Indeed, recent work has demonstrated that access to training or testing data should not be necessary for assessing the model quality of learning models \citep{martin2020predicting_NatComm}, though these findings have yet to be evaluated at scale in the NLP domain.
Furthermore, it is typically infeasible to train NLP models to interpolate the (frequently large) training set. Contrary to common practice for CV models, the training error on NLP datasets is often much larger than zero.
This becomes an issue when applying most existing generalization metrics as they compare models through the \emph{generalization gap} (i.e., the difference between training and test performance) rather than the test error itself.
Metrics that focus on ranking the generalization gap include most of the well-known metrics in CV, such as those based on the PAC-Bayesian framework \citep{neyshabur2017pac,mcallester1999pac} and margins \citep{bartlett2017spectrally,pitas2017pac,jiang2018predicting}.

To illustrate the issue, consider selecting between two models with test errors $e_1,e_2$, training errors $l_1,l_2$, and generalization gaps $g_1 = e_1 - l_1$ and $g_2 = e_2 - l_2$. Assuming a generalization metric can \emph{rank} the generalization gap perfectly (which is often the focus of prior studies on generalization metrics~\citep{jiang2019fantastic,jiang2020neurips,dziugaite2020search}) \footnote{As the report of the NeurIPS 2020 Competition on Predicting Generalization in Deep Learning \citep{jiang2020neurips} points out, the generalization metric ``should'' be able to order models' performance in a way similar to the generalization gap, and thus one hopes that it can be used for model selections or neural architecture search. However, see~\citet{MM21a_simpsons_TR} for a detailed exposition of issues and problems with this.}, we know only that one model has a larger training-test gap than another ($g_1 > g_2$).
For these two models, even if we have access to both models' exact training errors $l_1,l_2$, we still cannot determine which model exhibits smaller test error: if $l_1 < l_2$, we cannot determine whether $l_1 + g_1 > l_2 + g_2$ unless we know the training-test gaps $g_1,g_2$ \emph{explicitly}.
Therefore, if our objective is to construct a metric that correctly predicts model performance, rank correlation with the generalization gap is insufficient.
In this paper, we aim to study how generalization metrics rank correlate with model quality, for which we use test error as a close approximation.
As we will demonstrate (in Figure~\ref{fig:test_acc_vs_generalization_gap}), rank correlation with the generalization gap indeed does not imply rank correlation with model quality in practice, and in fact often orders models in the opposite order of their test errors.
From a practical point of view, for NLP tasks, we prefer generalization metrics that can directly predict trends in test error (or similar evaluation metrics in NLP, such as the test BLEU score~\citep{papineni2002bleu}) rather than trends in the generalization gap.

Naturally, we cannot expect a metric to be universally correlated with test error if evaluating the metric does not need data. 
However, within certain classes of models (e.g., stages of training in one model or across pre-trained models), they may be effective at diagnosing model quality. 
With these objectives in mind, among the generalization metrics in the literature, we take particular interest in those derived from the heavy-tail self regularization (HT-SR) theory \citep{martin2019traditional,martin2018implicit_JMLRversion} due to reasons summarized in the following:
\begin{center}
\begin{tcolorbox}
We choose HT-SR generalization metrics for model selection in NLP because they (i) predict test error directly instead of the generalization gap and (ii) do not require access to training (or testing) data.
\end{tcolorbox}
\end{center}

\ifisKDD
\else
In addition to these two advantages, actual data often follow heavy-tail distributions \citep{feldman2020does,martin2018implicit_JMLRversion,martin2020predicting_NatComm}, which can be even more evident in NLP than the more well-behaved datasets in CV \citep{li2017deep} that are often used to study~generalization. 
\fi

\noindent{\bf HT-SR theory and shape metrics.}
The core principle of HT-SR theory is that HT structures arise naturally in the ESDs of the weight matrices \footnote{The ESD of a weight matrix $\Wb$ refers to the empirical density of the eigenvalues of the squared weight matrix $\Wb^\top \Wb$. See ``Preliminary of ESDs of weight matrices'' at the end of the Introduction.} as the result of extracting various correlations in data during optimization~\citep{martin2018implicit_JMLRversion,martin2019traditional,martin2020heavy,martin2020predicting_NatComm,MM21a_simpsons_TR}.
Its primary practical consequence is that by
estimating the PL coefficient from the ESDs (requiring only weights), one can predict model quality, as smaller coefficients are reported to correspond to higher test accuracy. However, these estimators can be unstable, and so one must be careful not to rely on them alone. The quality of the PL fit itself should also point to similar conclusions~\citep{martin2018implicit_JMLRversion}, which can be a sanity check.

\begin{figure*}[ht]
    \centering
    \begin{subfigure}{0.3\textwidth}
        \includegraphics[width=\textwidth]{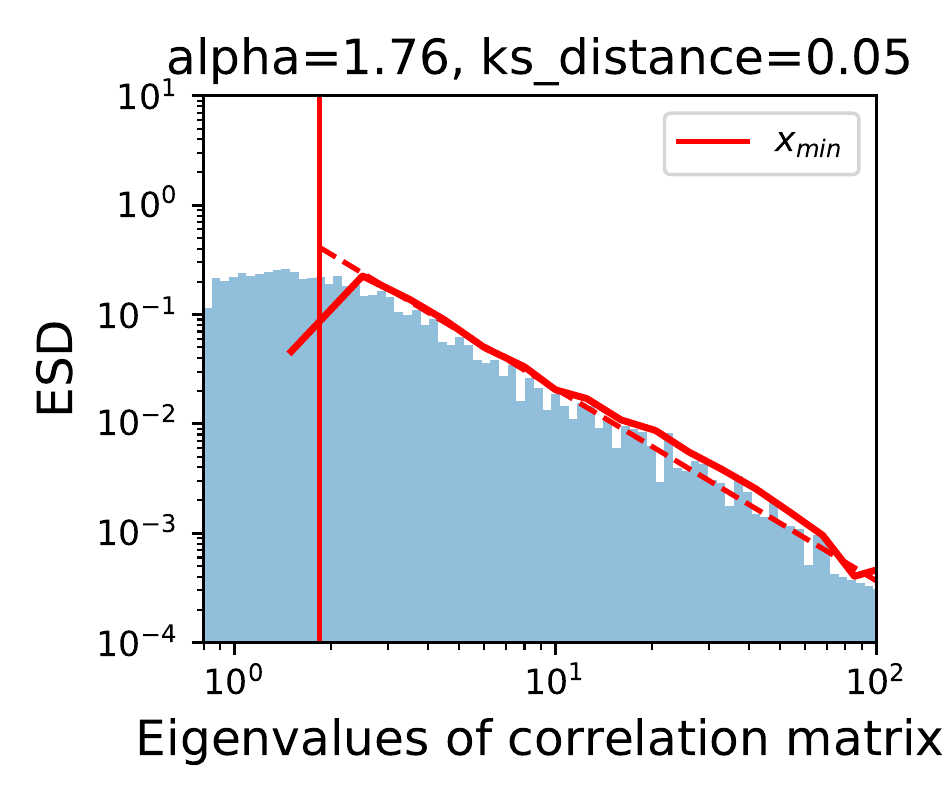}
        \caption{Small \KSDISTANCE.\vspace{3mm}}
        \label{fig:PL_good}
    \end{subfigure}
    \begin{subfigure}{0.3\textwidth}
        \includegraphics[width=\textwidth]{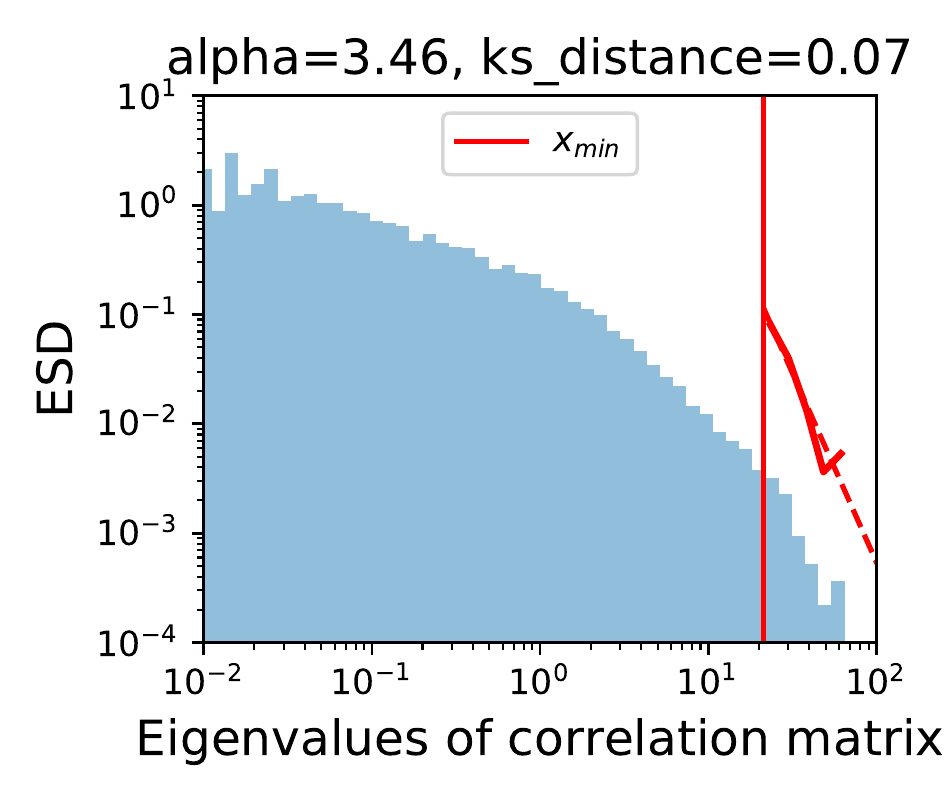}
        \caption{Mediocre \KSDISTANCE.\vspace{3mm}}
        \label{fig:PL_mediocre}
    \end{subfigure}
    \begin{subfigure}{0.3\textwidth}
        \includegraphics[width=\textwidth]{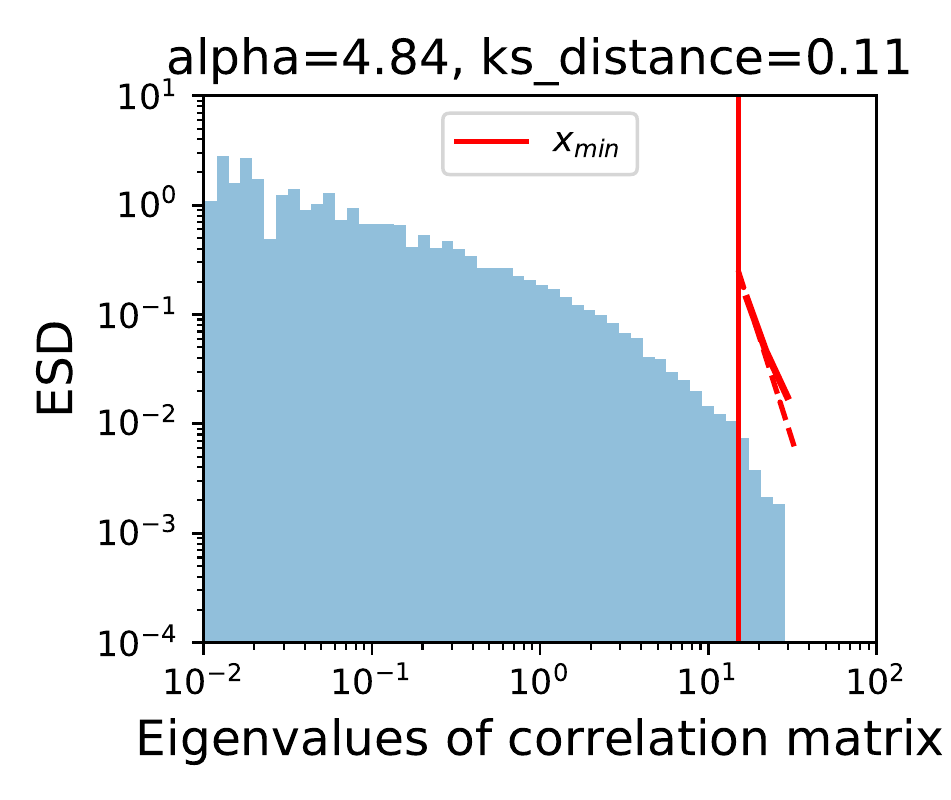}
        \caption{Large \KSDISTANCE.\vspace{3mm}}
        \label{fig:PL_bad}
    \end{subfigure}
    
    \begin{subfigure}{0.3\textwidth}
        \includegraphics[width=\textwidth]{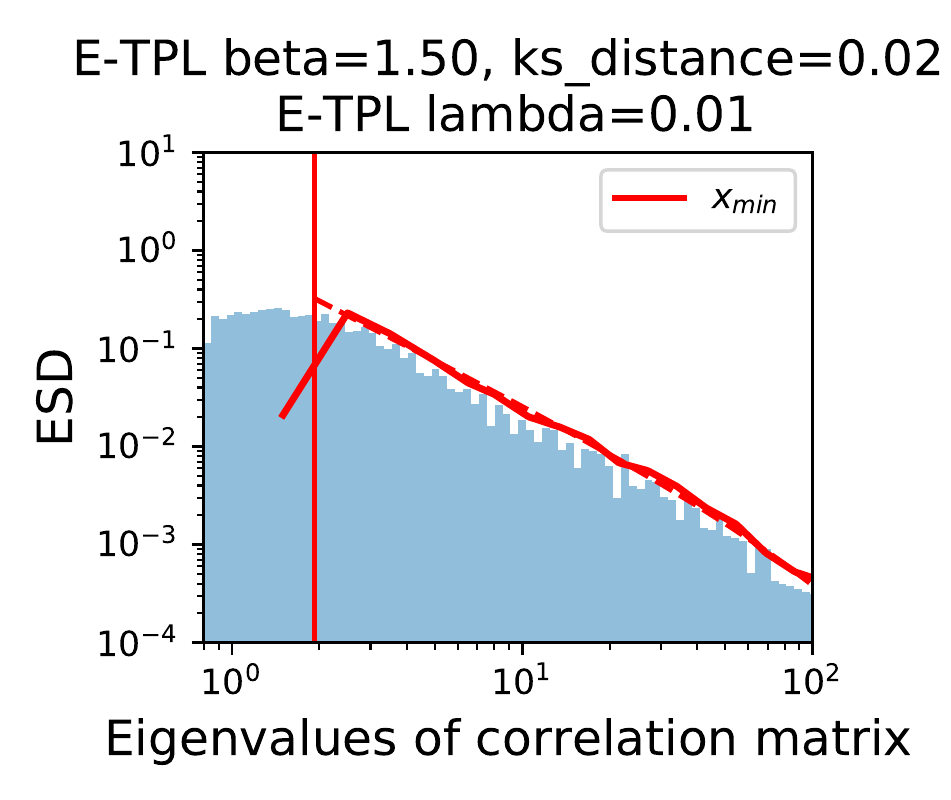}
        \caption{E-TPL fitting of the ESD.}
        \label{fig:TPL_good}
    \end{subfigure}
    \begin{subfigure}{0.3\textwidth}
        \includegraphics[width=\textwidth]{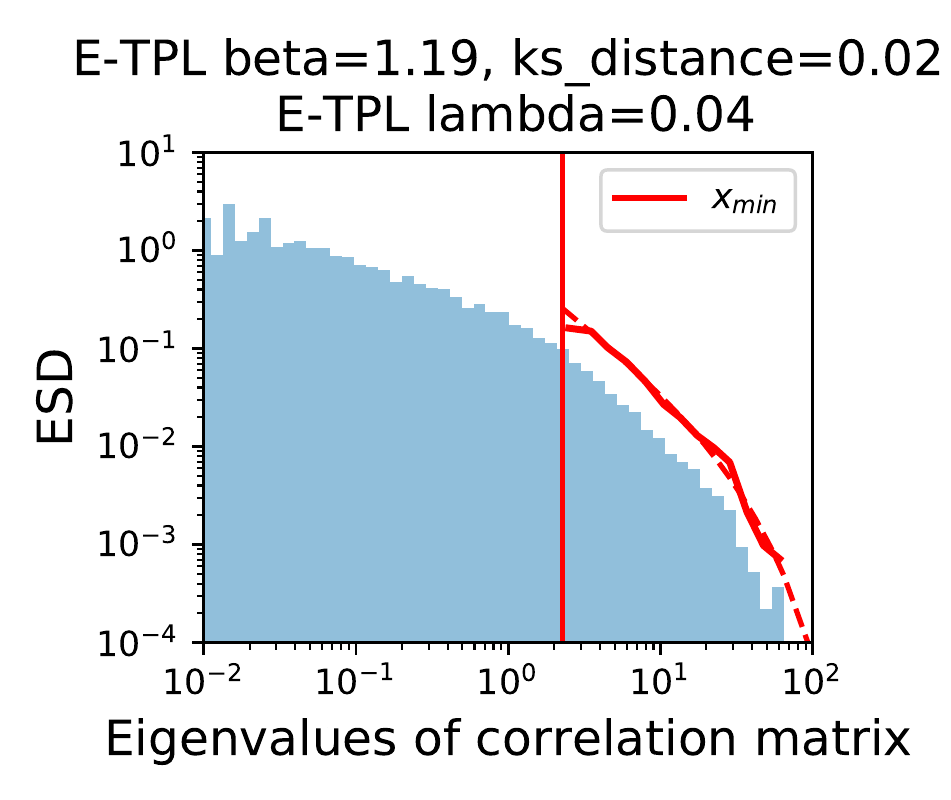}
        \caption{E-TPL fitting of the ESD.}
        \label{fig:TPL_mediocre}
    \end{subfigure}
    \begin{subfigure}{0.3\textwidth}
        \includegraphics[width=\textwidth]{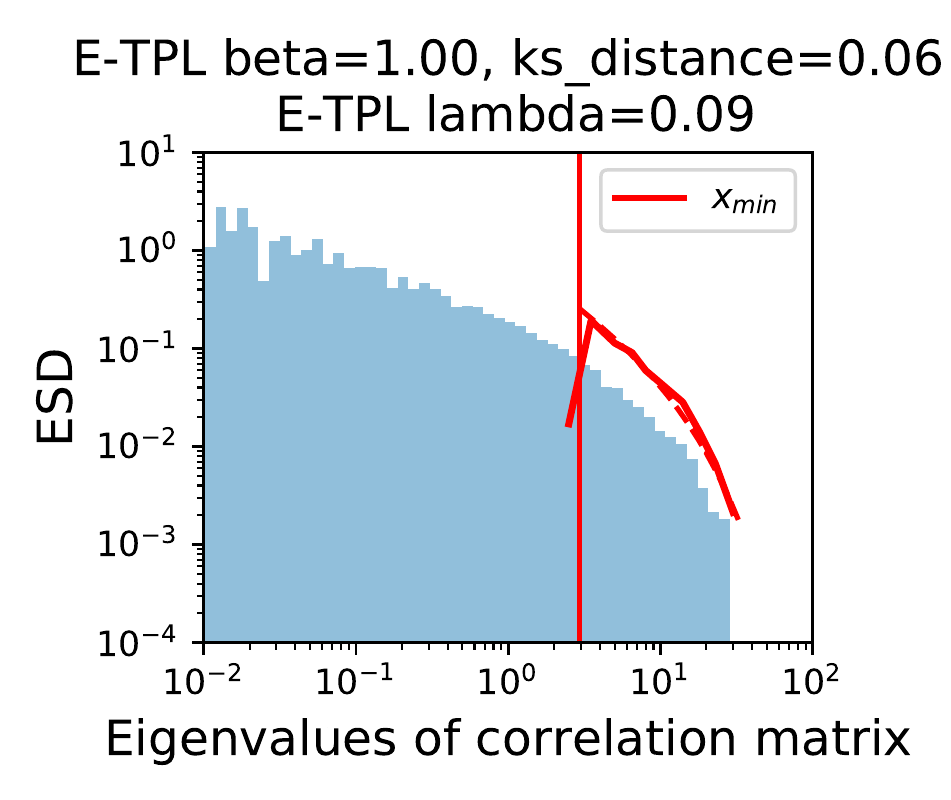}
        \caption{E-TPL fitting of the ESD.}
        \label{fig:TPL_bad}
    \end{subfigure}
    \caption{Comparing PL and E-TPL fitting. {\bf(First row).} Good, mediocre, and bad PL fittings measured by the \KSDISTANCE. {\bf(Second row).} E-TPL fitting of the ESD on the same column. Blue histograms represent the ESDs. Solid vertical lines represent the lower threshold $x_{\min}$ of the PL distribution found by the fitting procedure. Solid curves represent ESDs truncated using $x_{\min}$, and dashed curves represent the fitted HT~distributions. 
    }
    \label{fig:PL_vs_TPL}
\end{figure*}

The principles of HT-SR theory extend beyond fitting the PL coefficient, however, as ESDs can take many forms.
To this end, we study three different types of distributions to fit to the ESDs of weight matrices, including power laws (PL) in Eqn. \eqref{eqn:ALPHA}, exponentially truncated power laws (E-TPL) in Eqn. \eqref{eqn:EXPONENT}, and exponential laws (EXP) in Eqn. \eqref{eqn:EXP_DIST_EXPONENT}.
These are all commonly considered families of distributions in classical studies of PL \citep{clauset2009power}, and it is often hard in practice to predict which family fits data the
\ifisKDD
\else
best (as we show in this paper, this is true for deep NNs especially). \fi
Figure~\ref{fig:PL_vs_TPL} shows examples of comparing different HT fittings on the same ESD. Following \citet{MM21a_simpsons_TR}, we refer to the various metrics derived from HT-SR as \emph{shape metrics}.

\noindent\textbf{Contributions.} The following summarizes our main contributions.
\begin{itemize}[noitemsep,topsep=0pt,leftmargin=*,after=,before=]
    \item Deviating from prior work examining generalization metrics in CV \citep{jiang2019fantastic,dziugaite2020search},  we provide the first systematic correlational analysis on various generalization metrics in NLP. 
    Our detailed studies include:
    \begin{itemize}
        \item considering \numberofmodels transformers trained on WMT14~\citep{bojar2014findings} with varying hyperparameters, and eight families of pretrained SOTA transformers downloaded from Huggingface~\citep{wolf-etal-2020-transformers}, including BERT \citep{devlin2018bert}, GPT2 \citep{radford2019language}, ALBERT (both v1 and v2) \citep{lan2019albert}, etc;
        \item providing the first systematic study of applying generalization metrics to the model selection of Transformers without any training/validation/testing data;
        \item measuring the correlation between \numberofmetrics generalization metrics and the model quality (measured by test-time performance) over three different model classes: (i) models trained with the optimal hyperparameters, (ii) a single model at different stages of training, and (iii) a model trained with different hyperparameters (similar to \citet{jiang2019fantastic,MM21a_simpsons_TR}.)
        \ifisKDD\else
        \item corroborating results on multiple other datasets, including Wikitext-103, Reddit and MNLI; results using various ways of measuring rank correlations, including Spearman's rank correlation and Kendall's tau.\fi
    \end{itemize} 
    \item We revisit prior findings on data-dependent metrics motivated by margins and PAC-Bayesian bounds \citep{jiang2019fantastic,dziugaite2020search}, finding that while these metrics perform well in predicting the \emph{generalization gap}, none of them satisfactorily predicts test error directly.
    \item When applied appropriately, we find that HT-based shape metrics consistently perform better than scale metrics (or norm-based metrics) for predicting model quality.
    \item We extend prior studies on HT-SR theory and investigate alternative models to fit heavy-tail/light-tail distributions. Our results show that E-TPL fits are comparatively robust alternatives to PL fits on suboptimally-trained models. 
\end{itemize}
\ifisKDD
\else
This technical report is an extended version of the conference paper~\citep{yang2023test}. Compared to the conference version, this report provides full details of the correlational analysis when a single hyperparameter is changed (see Section~\ref{sec:Simpson}). It also provides corroborating results on multiple other datasets and evaluation methods.
\fi
In order that our results can be reproduced and extended, we have open-sourced our code.\footnote{\url{https://github.com/nsfzyzz/Generalization\_metrics\_for\_NLP}}

%% file: sections/Preliminaries.tex
\paragraph{Preliminary of ESDs of weight matrices.}
Consider a NN with $d$ layers and corresponding weight matrices $\Wb_1$, $\Wb_2$,..., $\Wb_d$.
For each weight matrix $\Wb_i$ with shape $N\times M$, assume without loss of generality that $N \geq M$ (otherwise, consider $\Wb_i^\top$). 
We define the correlation matrix as $\Xb_i=\Wb_i^\top \Wb_i$, and denote the eigenvalues of $\Xb_i$ as $\{\lambda_j\}_{j=1}^M$, so that $\lambda_j=\sigma_j^2$, where $\{\sigma_j\}_{j=1}^M$ are the singular values of $\Wb_i$.
Furthermore, we use $\lambda_{i,\max}$ to denote the maximum eigenvalue of the correlation matrix $\Xb_i$.
The ESD (empirical spectral density) of the weight matrix $\Wb_i$ refers to the empirical density of the eigenvalues of $\Xb_i$, typically represented through a histogram.
We let $p(x)$ denote the density function to fit the ESD taking values in the interval $(x_{\min}, x_{\max})$. For a power law, $p$ satisfies
\begin{equation}\label{eqn:ALPHA}
    p(x) \propto x^{-\alpha}, \quad x_{\min}<x<x_{\max}.
\end{equation}
From \citet{MM21a_simpsons_TR}, $x_{\max}$ is chosen to be the maximum eigenvalue of the empirical correlation matrix.
However, $x_{\min}$ is a variable to be optimized to improve the quality of PL fitting, and it is not equal to the minimum eigenvalue in general.

%% file: sections/HTtheory.tex
\section{Heavy-tail self-regularization}

Here, we provide a brief overview of the HT-SR theory, and discuss several metrics that can be derived from it. 
According to HT-SR theory, the ESDs of the weight matrices become more heavy-tailed during training as they become increasingly correlated.
One can quantify the extent of these correlations by fitting a PL to the ESD of a weight matrix, for example, by using the open-source \texttt{WeightWatcher} tool \footnote{\url{https://github.com/CalculatedContent/WeightWatcher}}\citep{martin2020predicting_NatComm}.
After computing the ESD of a weight matrix, we use the maximum likelihood estimate from~\citet{alstott2014powerlaw} to fit the PL distribution, the specific form of which has been defined in \eqref{eqn:ALPHA}. Let \ALPHA denote the PL coefficient averaged over layers; effectively the slope of the tail of the ESD of the pooled weights, on a log-log scale.

Correctly identifying and fitting PL distributions is well-known to be a challenge in practice.
For example, a density that appears as a straight line on a log-log scale plot need not follow a power law,
as there are many other distributions that could show a similar behavior, including lognormal and exponential-type distributions \citep{clauset2009power}.
Nested distributions such as E-TPL, which combine the pure PL and other distributional assumptions, can often improve the quality of fitting \citep{clauset2009power,alstott2014powerlaw}.
Therefore, in addition to PL (defined in~\eqref{eqn:ALPHA}), we consider several other distribution classes from the literature.
\begin{itemize}
    \item (\EXPONENT and \ETPLBETA) The ESDs are assumed to take a ``nested'' form in the interval $(x_{\min},x_{\max})$.
\begin{equation}\label{eqn:EXPONENT}
    p(x) \propto x^{-\beta}\exp(-\lambda x), \, x_{\min}<x<x_{\max}.
\end{equation}
    After fitting the E-TPL, we call the exponential truncation coefficient $\lambda$ the \EXPONENT metric, and we call the PL coefficient the \ETPLBETA metric.
    \item (\EXPDISTEXPONENT). The ESDs are assumed to take the following form, in the interval $(x_{\min},x_{\max})$.
    \begin{equation}\label{eqn:EXP_DIST_EXPONENT}
    p(x) \propto \exp(-\lambda x), \, x_{\min}<x<x_{\max}.
    \end{equation}
    After fitting the EXP, we call the exponential coefficient $\lambda$ the \EXPDISTEXPONENT metric.
\end{itemize}

For more details of the various metrics considered in this paper, see Table~\ref{tab:metrics}.
All of the metrics derived from HT-SR do \emph{not} require access to data,
\ifisKDD\else nor do they require model training/retraining, and thus \fi
they are relatively cheap to compute.
Our primary comparisons are between shape metrics (derived from HT-SR), and scale metrics (mostly norm-based). 
Scale metrics are mostly studied in prior work \citep{jiang2019fantastic,dziugaite2020search}, while shape metrics have received less attention.
\ifisKDD
\else
For the precise definitions of these metrics, see Appendix~\ref{sec:metrics}.
\fi

\input{sections/MetricTable}

\noindent{\bf Issues of PL fitting.}
It is well-known that subtle issues can arise when fitting the ESDs \citep{clauset2009power,alstott2014powerlaw,martin2017rethinking,MM21a_simpsons_TR}.
To best mitigate these issues in PL fits, we adopt the fitting strategies used in \texttt{WeightWatcher}~\citep{martin2017rethinking}.
For example, as in \citet{clauset2009power}, it is common to choose the lower threshold $x_{\min}$ which coincides with the best quality fit under the Kolmogorov--Smirnoff statistic
\ifisKDD
\else
(referred to as \PLKSDISTANCE for PL and \ETPLKSDISTANCE for E-TPL in the sequel; see Eqn.~\eqref{eqn:KS_DISTANCE}).
\fi
However, this method is time-consuming, especially for E-TPL as there are two parameters to fit. Instead, we adopt the \emph{fix-finger method} (see \texttt{WeightWatcher}) which selects $x_{\min}$ as the peak of the ESD when fitting E-TPLs. More than a simple speed improvement, we find this method also yields more stable results. 

\noindent{\bf Comparing PL and E-TPL fitting.}
Referring to Figure~\ref{fig:PL_vs_TPL}, we now discuss how E-TPL could partially address these fitting issues. On the first row of Figure~\ref{fig:PL_vs_TPL}, we show three typical cases of PL fitting.
In Figure~\ref{fig:PL_good}, the log-log scale reveals a ``linear region'' of the histogram, which the PL fitting correctly locates. The quality of fit, measured by the \KSDISTANCE, is within a typical range, as reported in Table 6 of~\citet{martin2018implicit_JMLRversion}.
In Figure~\ref{fig:PL_mediocre} and Figure~\ref{fig:PL_bad}, the ESDs do not exhibit a clear linear region on the log-log scale.
Following \citet{martin2018implicit_JMLRversion}, it is ill-advised to consider metrics derived from a PL fit in these scenarios. In practice, this typically occurs when \ALPHA~$>4$ (e.g., see Figure~\ref{fig:PL_bad}).
On the other hand, in these two cases, the corresponding E-TPL fits (shown on the second row in Figure~\ref{fig:PL_vs_TPL}) still closely match the empirical density function (see Figure~\ref{fig:TPL_mediocre} and Figure~\ref{fig:TPL_bad}),
and the \KSDISTANCE on the second row using a E-TPL fit is smaller than that for the PL fit on the first row, even when the fit on the second row clearly covers a larger part of the ESD.
\ifisKDD\else\footnote{We note that the value of \KSDISTANCE can be effectively made smaller if one restricts to a smaller part of the distribution, as is often done in practice by optimizing the $x_{\min}$ in the (truncated) PL distribution~\eqref{eqn:ALPHA}. This potential bias is alleviated by using the fix-finger method.
} \fi
In these two cases, the \EXPONENT plays a similar role as the \ALPHA in PL fitting, and provides an effective alternative when the ESD does not exhibit a proper PL.

%% file: sections/MetricTable.tex
\ifisKDD
\else

\begin{table*}[ht!]
\centering
{\scriptsize
\setlength\tabcolsep{2pt}
\begin{tabular}{ccccccccc}
\hline
\gb \multicolumn{1}{c}{Name} & \multicolumn{1}{c}{Eqn} & \multicolumn{1}{c}{Ref} & \multicolumn{1}{c}{\shortstack{Need initial\\ weights?}} & \multicolumn{1}{c}{\shortstack{Scale or \\shape}} & \multicolumn{1}{c}{\shortstack{Need \\data?}} & \multicolumn{1}{c}{\shortstack{Need \\gpu?}} & \multicolumn{1}{c}{\shortstack{Predicting model quality \\ or generalization gap?}} \\
\hline
\hline
\PARAMNORM &\eqref{eqn:PARAM_NORM} & \cite{jiang2019fantastic} & No & Scale & No & No & Generalization gap \\ 
\gb \FRODIST &\eqref{eqn:FRO_DIST}& \cite{jiang2019fantastic} & Yes & Scale  & No & No & Generalization gap \\ 
\LOGNORM & \eqref{eqn:LOG_NORM}  & \cite{martin2018implicit_JMLRversion} & No & Scale & No & No & Generalization gap \\ 
\gb\LOGSPECTRALNORM & \eqref{eqn:LOG_SPECTRAL_NORM} & \cite{MM21a_simpsons_TR} & No & Scale & No & No & Generalization gap \\ 
\DISTSPECINIT &\eqref{eqn:DIST_SPEC_INIT} & \cite{jiang2019fantastic} & Yes & Scale & No & No & Generalization gap \\
\gb \PATHNORM & \eqref{eqn:PATH_NORM} & \cite{neyshabur2015norm} & No & Scale & No & No & Generalization gap \\ 
\MPSOFTRANK & \eqref{eqn:MP_SOFTRANK} & \cite{martin2018implicit_JMLRversion} & No & Scale/Shape & No & No & Model quality \\
\gb \STABLERANK &  \eqref{eqn:STABLE_RANK} & \cite{martin2018implicit_JMLRversion} & No & Scale/Shape & No & No & Model quality \\ 
\ALPHA & \eqref{eqn:ALPHA}  & \cite{martin2018implicit_JMLRversion} & No & Shape & No & No & Model quality \\ 
\gb \ETPLBETA & \eqref{eqn:EXPONENT}  & \shortstack{This paper\\ WeightWatcher} & No & Shape & No & No & Model quality \\ 
\EXPONENT & \eqref{eqn:EXPONENT}  & \shortstack{This paper\\ WeightWatcher} & No & Shape & No & No & Model quality \\ 
\gb \EXPDISTEXPONENT & \eqref{eqn:EXP_DIST_EXPONENT}  & \shortstack{This paper\\ WeightWatcher}  & No & Shape & No & No & Model quality \\ 
\PLKSDISTANCE & \eqref{eqn:KS_DISTANCE} & \cite{martin2018implicit_JMLRversion} & No & Shape & No & No & Model quality \\ 
\gb \ETPLKSDISTANCE & \eqref{eqn:KS_DISTANCE} & \shortstack{This paper \\ \cite{martin2018implicit_JMLRversion}} & No & Shape & No & No & Model quality \\ 
\ALPHAWEIGHTED & \eqref{eqn:ALPHA_WEIGHTED} & \cite{martin2018implicit_JMLRversion} & No & Hybrid & No & No & Model quality \\ 
\gb \LOGALPHANORM & \eqref{eqn:LOG_ALPHA_NORM} & \cite{MM21a_simpsons_TR} & No & Hybrid & No & No & Model quality \\
\INVERSEMARGIN & \eqref{eqn:INVERSE_MARGIN} & \cite{jiang2019fantastic} & No & Scale & Yes & Maybe & Generalization gap\\
\gb \LOGPRODOFSPECOVERMARGIN & \eqref{eqn:LOG_PROD_OF_SPEC_OVER_MARGIN} & \shortstack{\citet{bartlett2017spectrally}\\\citet{pitas2017pac}}  & No & Scale & Yes & Maybe & Generalization gap \\
\LOGSUMOFSPECOVERMARGIN & \eqref{eqn:LOG_SUM_OF_SPEC_OVER_MARGIN} & \shortstack{\citet{bartlett2017spectrally}\\\citet{pitas2017pac}} & No & Scale & Yes & Maybe & Generalization gap \\
\gb \LOGPRODOFFROOVERMARGIN & \eqref{eqn:LOG_PROD_OF_FRO_OVER_MARGIN} & \shortstack{\citet{bartlett2017spectrally}\\\citet{pitas2017pac}} & No & Scale & Yes & Maybe & Generalization gap \\
\LOGSUMOFFROOVERMARGIN & \eqref{eqn:LOG_SUM_OF_FRO_OVER_MARGIN} & \shortstack{\citet{bartlett2017spectrally}\\\citet{pitas2017pac}} & No & Scale & Yes & Maybe & Generalization gap \\
\gb \PATHNORMOVERMARGIN & \eqref{eqn:PATH_NORM_OVER_MARGIN} & \cite{neyshabur2015norm} & No & Scale & Yes & Maybe & Generalization gap \\
\PACBAYESINIT & \eqref{eqn:PACBAYES_INIT} & \cite{neyshabur2017exploring} & Yes & Scale & Yes & Yes & Generalization gap \\
\gb \PACBAYESORIG & \eqref{eqn:PACBAYES_ORIG} & \cite{neyshabur2017exploring} & No & Scale & Yes & Yes & Generalization gap  \\
\PACBAYESFLATNESS & \eqref{eqn:PACBAYES_FLATNESS} & \cite{neyshabur2017exploring} & No & Scale & Yes & Yes  & Generalization gap \\
\gb \PACBAYESMAGINIT & \eqref{eqn:PACBAYES_MAG_INIT} & \cite{jiang2019fantastic} & Yes & Scale & Yes & Yes & Generalization gap  \\
\PACBAYESMAGORIG & \eqref{eqn:PACBAYES_MAG_ORIG} & \cite{jiang2019fantastic} & No & Scale & Yes & Yes & Generalization gap  \\
\gb \PACBAYESMAGFLATNESS & \eqref{eqn:PACBAYES_MAG_FLATNESS} & \cite{jiang2019fantastic} & No & Scale & Yes & Yes & Generalization gap  \\
\bottomrule
\end{tabular}
}
\caption{Overview of the generalization metrics considered in this paper. We focus on the \emph{shape} metrics derived from the ESDs of weight matrices.
Due to the space constraint, definitions of some metrics are presented in Appendix~\ref{sec:metrics}.
\vspace{-.0cm}
}
\label{tab:metrics}
\end{table*}

\fi

%% file: sections/Experiments.tex
\section{Empirical results}
\label{sec:experiments}

\begin{figure*}
    \centering
    \includegraphics[width=0.32\textwidth]{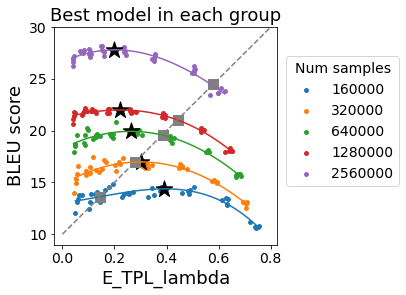}
    \includegraphics[width=0.32\textwidth]{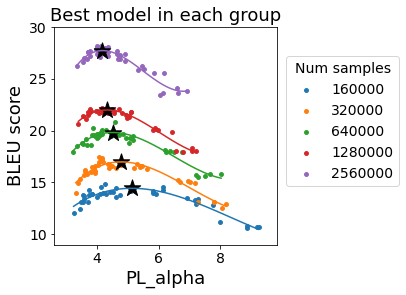}
    \includegraphics[width=0.32\textwidth]{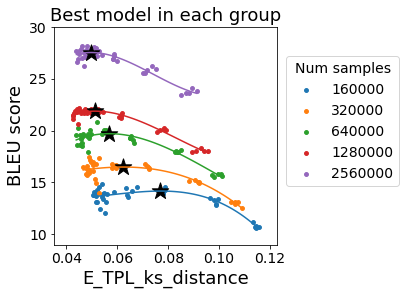}
     \includegraphics[width=0.32\textwidth]{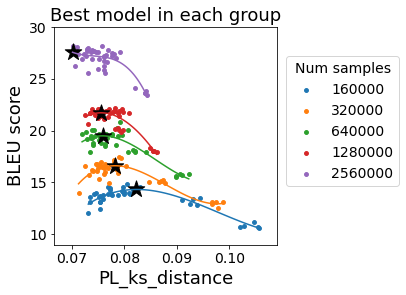}
    \includegraphics[width=0.32\textwidth]{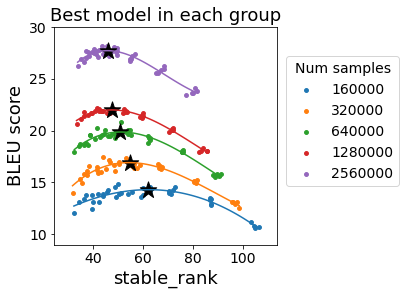}
    \includegraphics[width=0.32\textwidth]{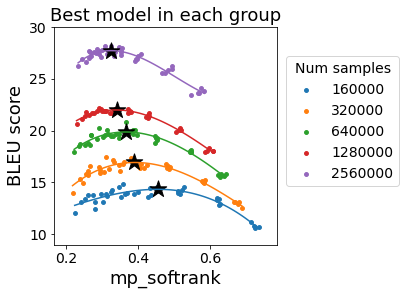}
    
    \caption{BLEU-score vs. six shape metrics for 200 Transformers trained on WMT14 with varying hyperparameters. HT-SR theory applies for optimally-tuned models (black stars), that is, for optimally-tuned models indicated by the black stars, models that have better BLEU scores exhibit heavier-tailed ESDs.
    \ifisKDD
\else
    HT-SR theory is a predictive theory designed to provide predictions for state-of-the-art-models~\citep{martin2018implicit_JMLRversion,martin2020predicting_NatComm}; thus, for \fi
    suboptimal models, the HT-SR metrics can be anti-correlated with model quality, see e.g. the grey dotted line in the first subfigure. 
    }
    \label{fig:Optimal_models_shape_metric}
\end{figure*}

\ifisKDD\else
In this section, we first give full details of the experimental setup, in Section~\ref{sec:setup}. Then, we provide the analyses of the empirical results:
in Section~\ref{sec:trained_models}, we study Transformers trained from scratch with different hyperparameters; and in Section~\ref{sec:Huggingface}, we focus on pretrained Transformers from the Huggingface website.\fi

\subsection{Experimental setup}\label{sec:setup}

\noindent\textbf{Dataset.} In Section~\ref{sec:trained_models}, we study models trained on the WMT14 German to English (DE-EN) dataset~\citep{bojar2014findings}, commonly used as a benchmark for neural machine translation \citep{ott2018scaling,vaswani2017attention,shen2020simple,edunov2018understanding}.
WMT14 consists of 4.5 million sentence pairs for training.

\noindent\textbf{Hyperparameters.} To conduct correlational analysis, and to capture the relationship between the generalization metrics and model quality in different settings, we vary several hyperparameters: the number of samples (either 160K, 320K, 640K, 1.28M, 2.56M samples), the initial learning rate during training (across eight different rates), the model width (embedding dimension either 256, 384, 512, 768, or 1024), and the model depth (\{4, 5, 6, 7, 8\}-layer transformers).
Similar to prior works on correlational analysis \citep{jiang2019fantastic} for model selection, we construct a high-dimensional grid of different hyperparameters $\mathbf{\Theta} = \{(\theta_1, \ldots, \theta_K): \theta_1\in \mathbf{\Theta}_1, \ldots, \theta_K\in \mathbf{\Theta}_K\}$, so that we can compare models when one of the hyperparameters is varied.
Two separate high-dimensional grids with dimension $K=3$ are considered: (1) sample$\times$learning rate$\times$width; (2) sample$\times$learning rate$\times$depth.
Each grid contains 5$\times$8$\times$5=\numberofmodelsWMT of these training settings. In total, there are \numberofmodels trained models because the two high-dimensional grids overlap each other, and 40 models belong to both grids.
We will conduct three correlational analyses in the following to evaluate model selection performance.

\noindent\textbf{Task one, correlation evaluated on optimally trained models.}
In the first task (Section~\ref{sec:Optimal_training}), we measure the (rank) correlation between model quality and generalization metrics on models trained with the optimal choice of training hyperparameters, that is, if we are allowed to grid-search the best training hyperparameters, can we predict the best data size or model size parameters?

\noindent\textbf{Task two, correlation in time.} 
In the second task (Section~\ref{sec:time-wise}), we track BLEU score and generalization metrics during training, assessing time-wise correlation to model quality.
This task has been considered in the literature \citep{bartlett2017spectrally}, and from a practical point of view, capturing the time-wise dependence during training could potentially lead to better early stopping and regularization methods.

\noindent\textbf{Task three, correlation when a single hyperparameter is varied.}
In the third task \ifisKDD\else(Section~\ref{sec:Simpson})\fi, we study the relationship between the model quality and the generalization metrics when a single hyperparameter is varied.
Metrics that achieve a high (rank) correlation for all the hyperparameters are good candidates for model selection.

\noindent\textbf{Training and model setup.}
For the details of training Transformers on WMT14,
\ifisKDD
\else
see Appendix~\ref{app:experiment_setup}.
\fi

\subsection{Correlational analyses on Transformers trained in different settings}\label{sec:trained_models}

In this subsection, we study \numberofmetrics generalization metrics (with details provided in Table~\ref{tab:metrics}) and examine their correlations with BLEU score \citep{papineni2002bleu}, the most commonly used metric to evaluate machine translation \footnote{Several empirical metrics have been designed to measure the quality of text generation, such as BERTScore \citep{zhang2019bertscore} and BARTScore \citep{yuan2021bartscore}.
Our work is different because we do not need any data, and we do model selection using the ESDs of weight matrices only.
BERTScore and BARTScore evaluate the text quality, and thus they need source or reference texts generated by humans. These metrics can serve as alternatives to BLEU, which is viewed as ground truth in our work.}.
Note that BLEU score here is used as a close approximation of model quality, mimicking the role of test accuracy in image classification.
We also consider correlation between these metrics and the generalization gap, defined as the BLEU score for training data subtracted by the BLEU score for test data.
We intend to find generalization metrics that strongly correlate with model quality instead of the generalization gap.

\input{sections/Experiments_Optimal-hyperparameter}
\input{sections/Experiments_Time-wise}
\ifisKDD
\else
\input{sections/Experiments_Simpson_Full}
\fi

\noindent{\bf Corroborating results.} We extend our empirical evaluations to other datasets and evaluation methods.
First, we consider pretrained Huggingface Transformers in Section~\ref{sec:Huggingface}, providing model selection results in a broad range of NLP tasks.
Then, we consider three other language processing tasks trained with different Transformers, including
\begin{itemize}
    \item Roberta \citep{liu2019roberta} trained on the masked language modeling task using Wikitext-103 \citep{merity2016pointer}, and then finetuned on MNLI \citep{williams2018broad};
    \item Six-layer base Transformers trained on the language modeling task using the Wikitext-103 dataset \citep{merity2016pointer};
    \item Six-layer base Transformers trained on the next-word prediction task using the Reddit dataset, following the implementation in~\citet{bagdasaryan2020backdoor}.
\end{itemize}
\ifisKDD
\else
All extended results can be found in Appendix~\ref{app:additional_results}. Also, in Appendix~\ref{app:additional_results},
\fi
we provide additional results on conducting correlational analysis using Kendall's tau instead of Spearman's rank correlation.

\noindent{\bf Computational cost and carbon emission.}
We believe it is extremely important that papers relying on large-scale empirical analysis accurately report the computational cost.
The overall training cost is 7301.66 GPU hours. We use GPU nodes with TITAN RTX for our training. The overall carbon emission depends on carbon efficiency. Using the default values from the online Machine Learning Emissions Calculator\footnote{https://mlco2.github.io/impact/\#compute}, the total emissions are estimated to be 883.21 kg CO2 eq.

%% file: sections/Experiments_Optimal-hyperparameter.tex
\subsubsection{Task one: Evaluating correlations on optimally trained models only}
\label{sec:Optimal_training}

Here, we group models using the number of training samples, and select the best model from each group when the model depth and the learning rate are varied. In Figure~\ref{fig:Optimal_models_shape_metric}, each curve represents a group of models trained with a certain number of training samples. The black star on each curve represents training with optimal hyperparameters (learning rate and depth in our setting), obtained by searching for the optimum on a third-order polynomial fit of each curve.
From Figure~\ref{fig:Optimal_models_shape_metric}, we see that the shape metrics correctly predict the model quality for models trained with the optimal training hyperparameters, i.e., the BLEU scores should be higher when the metric values are smaller on the optimal models represented using black stars.
Since all six shape metrics show similar trends, a pairing of these metrics can be considered as a sanity check.

\noindent{\bf Comparison with scale metrics.}
\ifisKDD
\else
We compare scale metrics and shape metrics in Section~\ref{sec:Simpson}.
\fi
We show that shape metrics predict the correct trends in test BLUE scores, while scale metrics predict wrongly because they are correlated with the generalization gap.

\begin{remark}\normalfont
Figure~\ref{fig:Optimal_models_shape_metric} points out an important but subtle issue in empirically evaluating the HT-SR theory.
In Figure~\ref{fig:Optimal_models_shape_metric}, one can make a model less well-trained---and artificially anti-correlate the generalization metric with the task accuracy.
For example, see the gray dotted line in the first subfigure in Figure~\ref{fig:Optimal_models_shape_metric}.
\ifisKDD\else
In Section~\ref{sec:Huggingface}, we study models from the Huggingface website, and therefore, our results will strongly depend on how well-trained these models are.\fi
\end{remark}

%% file: sections/Experiments_Time-wise.tex
\subsubsection{Task two: Time-wise correlations and rank correlation results}\label{sec:time-wise}

\begin{figure*}
    \centering
    \includegraphics[width=\textwidth]{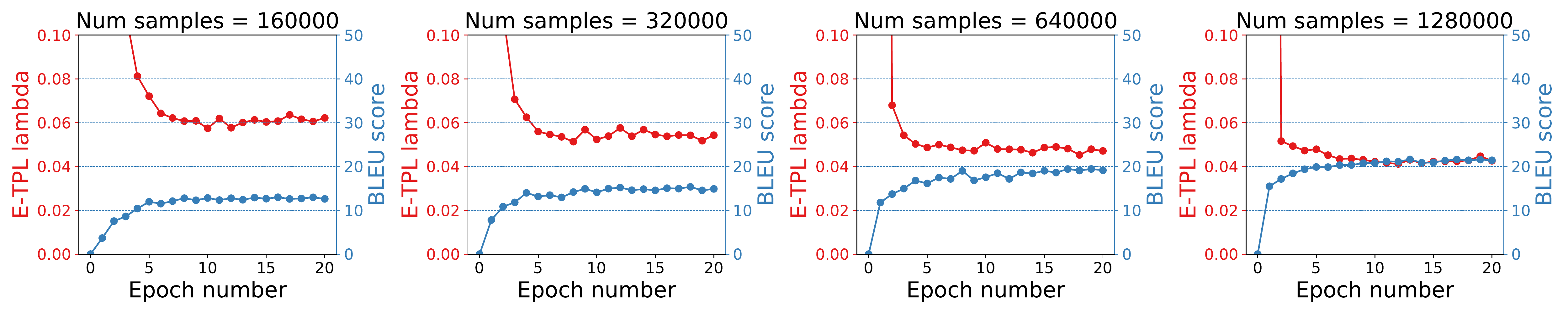}
    \includegraphics[width=\textwidth]{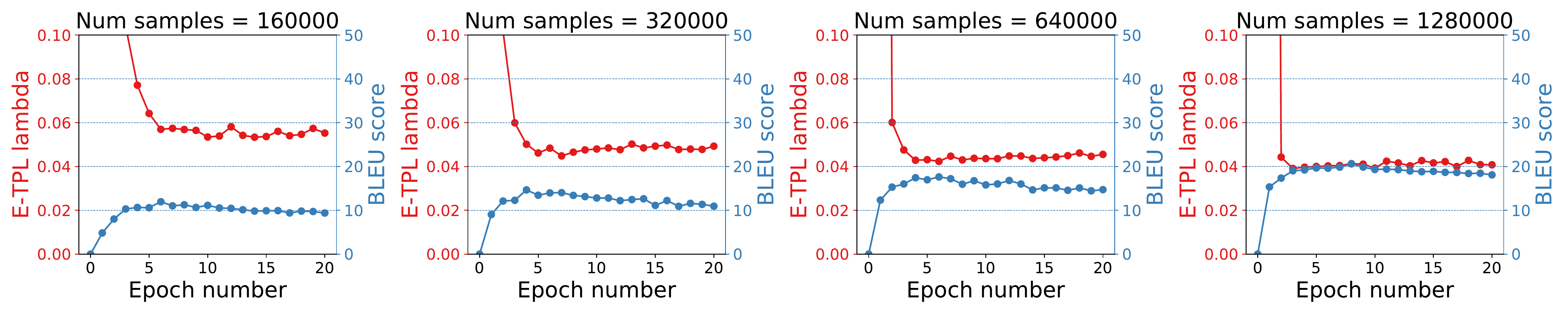}
    \caption{\EXPONENT closely tracks the BLEU score, i.e., BLEU score increases when the \EXPONENT drops. Results are shown for Transformers trained on WMT14 with different number of samples. {\bf(First row).} Training with dropout 0.1. {\bf(Second row).} Training without dropout.\vspace{-.01cm}}
    \label{fig:generalization}
\end{figure*}

\begin{figure*}
    \centering
    \begin{subfigure}{0.49\textwidth}\centering
        {\small\textbf{Correlations with model quality}} \\
         \includegraphics[width=0.95\textwidth]{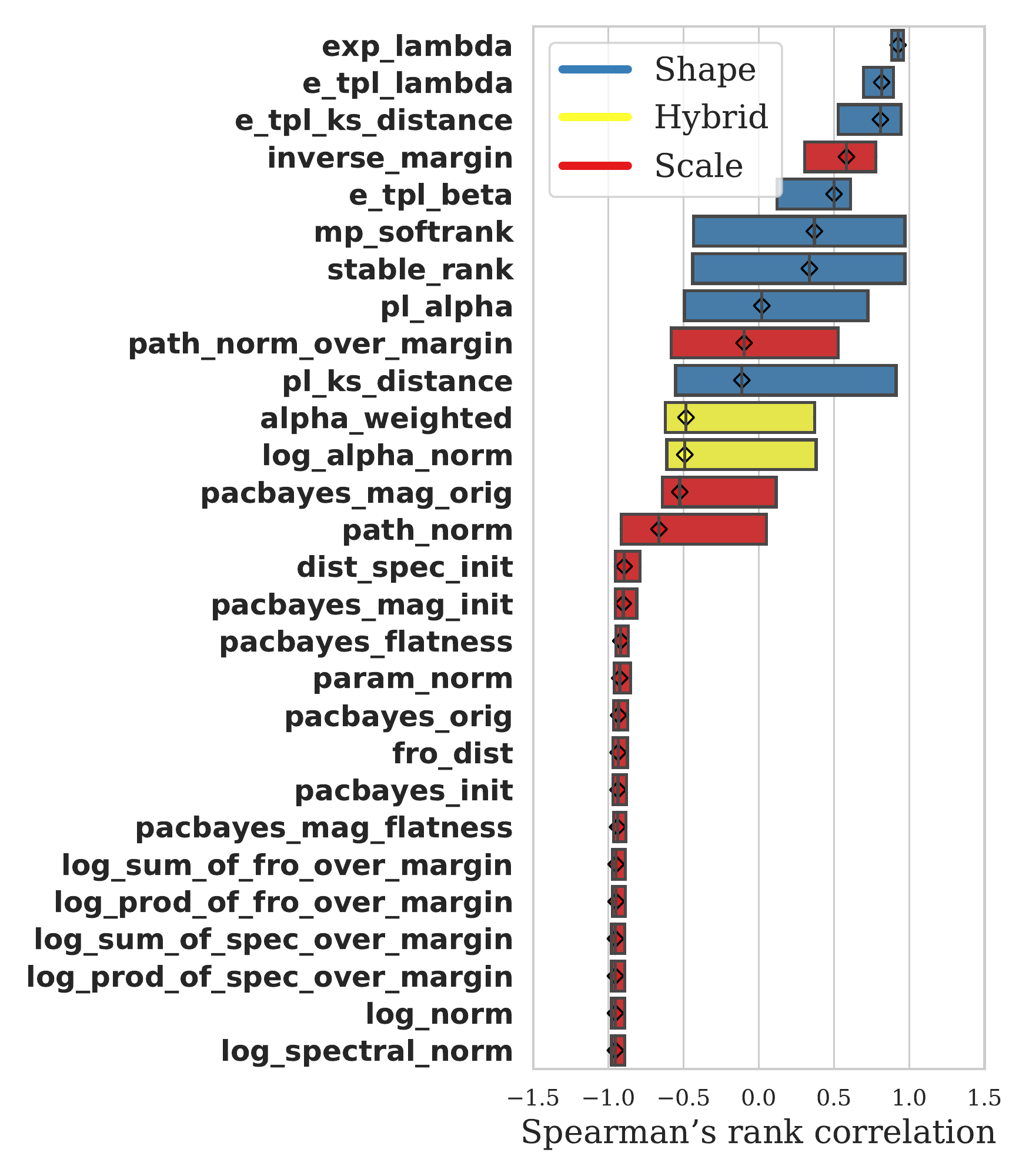}
        \caption{\textbf{Correlations with model quality.} Spearman’s rank correlation between various generalization metrics and BLEU.
}
        \label{fig:ave_worst_rank_correlations}
    \end{subfigure}\hfill
    \begin{subfigure}{0.49\textwidth}\centering
        {\small\textbf{Correlations with generalization gap}} \\
        \includegraphics[width=0.95\textwidth]{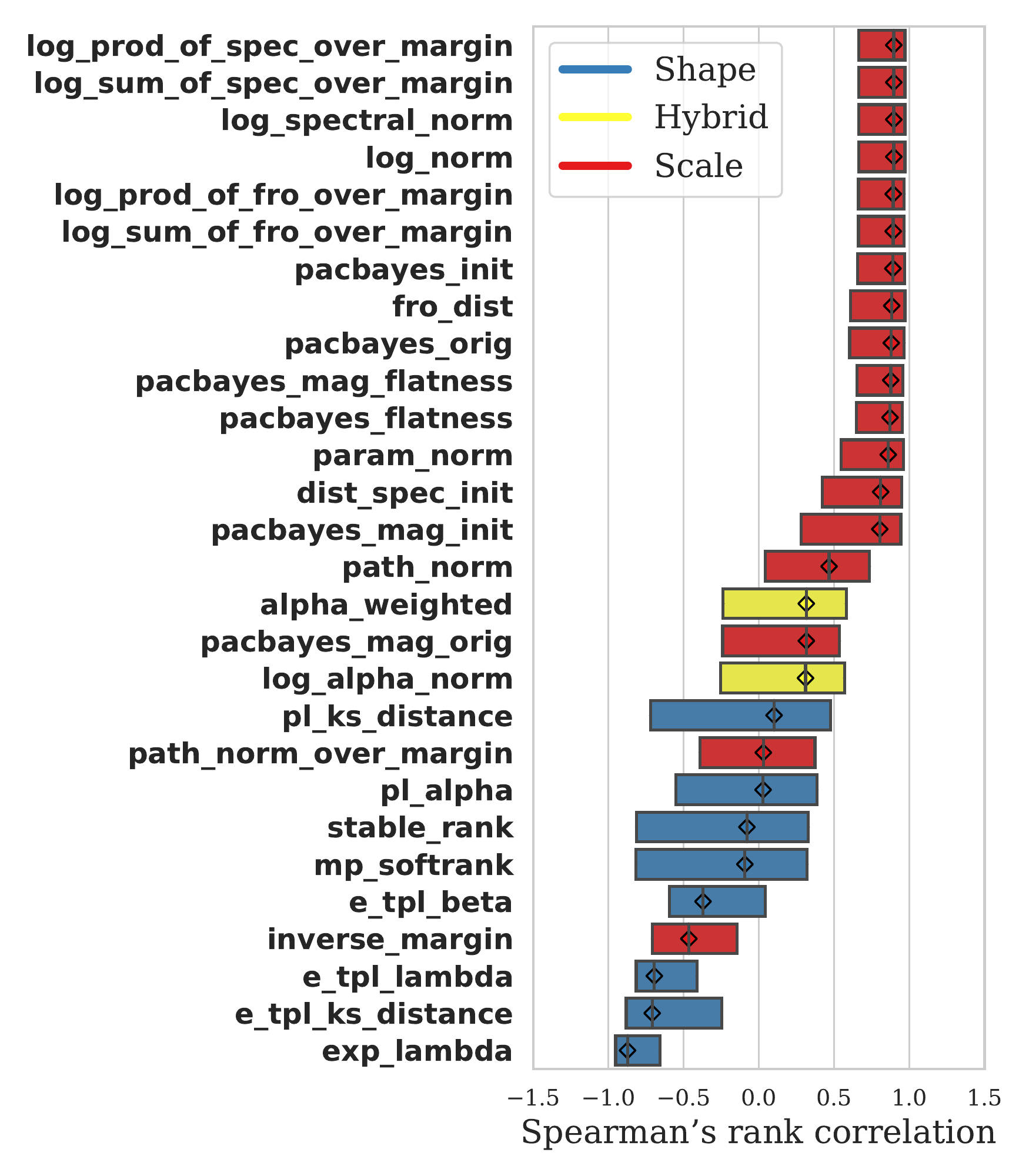}
        \caption{\textbf{Correlations with generalization gap.} Spearman’s rank correlation between various generalization metrics and generalization gap. 
        }
        \label{fig:ave_worst_rank_correlations_gap}
    \end{subfigure}
    \caption{Comparing multiple generalization metrics for predicting BLEU score (on the left) or the generalization gap (on the right).
    Lines on each box delineate the 25/50/75 percentiles of the rank correlations in \numberofmodels different settings (including different amount of data, different network depths, different network widths, and different initial learning rates). \vspace{-.0cm}
    }
    \label{fig:test_acc_vs_generalization_gap}
\end{figure*}

In this subsection, we study time-wise correlation between our chosen metrics and the BLEU scores. \ifisKDD\else
In other words, we see if a generalization metric can track the test curve during training.\fi

\noindent{\bf \EXPONENT tracks the BLEU score.}
As a warm-up, we consider how well the \EXPONENT metric defined in \eqref{eqn:EXPONENT} tracks the BLEU score (recalling that \EXPONENT assumes the ESDs follow E-TPLs). 
We use training with and without dropout to study the effect of training schemes, and we consider different quantities of data to test robustness when the size of data changes.
In Figure~\ref{fig:generalization}, the first row considers models trained with dropout,
while the second row considers models trained without dropout.
The multiple columns track \EXPONENT and the BLEU score throughout training for different amounts of data.
We can see that \EXPONENT not only successfully tracks BLEU scores but also differentiates underfitting (first row, with dropout) from overfitting (second row, without dropout) in this~experiment.

\noindent{\bf Shape metrics predict model quality, while scale metrics predict the generalization gap.} Now we consider the rank correlations between our chosen metrics and the test BLEU score. The rank correlations are evaluated across training, i.e., for each of the \numberofmodels settings of the hyperparameters, we calculate the Spearman's rank correlation between BLEU scores and the values of each generalization metric over all epochs.
The summarized results are presented in Figure~\ref{fig:ave_worst_rank_correlations}.
A positive Spearman's rank correlation (with BLEU) suggests that the generalization metric is useful in tracking BLEU during training. A negative Spearman's rank correlation, on the other hand, implies that the metric often gives the incorrect prediction.
In~Figure~\ref{fig:ave_worst_rank_correlations}, we use the average rank correlations for all settings to study the effectiveness of each metric, and present 25\% quantile rank correlations to indicate robustness across runs.

In Figure~\ref{fig:ave_worst_rank_correlations}, we find shape metrics, such as~\ETPLKSDISTANCE, \EXPDISTEXPONENT, \EXPONENT, and \ETPLBETA, exhibit some of the highest rank correlations with BLEU score. The \EXPDISTEXPONENT metric, which assumes a EXP distribution on the ESDs, achieves the highest median rank correlation, while the \EXPONENT metric, which assumes a E-TPL distribution on the ESDs, achieves the second highest.
\ifisKDD\else
We discuss the \INVERSEMARGIN metric in Appendix~\ref{app:scale_metrics}.\fi

In Figure~\ref{fig:ave_worst_rank_correlations_gap}, we plot the rank correlations to the generalization gap across our chosen metrics. While it is encouraging that most existing generalization metrics yield correct predictions, as previously discussed, correct predictions of the generalization gap do \emph{not} imply accurate predictions on the best-performing models~here.

\noindent{\bf Details of the rank correlation calculations.}
When calculating the rank correlation with the test accuracy, we associate a negative sign to all the generalization metrics, i.e., a positive rank correlation in Figure~\ref{fig:ave_worst_rank_correlations} means that a generalization metric is negatively correlated with the BLEU score. We use this procedure to follow the conventional wisdom that a smaller value of the complexity metric leads to better generalization~\citep{jiang2019fantastic}.
On the other hand, for Figure~\ref{fig:ave_worst_rank_correlations_gap}, a positive rank correlation means that the metric is positively correlated with the generalization gap. Thus, for both Figure~\ref{fig:ave_worst_rank_correlations} and \ref{fig:ave_worst_rank_correlations_gap}, a strong positive correlation corresponds to the expected trend.

\noindent{\bf Can we utilize anti-correlation for prediction?}
One may ask if the anti-correlation shown in Figure~\ref{fig:ave_worst_rank_correlations_gap} implies that scale metrics can also predict model quality.
Indeed, from Figure~\ref{fig:ave_worst_rank_correlations_gap} alone, it seems that one can negate the predicted results of scale metrics to obtain an accurate prediction.
However, note this strong negative correlation of scale metrics only holds in this one particular scenario.
In other scenarios, such as in~\citet{jiang2019fantastic,dziugaite2020search}, the correlation is strong in the other direction.
Broadly speaking, if a particular theory says that a quantity should go up with model quality, and it goes down sometimes instead, then the theory is incomplete, regardless of how strong the correlation is.
A prominent claim in our paper is that the correlation between test error and the generalization gap can sometimes be reversed. Therefore, it is insufficient to study metrics that have a large rank correlation with the generalization gap.

%% file: sections/Experiments_Simpson_KDD.tex
\subsubsection{Task three: evaluating correlation when a single hyperparameter is varied}\label{sec:simpson_partial_results}

In this subsection, we assess whether the generalization metrics can predict trends in BLEU score when a single hyperparameter is changed.
Specifically, for a hyperparameter space $\mathbf{\Theta} = \{(\theta_1, \ldots, \theta_K): \theta_1\in \mathbf{\Theta}_1, \ldots, \theta_K\in \mathbf{\Theta}_K\}$, we consider each one-dimensional slice of the form \[\{(\theta_1, \ldots, \theta_K): \theta_i \in \mathbf{\Theta}_i \text{ while other parameters $\theta_j, j\neq i$ are fixed}\},\] 
and we calculate the rank correlation using the models in each such slice.
Then, we aggregate the rank correlations from all the one-dimensional slices and plot the distributions of the rank correlations.
For example, if we evaluate the trends when the initial learning rate is varied, we choose $\mathbf{\Theta}_i$ to be the set of eight different initial learning rates mentioned in Section~\ref{sec:setup}, ``Hyperparameters''.
As another example, we can define $\mathbf{\Theta}_i$ to be the set of five different numbers of samples to study the (rank) correlation when the number of samples is varied.

Similar to Figure~\ref{fig:test_acc_vs_generalization_gap}, we provide the rank correlation results on both the test BLEU scores and the generalization gap. See Section 3.2.3 of our report online~\citep{yang2022evaluating}.
Again, shape metrics have better rank correlations with model quality, while scale metrics are better correlated with the generalization gap.

%% file: sections/Experiments_Simpson_Full.tex
\subsubsection{Task three: Rich correlational structures and the Simpson's paradox when data size, model size and training hyperparameters are varied}
\label{sec:Simpson}

\begin{figure*}
    \centering
    \includegraphics[width=0.32\textwidth]{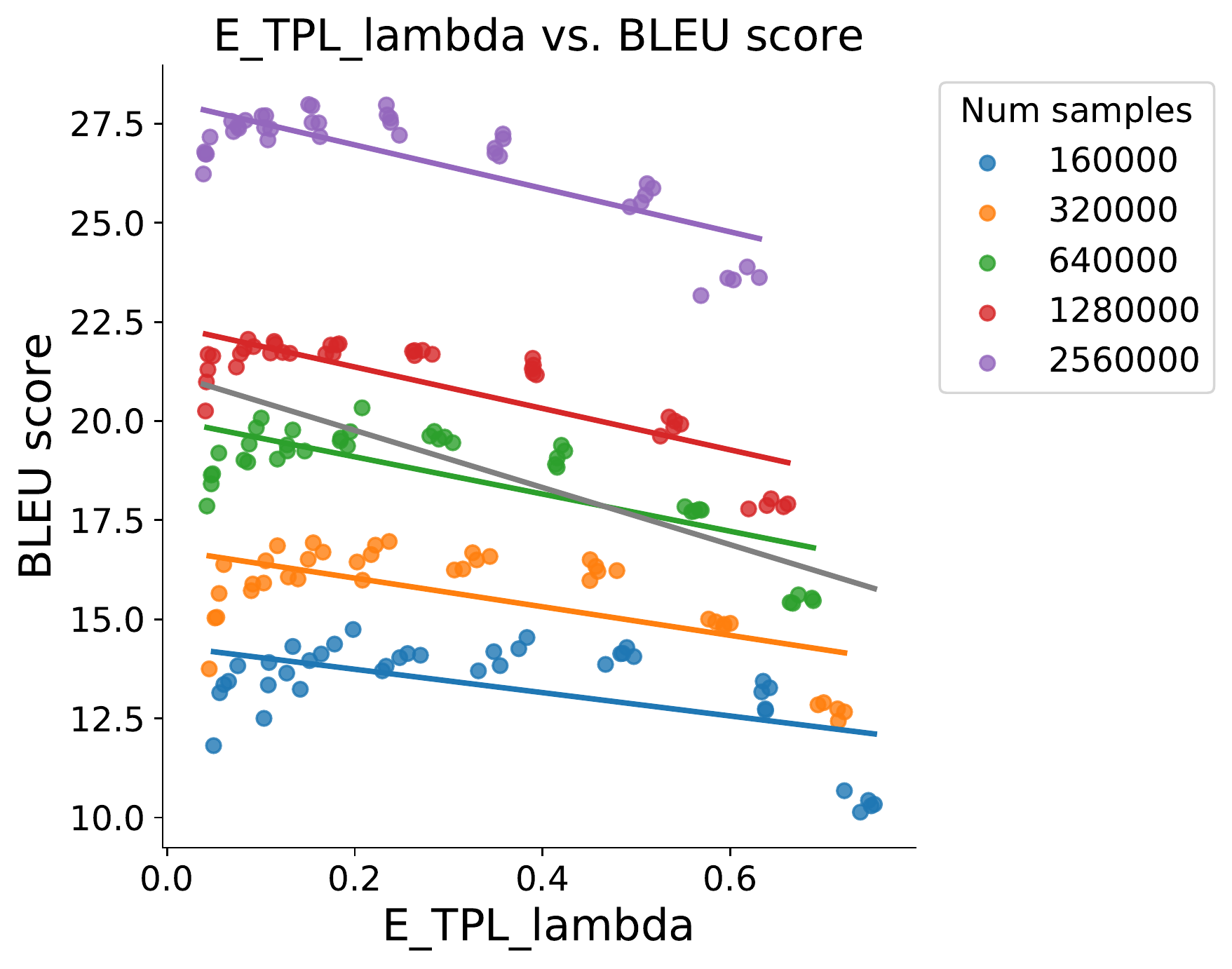}
    \includegraphics[width=0.32\textwidth]{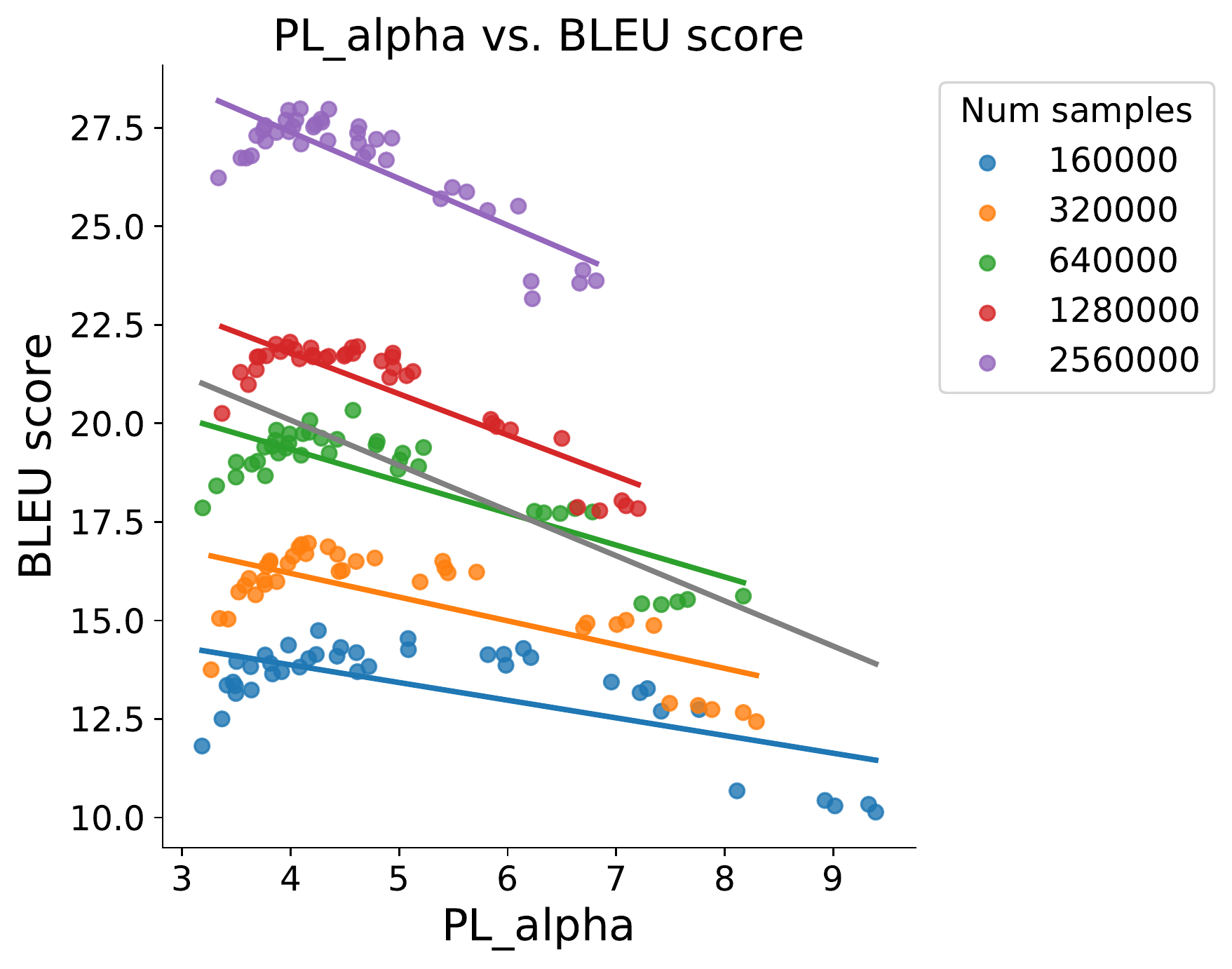}
    \includegraphics[width=0.32\textwidth]{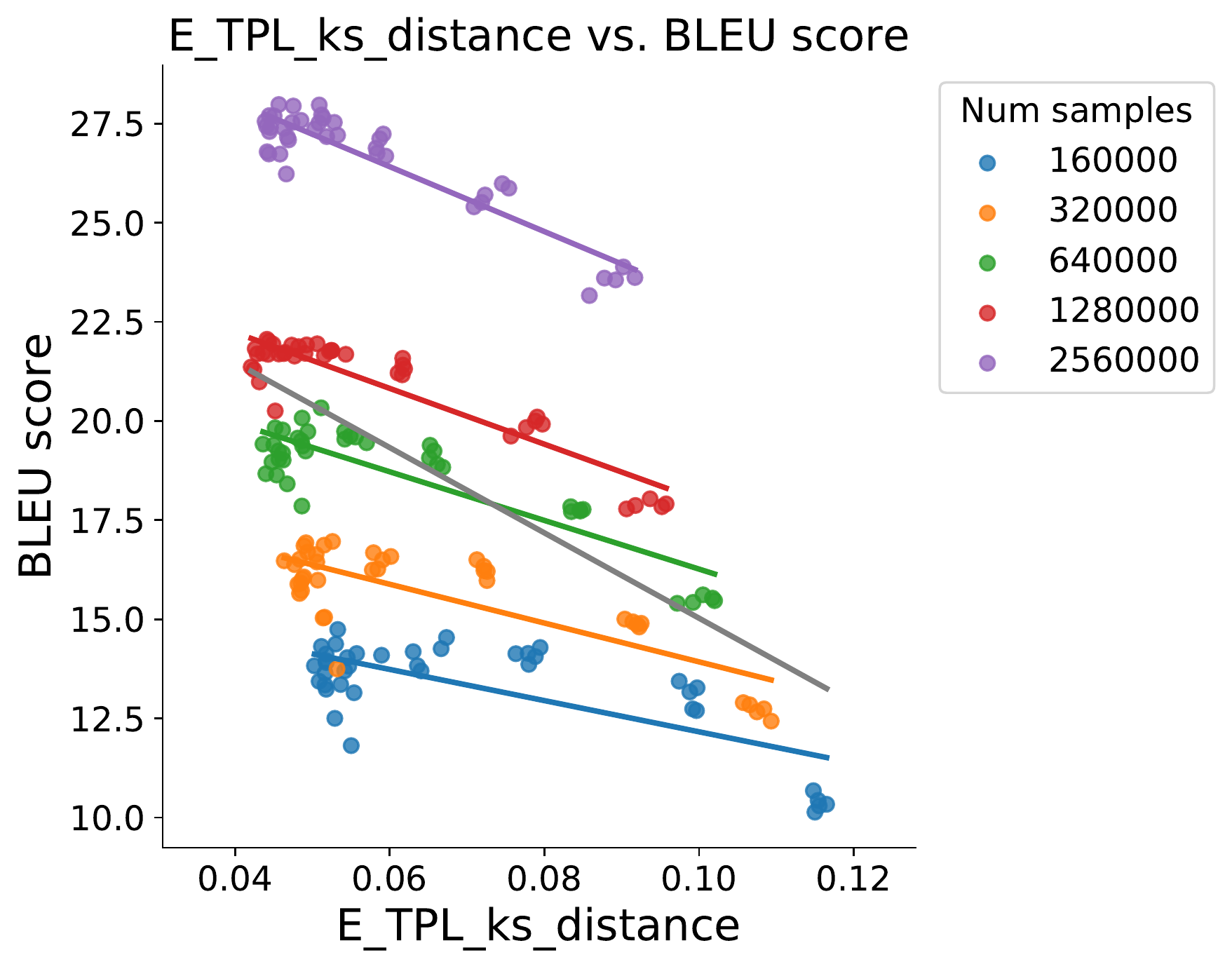}
    \includegraphics[width=0.32\textwidth]{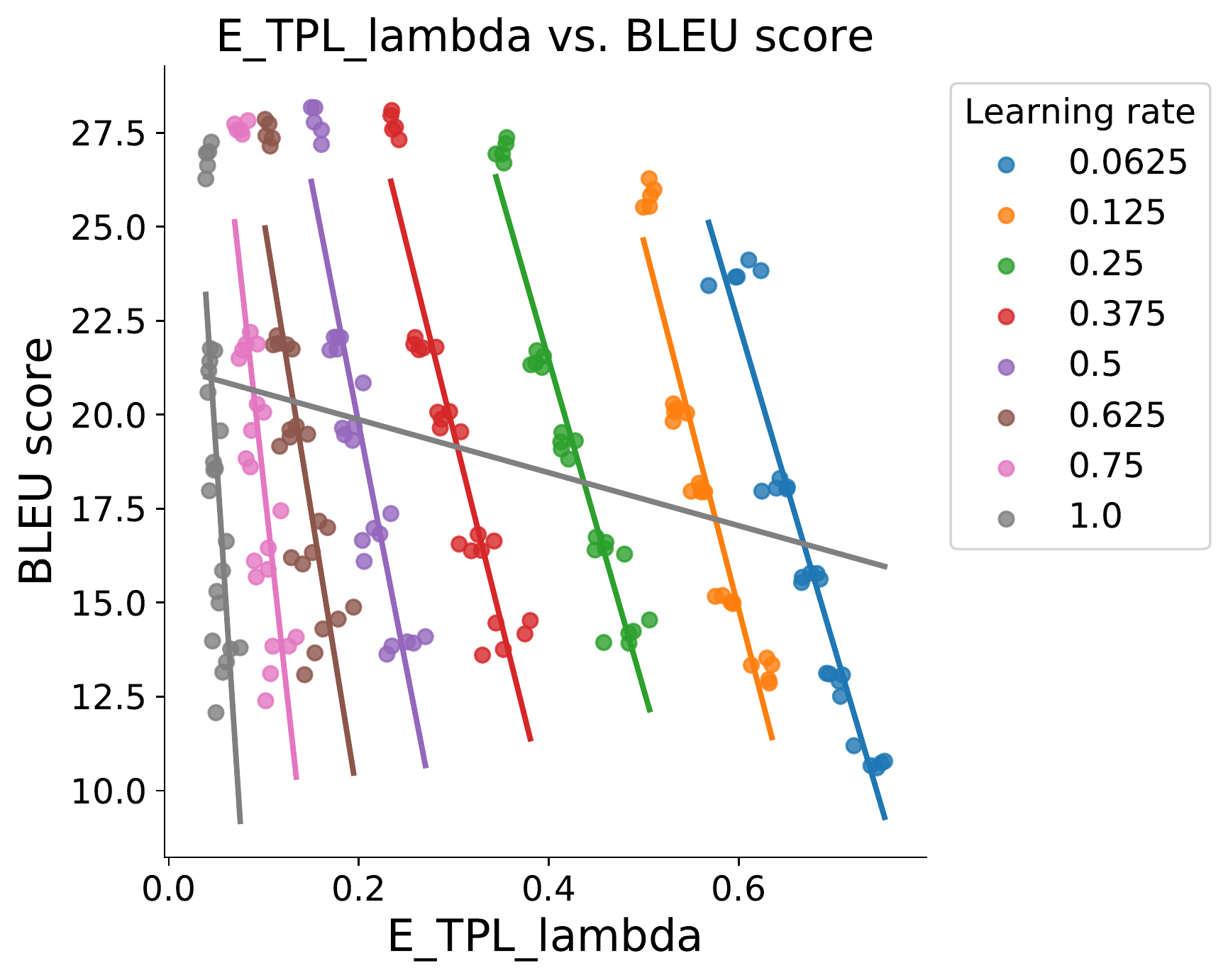}
    \includegraphics[width=0.32\textwidth]{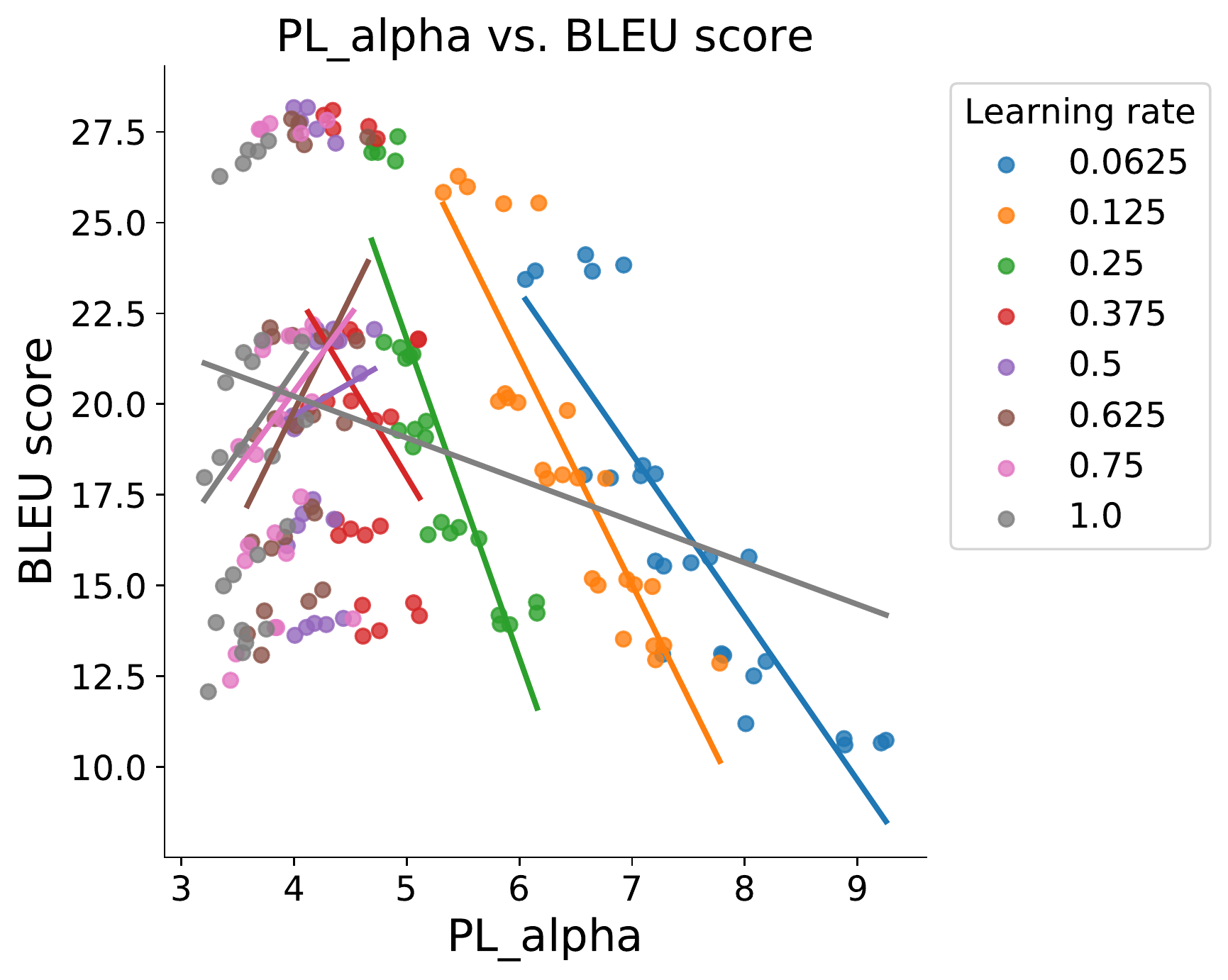}
    \includegraphics[width=0.32\textwidth]{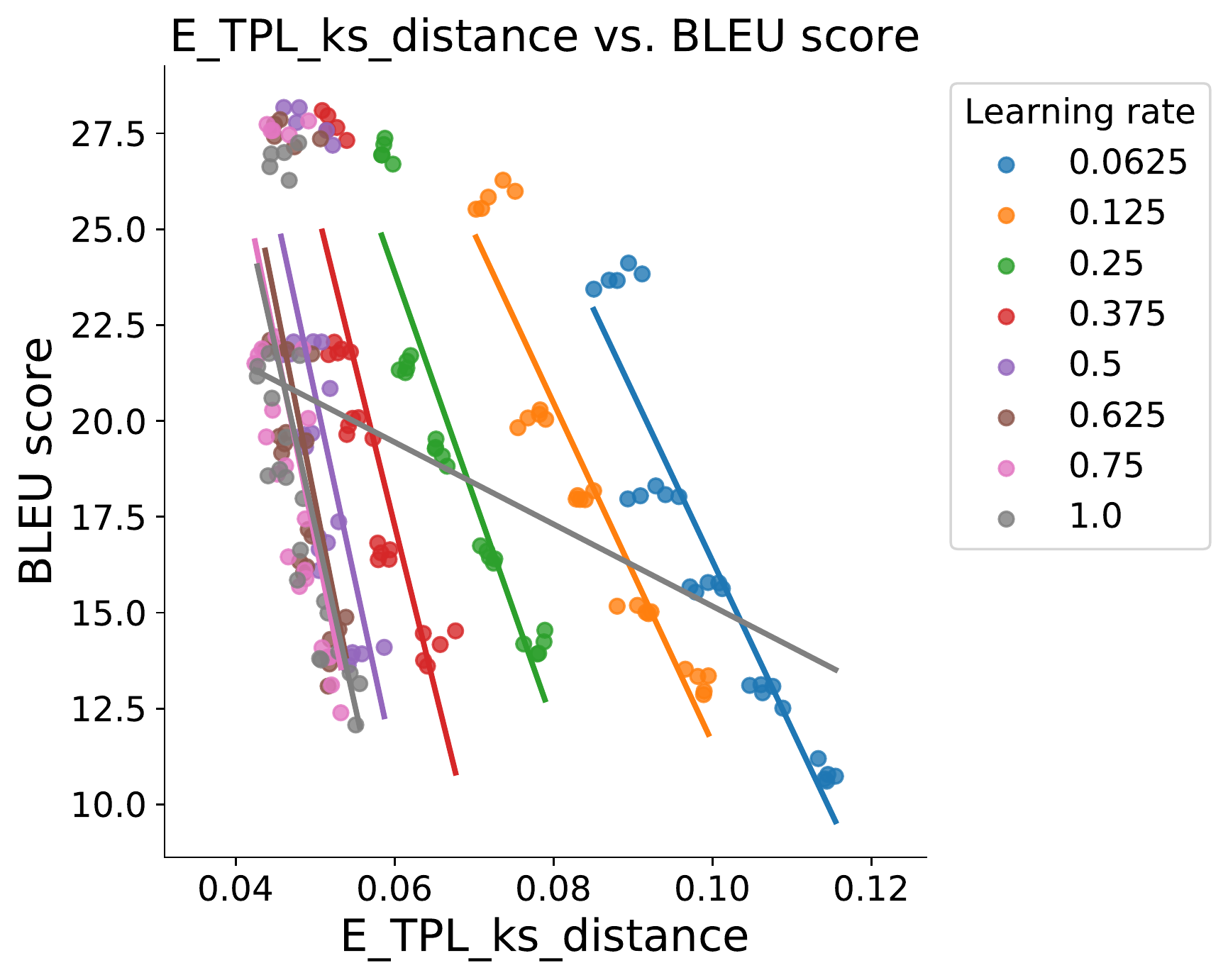}
    \caption{BLEU-score versus shape metrics for \numberofmodelsWMT Transformers trained on WMT14 with different hyperparameters.
    {\bf(First row)} Trained models grouped by the number of samples.
    {\bf(Second row)} Trained models grouped by the learning rate.
    The BLEU scores and the evaluated shape metrics display the correct (downward) trend.}
    \label{fig:Shape_metrics}
\end{figure*}

\begin{figure*}
    \centering
    \includegraphics[width=0.32\textwidth]{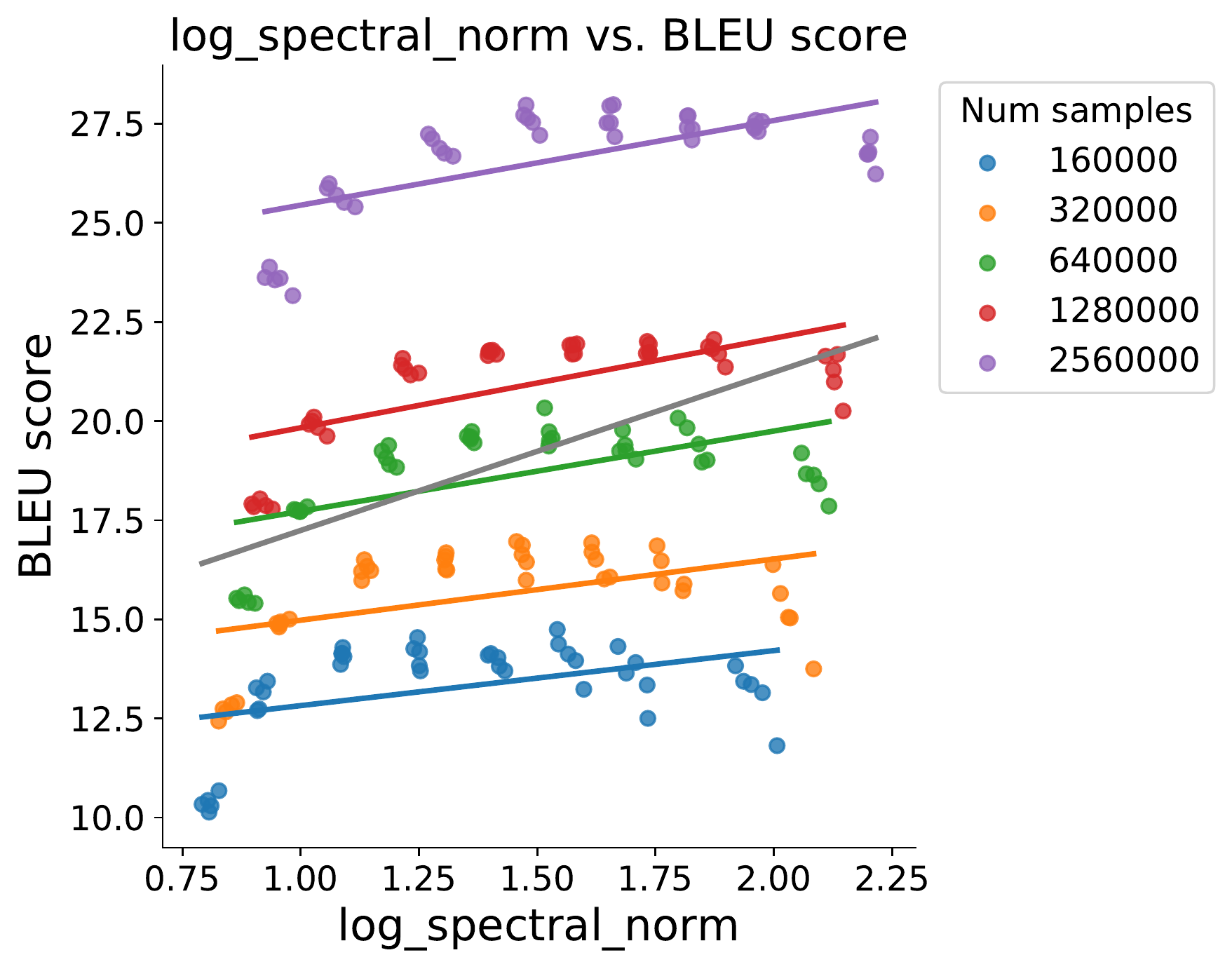}
    \includegraphics[width=0.32\textwidth]{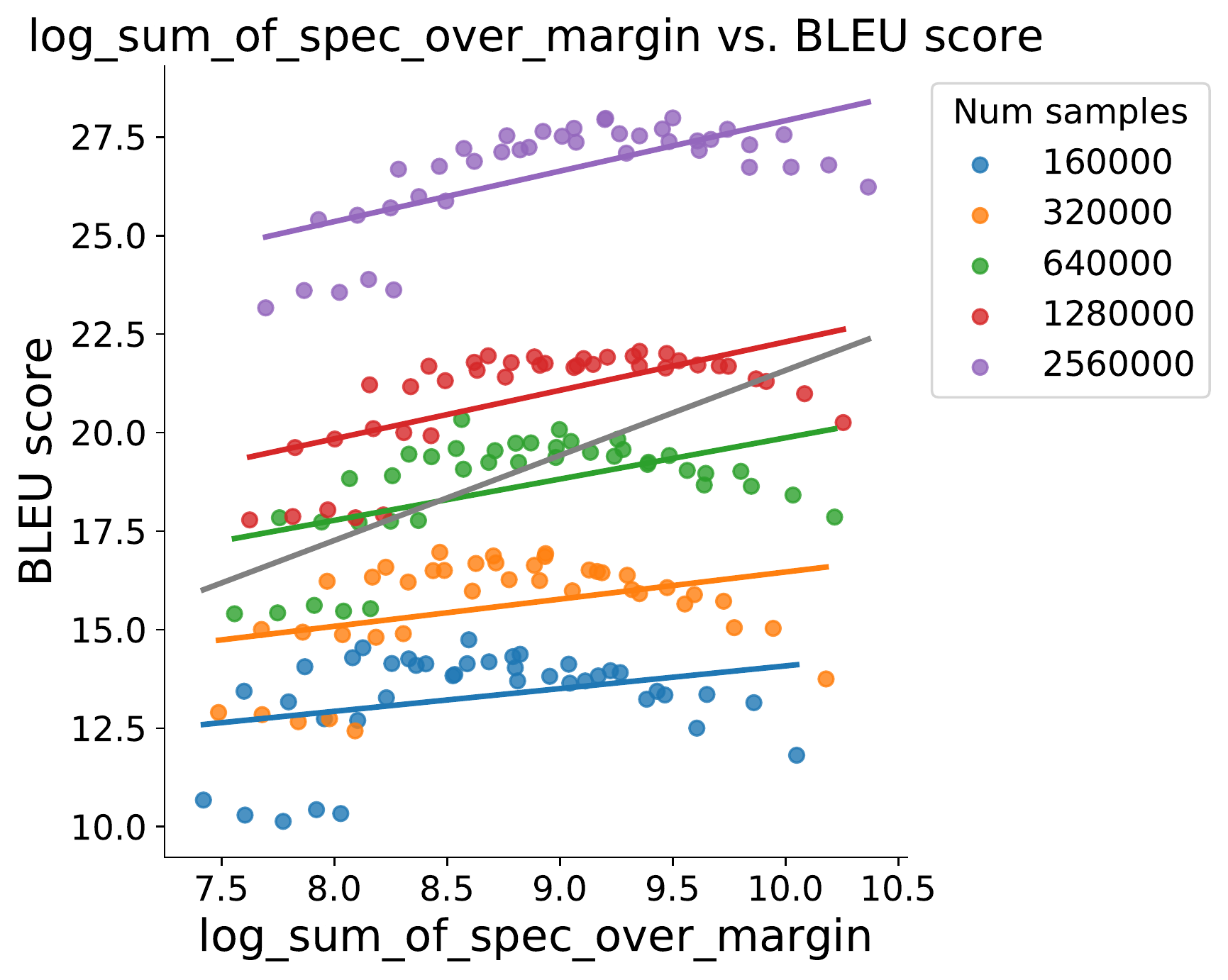}
    \includegraphics[width=0.32\textwidth]{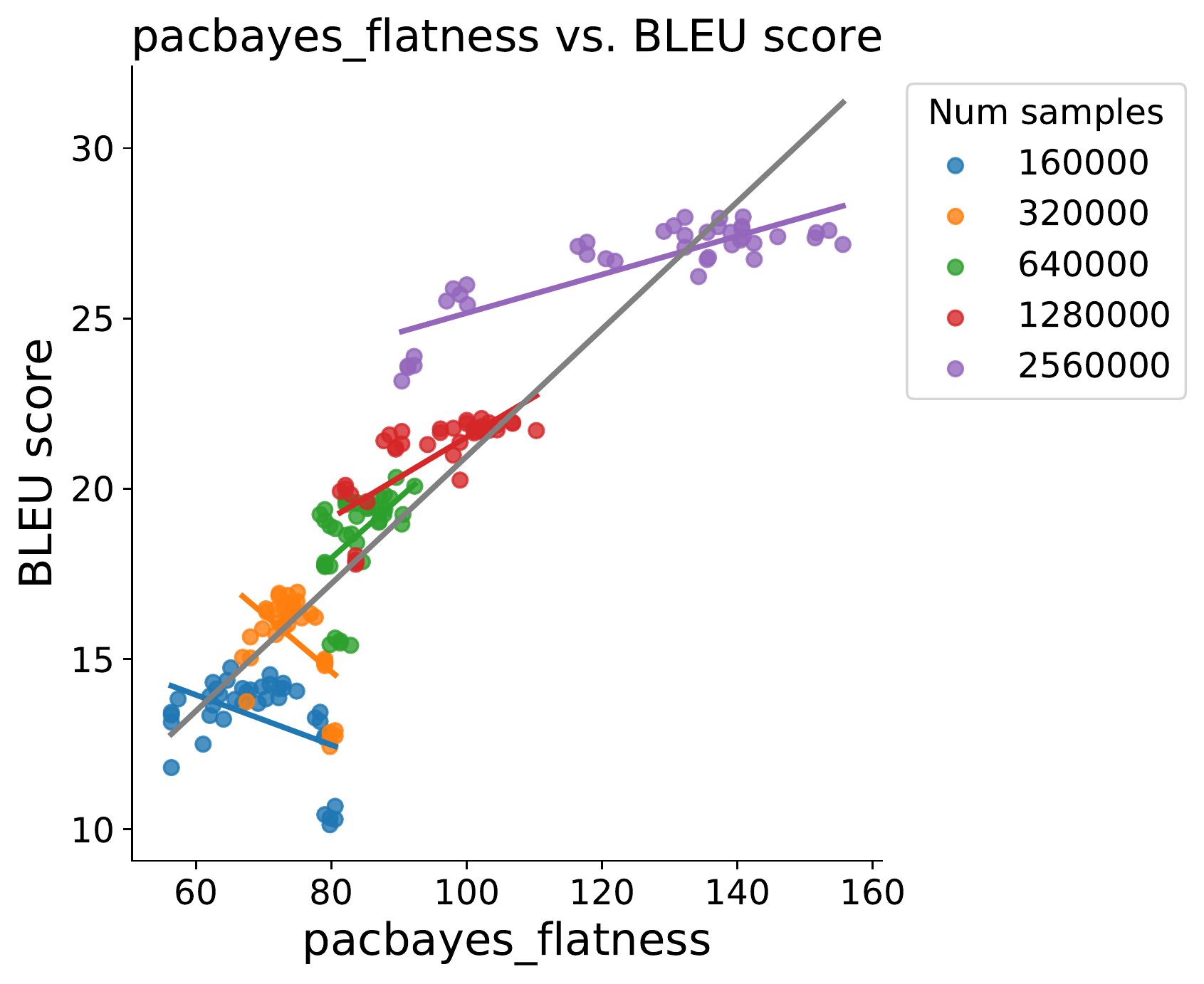}
    \includegraphics[width=0.32\textwidth]{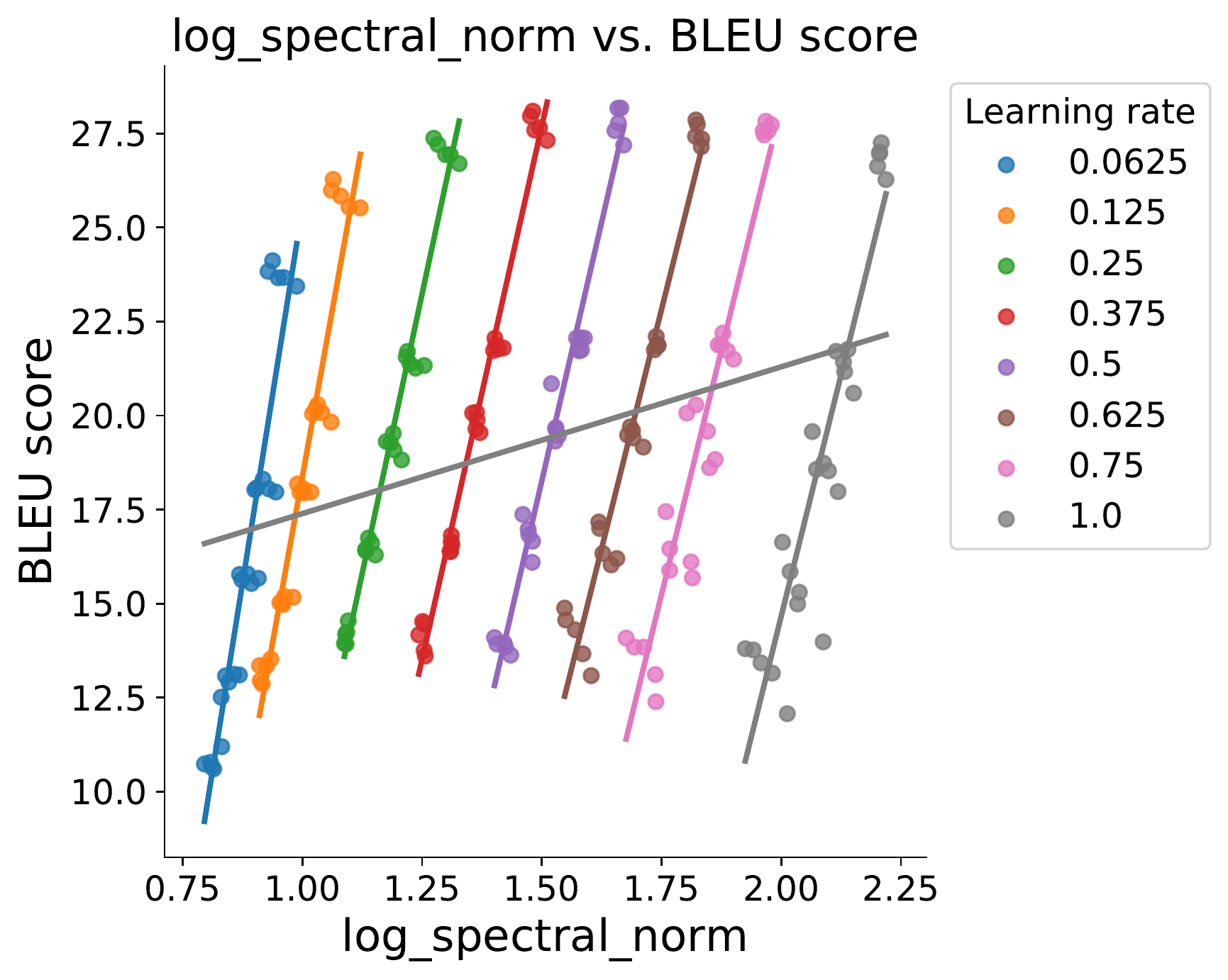}
    \includegraphics[width=0.32\textwidth]{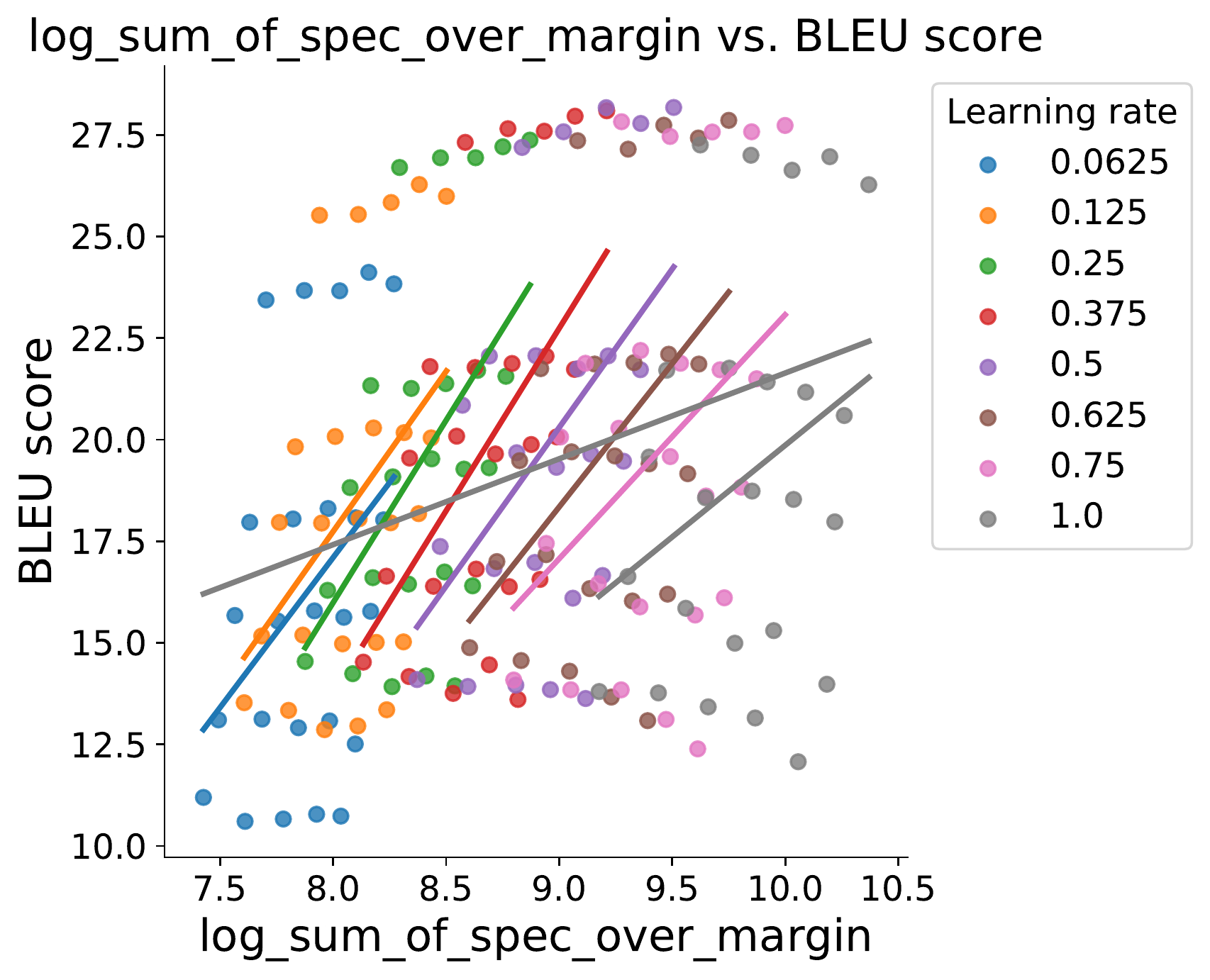}
    \includegraphics[width=0.32\textwidth]{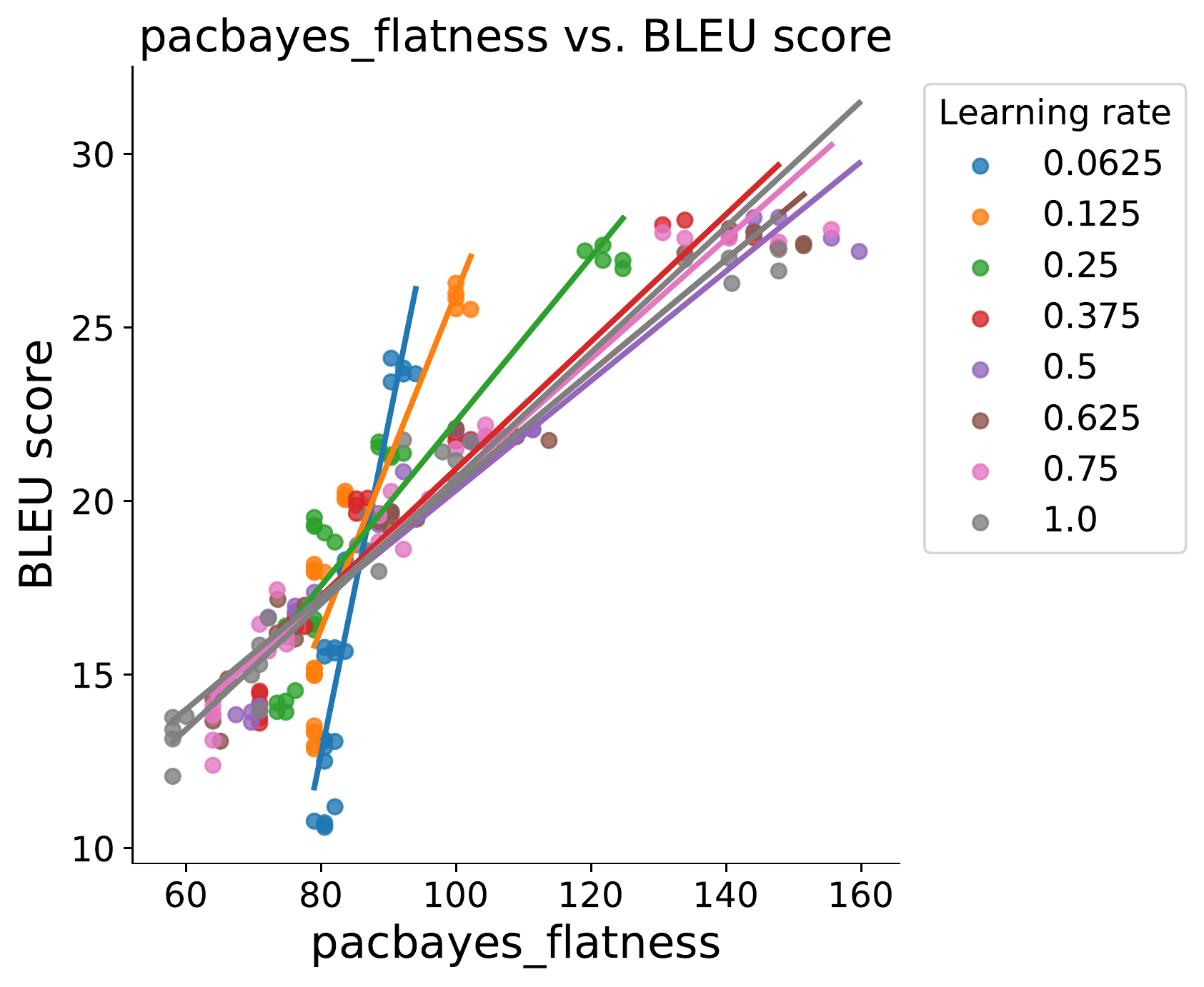}
    \caption{BLEU-score versus scale metrics for \numberofmodelsWMT Transformers trained on WMT14 with different hyperparameters. {\bf(First row)} Trained models grouped by the number of samples.
    {\bf(Second row)} Trained models grouped by the learning rate.
    The BLEU scores and the evaluated shape metrics display the wrong (upward) trend.}
    \label{fig:Scale_metrics}
\end{figure*}

\begin{figure}
    \centering
    \begin{subfigure}{0.32\textwidth}
    \includegraphics[width=\textwidth]{figs/id_bleu_E_TPL_lambda_lr_simpson.pdf}
    \caption{Grouping by learning rates.\label{fig:group_by_lr}}
    \end{subfigure}
    \begin{subfigure}{0.32\textwidth}
    \includegraphics[width=\textwidth]{figs/id_bleu_E_TPL_lambda_sample_simpson.pdf}
    \caption{Grouping by number of samples.\label{fig:group_by_sample}}
    \end{subfigure}
    \begin{subfigure}{0.32\textwidth}
    \includegraphics[width=0.86\textwidth]{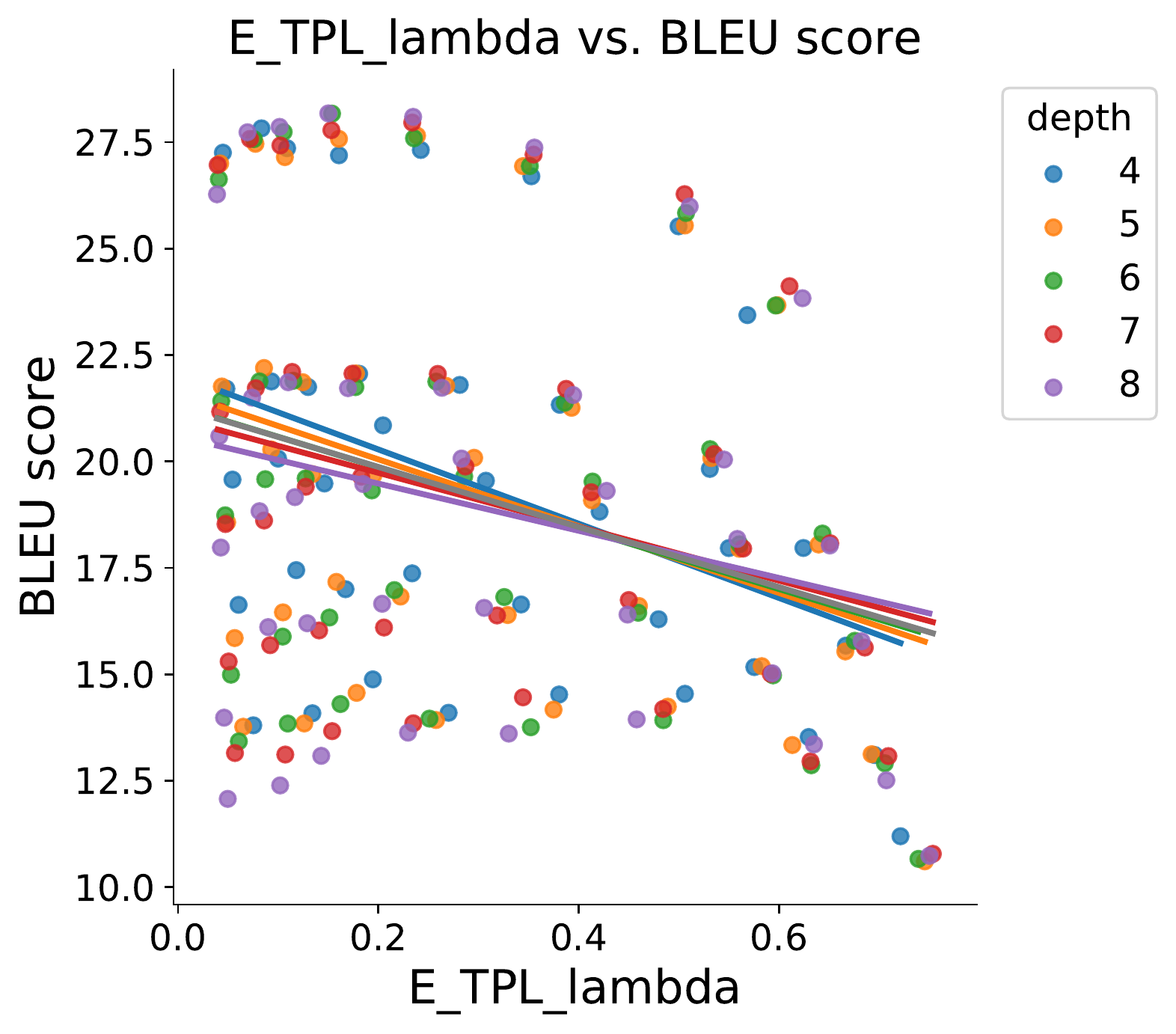}
    \caption{Grouping by network depths.\label{fig:group_by_depth}}
    \end{subfigure}
    \caption{\EXPONENT vs BLEU for the same set of trained models grouped by different hyperparameters. While BLEU scores change significantly when the initial learning rates and the number of samples are varied, they almost remain fixed for varying network depths.}
    \label{fig:depth}
\end{figure}

For our final task, we vary each of the hyperparameters and study the trends of the generalization metrics.
We first focus on the sample$\times$learning rate$\times$depth hyperparameter grid---
see Figures~\ref{fig:Shape_metrics} and \ref{fig:Scale_metrics} for plots of BLEU score against shape and scale metrics, respectively.
Since we vary the number of samples, the learning rate, and the model depth to obtain different models, we group these models to visualize trends over each hyperparameter.
In each subfigure, we color-code the models by either learning rate or the number of samples.
We discuss grouping models by depth later in Figure~\ref{fig:group_by_depth}.
Note that Figure~\ref{fig:Shape_metrics} partially overlaps with Figure~\ref{fig:Optimal_models_shape_metric}, except for different fitting methods.

For each curve, we expect the generalization metrics to be negatively correlated with the models’ quality measured using the BLEU score, i.e., the regression lines should have negative slopes. Comparing Figure~\ref{fig:Shape_metrics} and~\ref{fig:Scale_metrics}, one can see that the \emph{shape metrics tend to show the correct trends (which are more negatively correlated) than the scale metrics}.

\begin{remark}\normalfont
In Figure~\ref{fig:Shape_metrics}, constrained by the least-squares fitting, some regression lines are not aligned well with data, e.g., the second figure on the second row. In Appendix~\ref{sec:odr}, we fit the data using the orthogonal distance regression to mitigate this issue.
\end{remark}

\begin{figure*}
    \centering
    \includegraphics[width=0.32\textwidth]{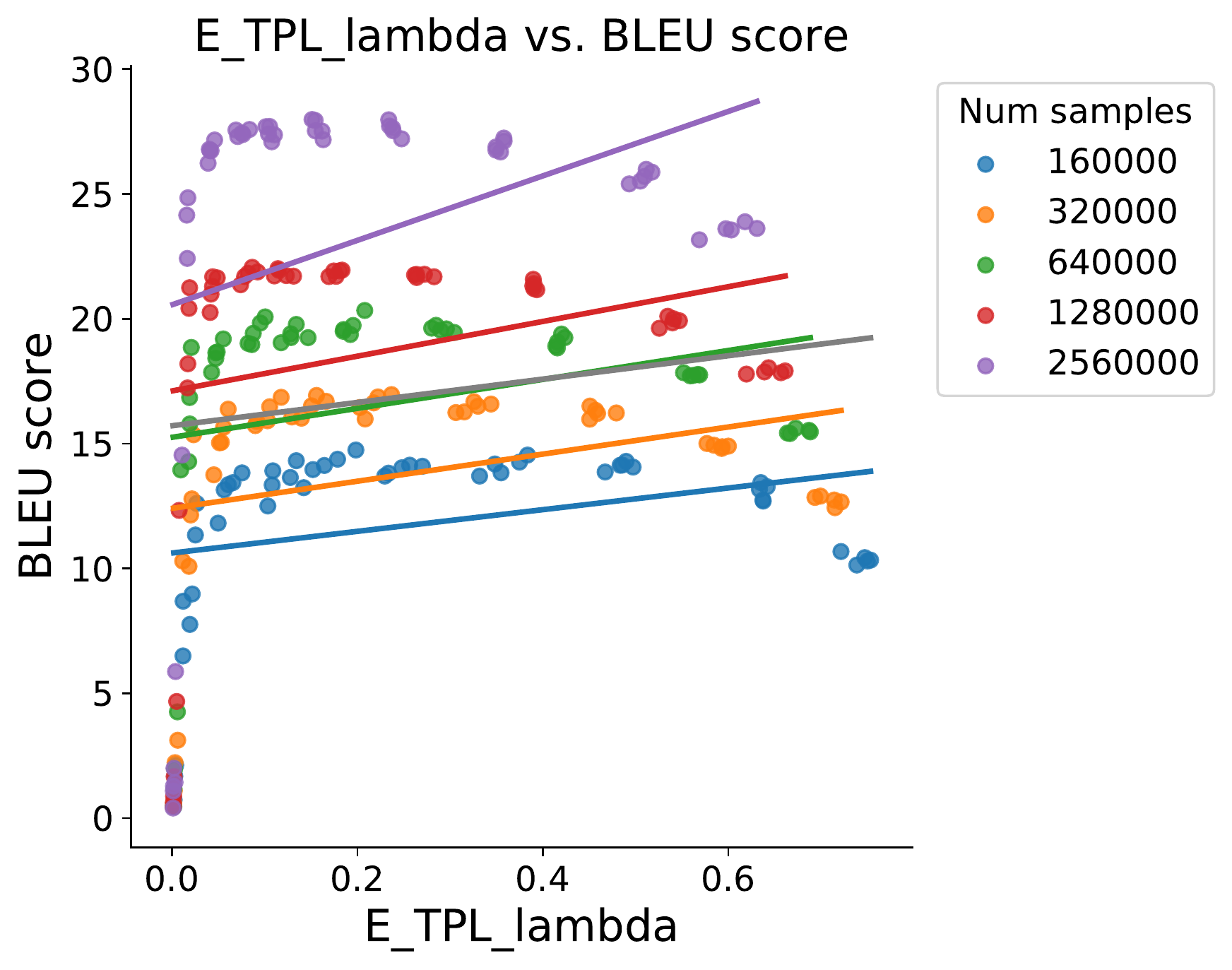}
    \includegraphics[width=0.32\textwidth]{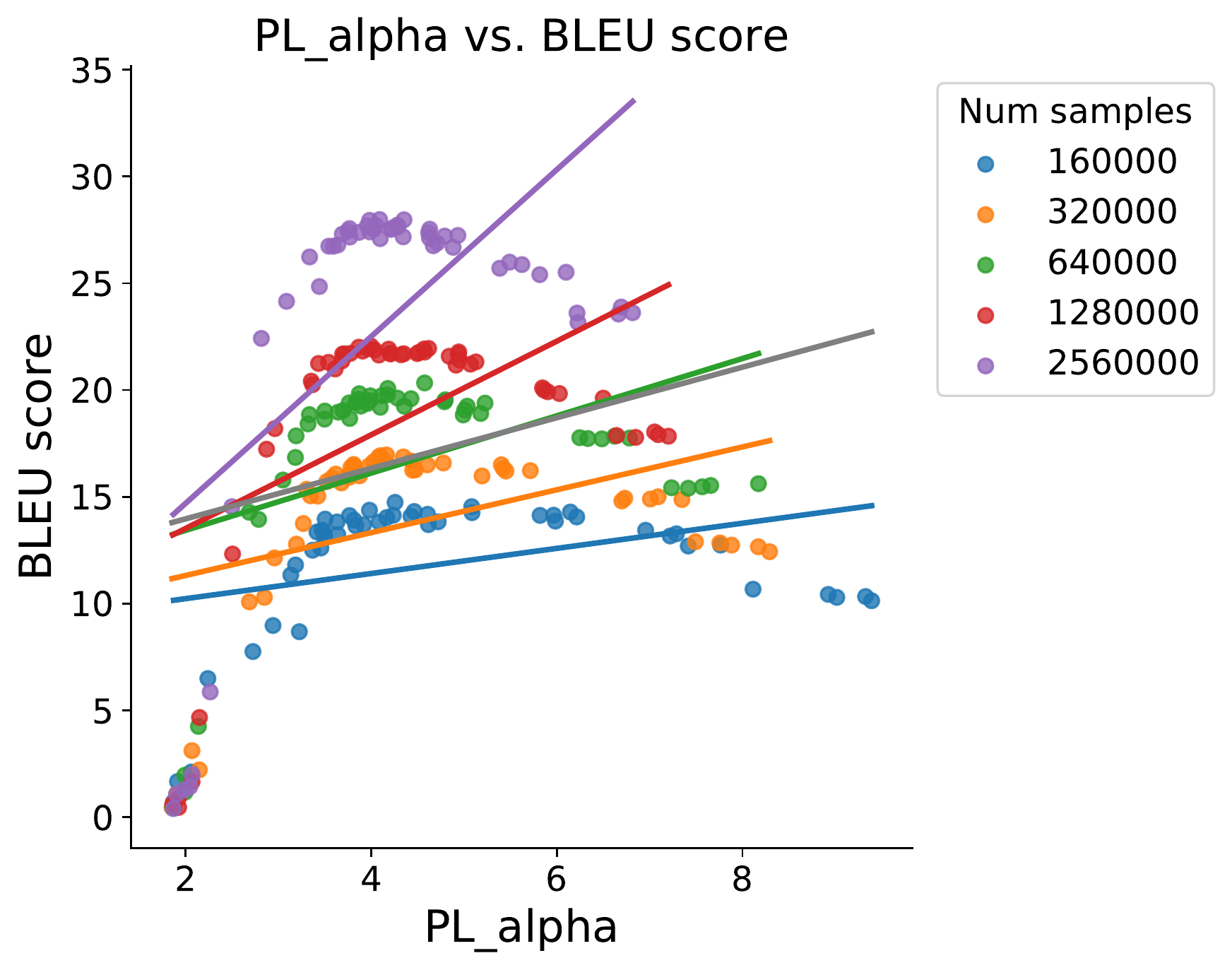}
    \includegraphics[width=0.32\textwidth]{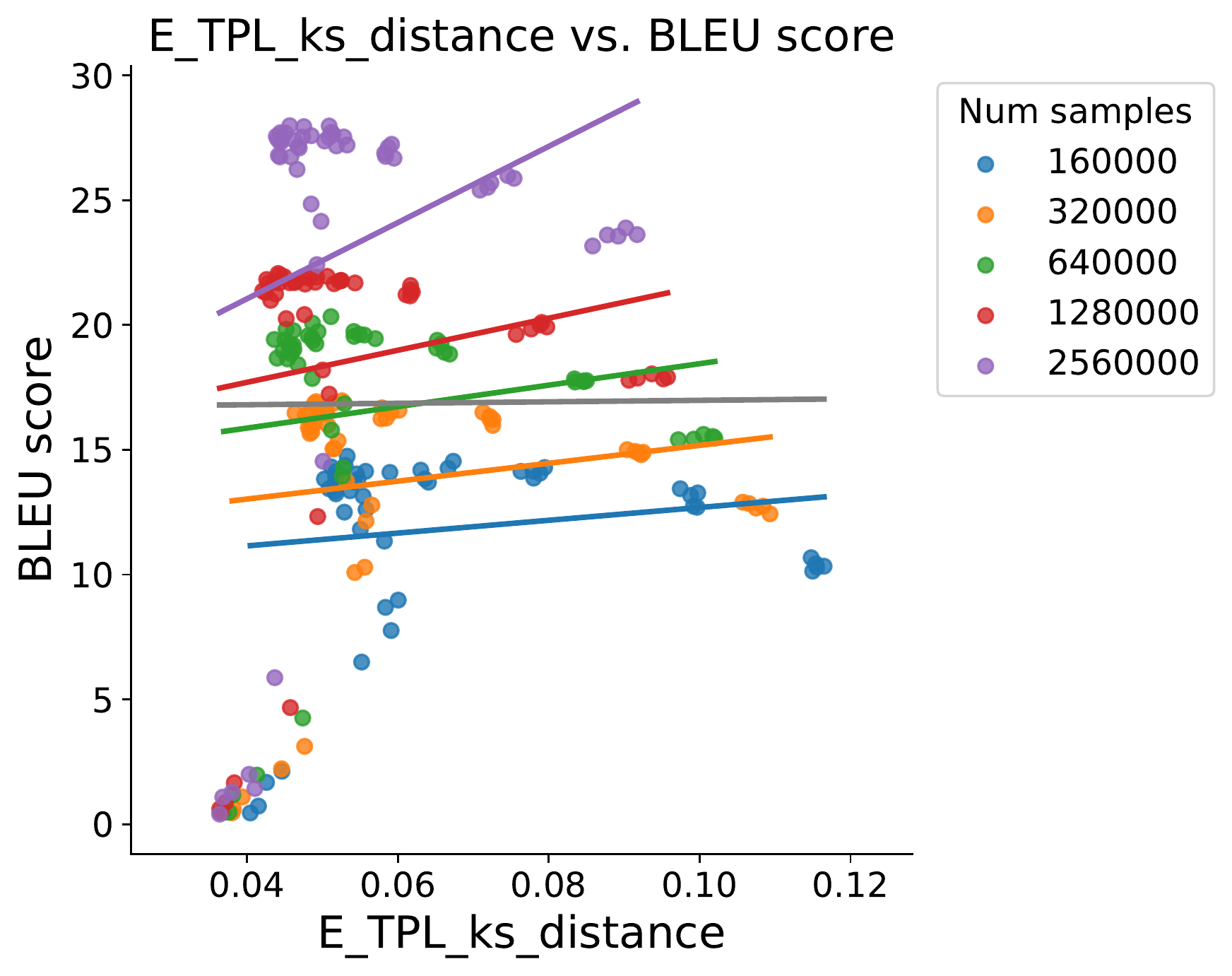}
    \includegraphics[width=0.32\textwidth]{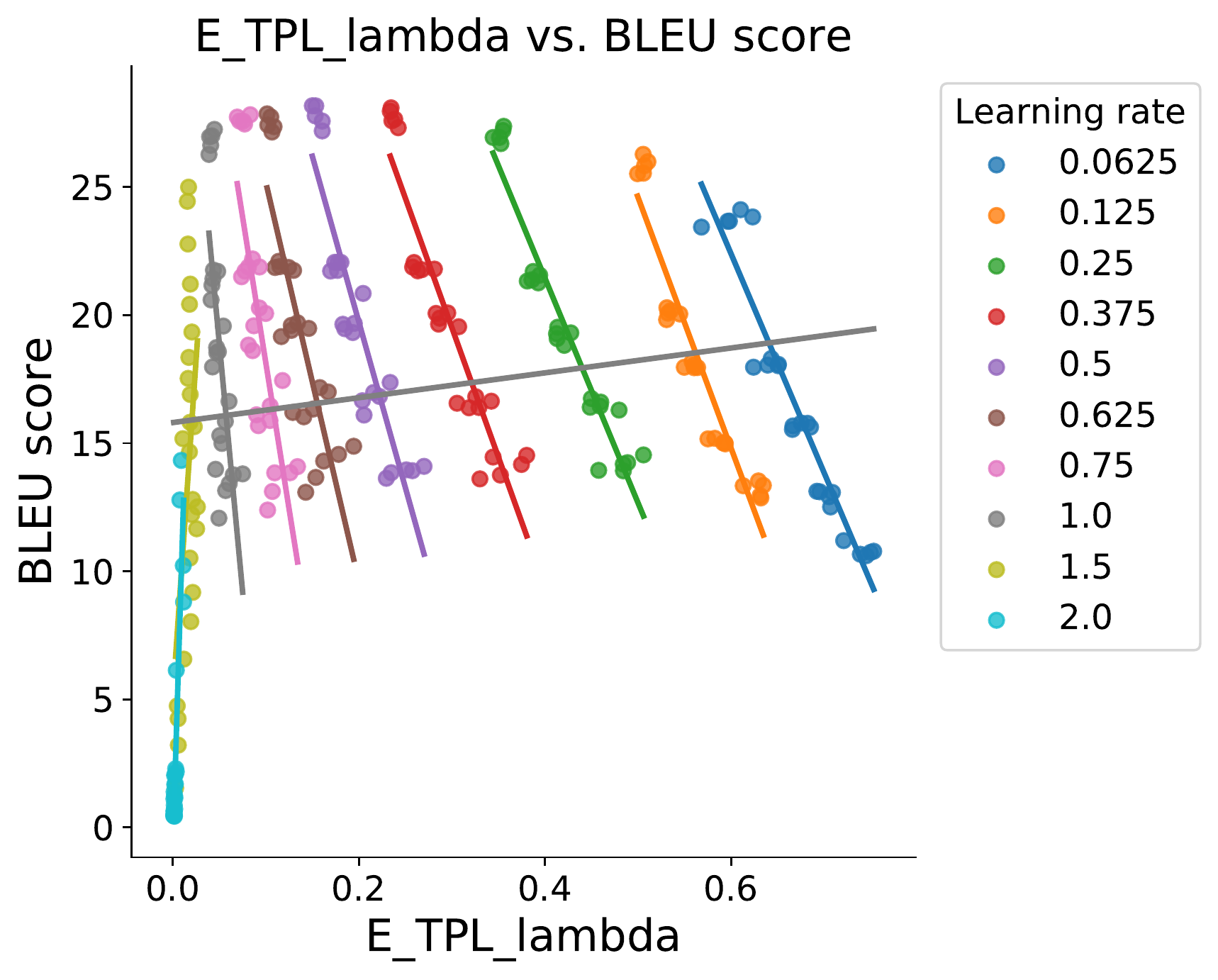}
    \includegraphics[width=0.32\textwidth]{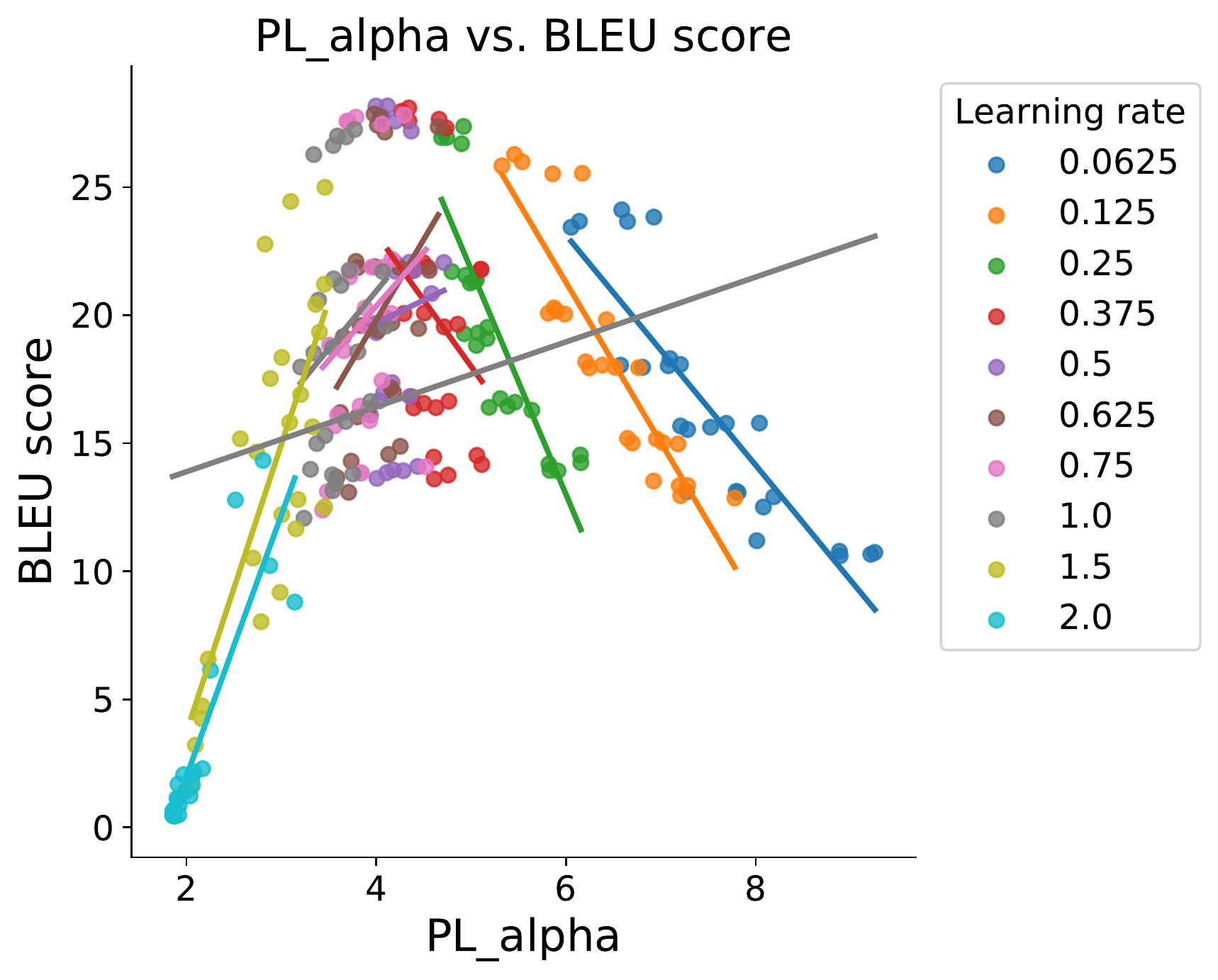}
    \includegraphics[width=0.32\textwidth]{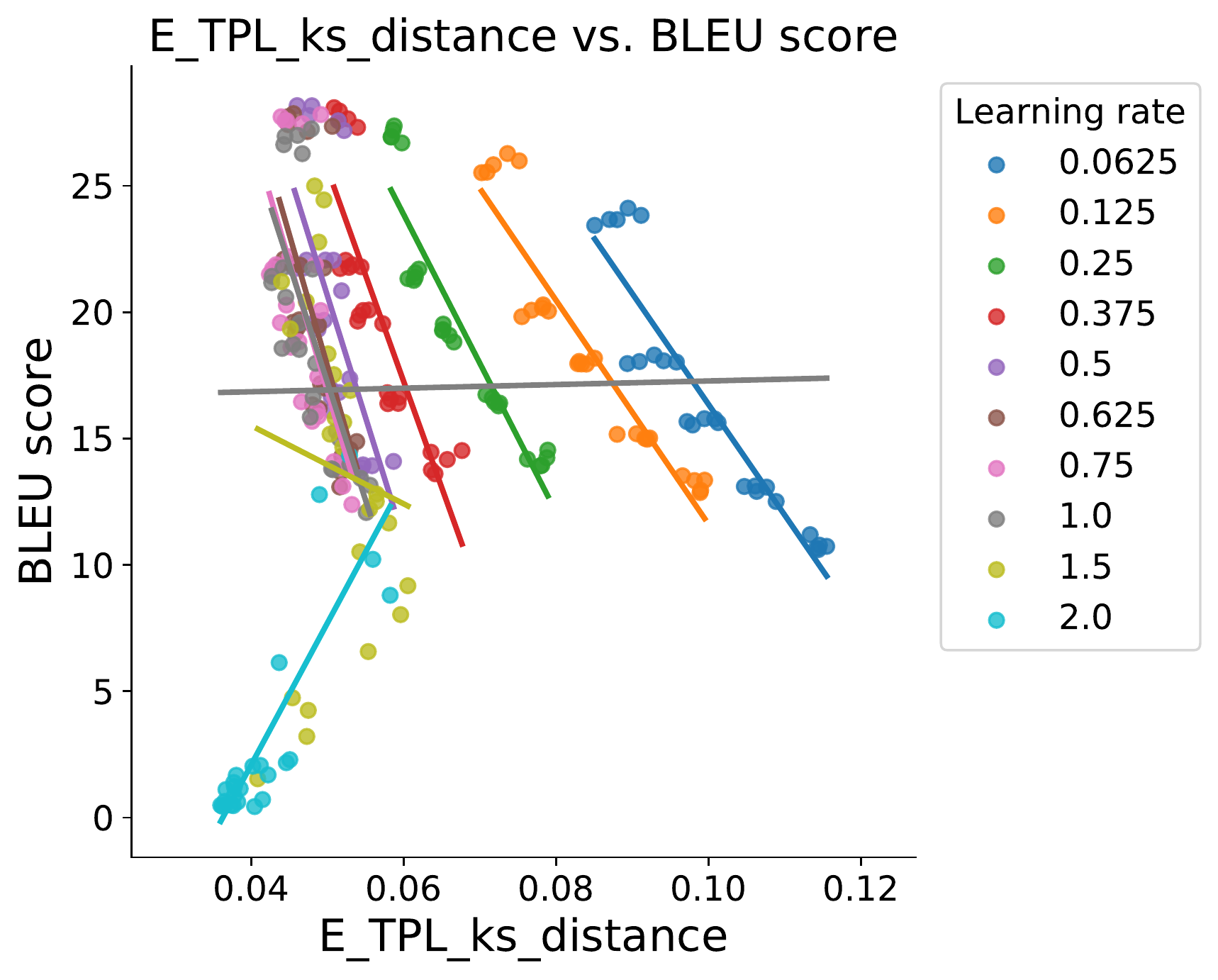}
    \caption{BLEU-score versus shape metrics with particularly large learning rates. {\bf(First row).} Trained models grouped by the learning rate. {\bf(Second row).} Trained models grouped by the number of samples. The BLEU scores and the evaluated shape metrics display Simpson's paradox when there are models trained with particularly large learning rates (1.5 and 2.0).}
    \label{fig:Shape_metrics_large_lr}
\end{figure*}

\noindent\textbf{Abnormal hyperparameters lead to the ``Simpson's paradox''.}
From Figure~\ref{fig:Shape_metrics}, we can see that the prediction of trends degrades for relatively large learning rates. 
For a fixed number of samples, when the learning rate becomes larger, the trends deviate from a perfectly linear fit.
Now, we include more models trained with particularly large learning rates, and we show the results in Figure~\ref{fig:Shape_metrics_large_lr}.
We see that the results potentially display a ``Simpson's paradox'' (similar results for a different corpus of models have been reported previously by~\citet{MM21a_simpsons_TR}), i.e., the overall correlation trends are opposite to the trends in individual groups.
In these figures, the regression lines are strongly influenced by the models trained with large learning rates and are biased towards the models with low BLEU scores. This phenomenon is known in the HT-SR literature \citep{MM21a_simpsons_TR,martin2020predicting_NatComm}, and one often has to avoid biasing the results with these poorly trained models \footnote{From a theoretical point of view, this phenomenon is caused by the change of the HT random matrix universality class at the point of \ALPHA = 2. For more details, see Table~1 of~\citet{martin2018implicit_JMLRversion}.}.

\noindent\textbf{BLEU scores are not significantly influenced by network depth.}
Unlike learning rates and number of samples, we find that the BLEU scores are almost identical when we vary the number of layers from 4 to 8.
In Figure~\ref{fig:group_by_lr} and~\ref{fig:group_by_sample}, we show \EXPONENT vs BLEU for models grouped by different learning rates and number of samples.
These two subfigures are repeated from the first column of Figure~\ref{fig:Shape_metrics_large_lr}.
From these two figures, we see that the BLEU scores vary significantly when these two hyperparameters are varied.
In Figure~\ref{fig:group_by_depth}, we show the same set of models color-coded by the network depth.
We can see that the BLEU scores almost remain identical.
This is because, from Figure~\ref{fig:group_by_lr} and~\ref{fig:group_by_sample}, we see that these models are roughly divided into ``vertical'' groups when the learning rate is varied, and they are roughly divided into ``horizontal'' groups when the number of samples is varied.
Therefore, each small ``cluster'' in Figure~\ref{fig:group_by_depth} corresponds to a group of models trained with the same learning rate and the number of samples but different depths, and these clusters show that the BLEU scores almost remain fixed when the depth is varied.
This phenomenon suggests that the rank correlations calculated for varying depths may not be informative.

\begin{figure}[ht!]
{\small\textbf{Correlations with model quality when different hyperparameters are varied.}}
    \centering
    \begin{subfigure}{0.49\textwidth}
         \includegraphics[width=0.95\textwidth]{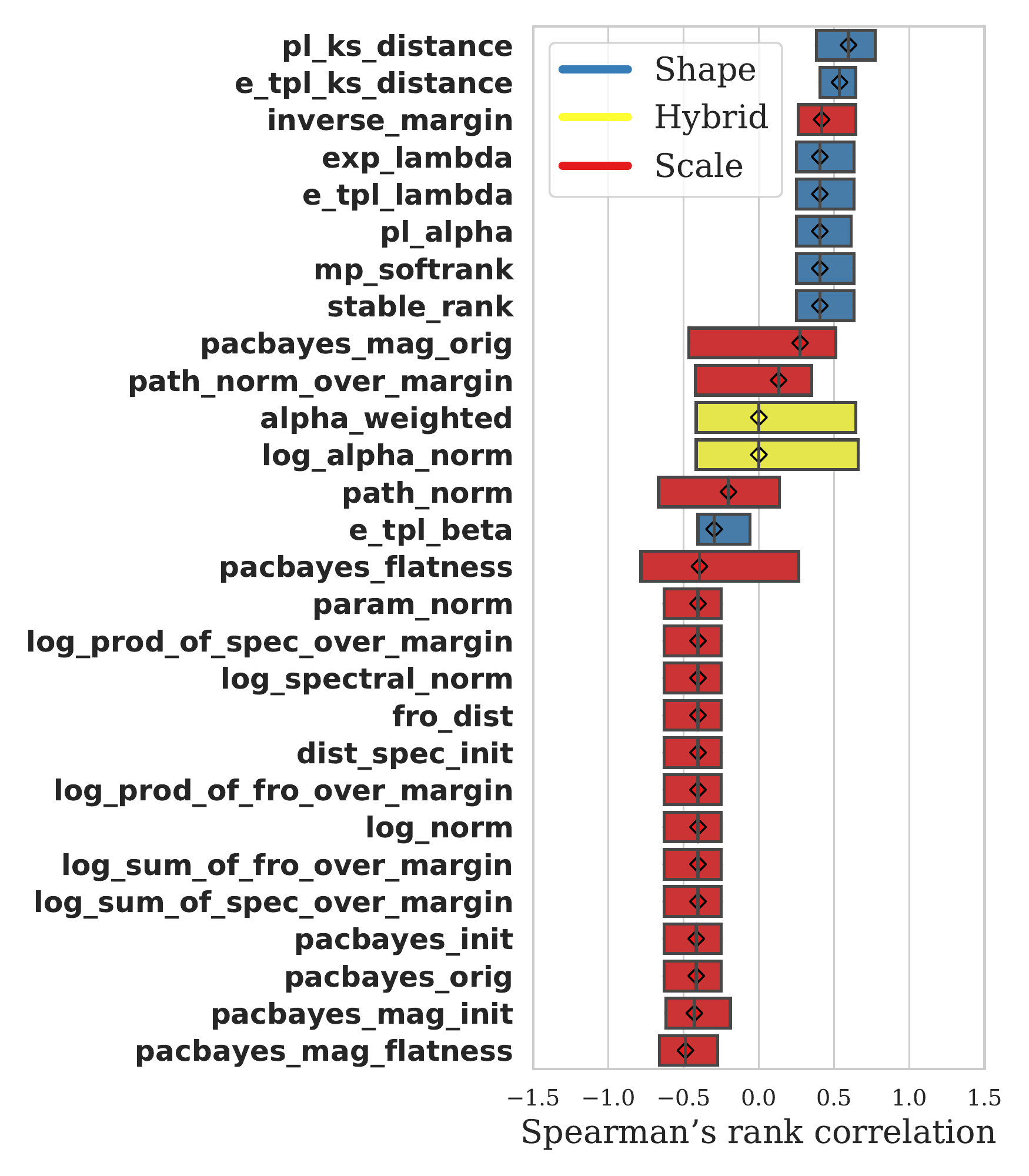}
        \caption{Varying learning rate.}
        \label{fig:rank_correlations_varying_lr}
    \end{subfigure}\hfill
    \begin{subfigure}{0.49\textwidth}
        \includegraphics[width=0.95\textwidth]{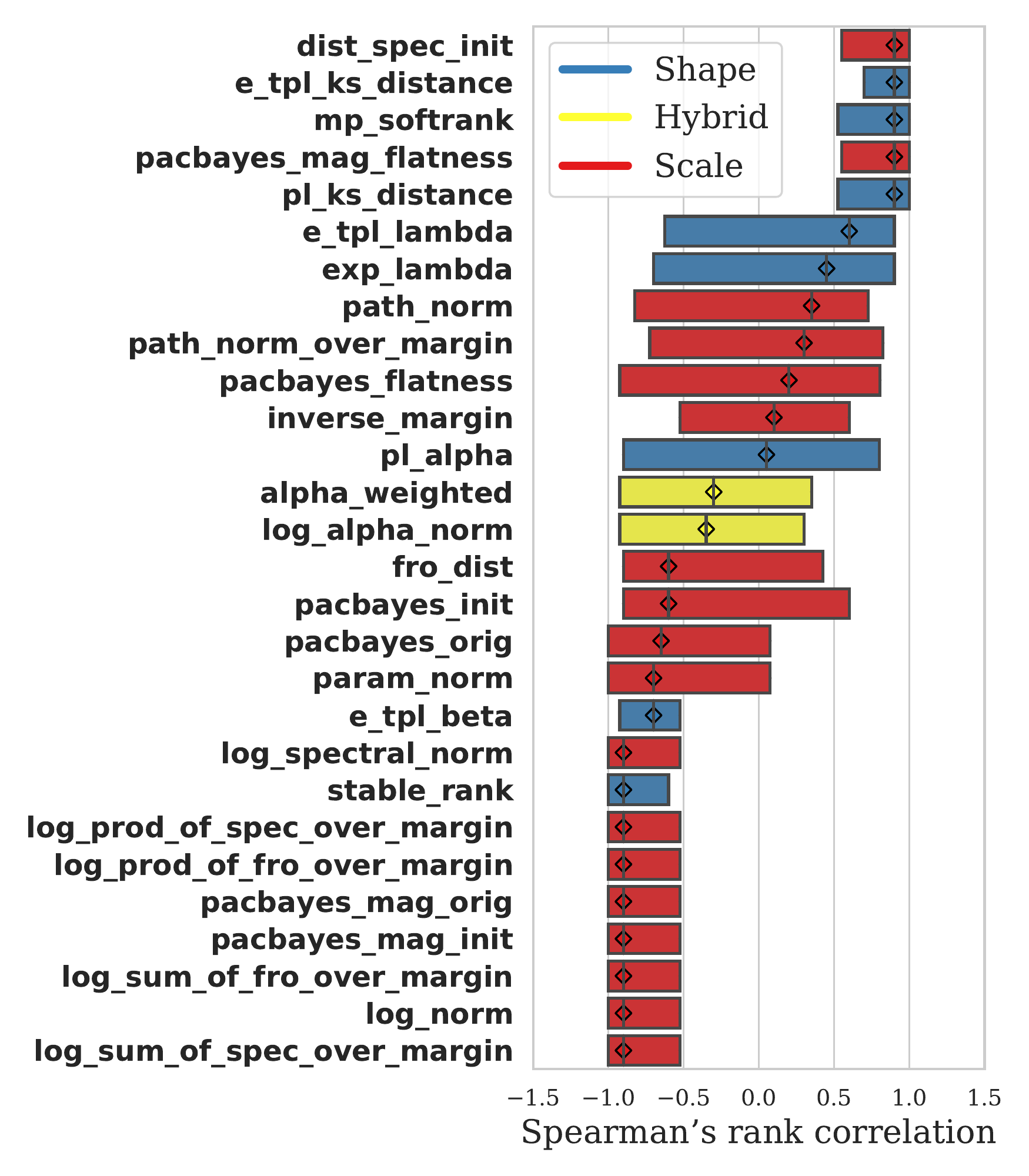}
        \caption{Varying network width.}
        \label{fig:rank_correlations_varying_width}
    \end{subfigure}
    \begin{subfigure}{0.49\textwidth}
        \includegraphics[width=0.95\textwidth]{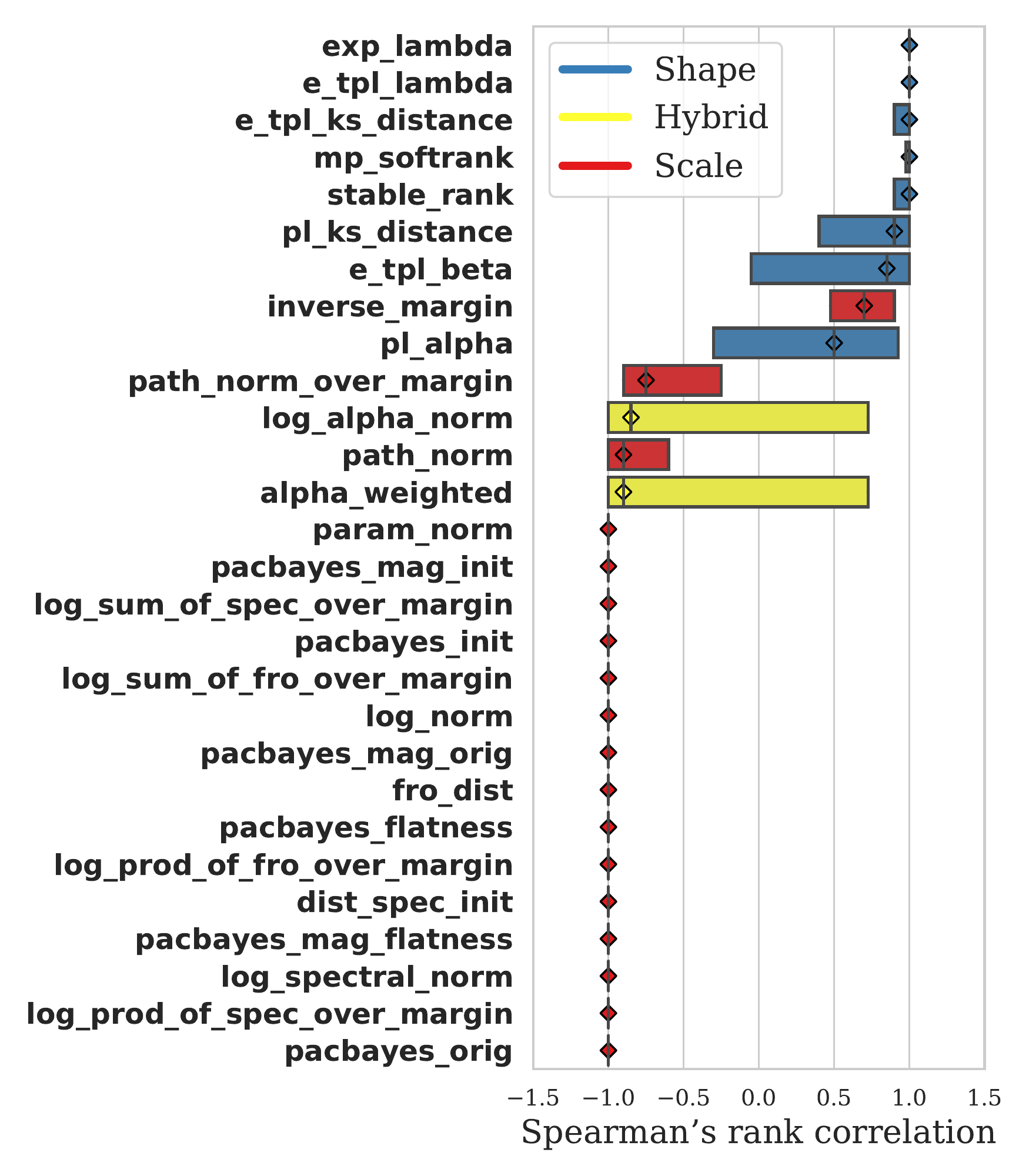}
        \caption{Varying number of training samples. In this subplot, metrics from \citet{jiang2019fantastic} are \emph{not} normalized by the number of training samples.}
        \label{fig:rank_correlations_varying_sample}
    \end{subfigure}\hfill
    \begin{subfigure}{0.49\textwidth}
        \includegraphics[width=0.95\textwidth]{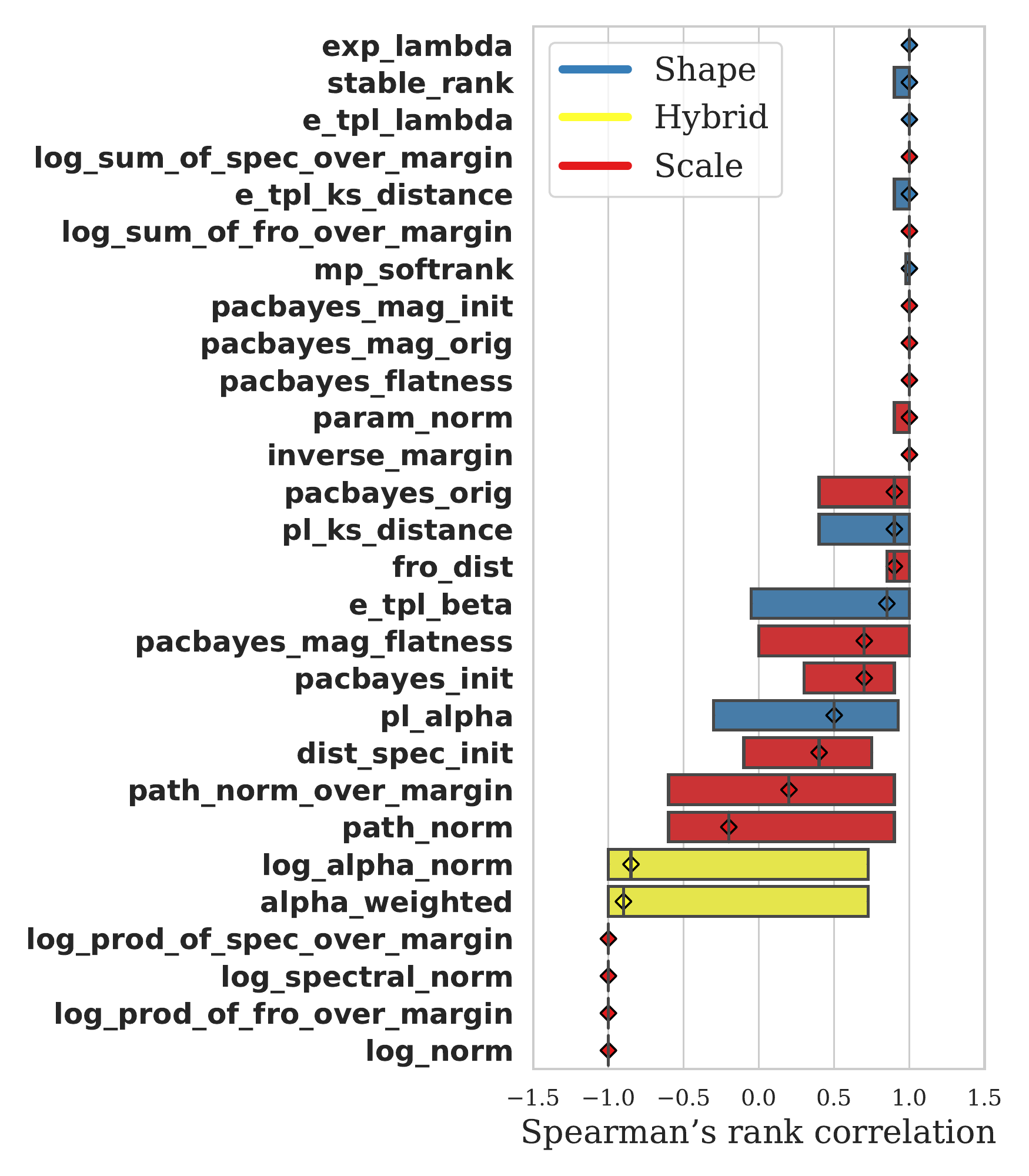}
        \caption{Varying number of training samples. In this subplot, metrics from \citet{jiang2019fantastic} are normalized by the number of training samples.}
        \label{fig:rank_correlations_varying_sample_normalized}
    \end{subfigure}
    \caption{Comparing multiple generalization metrics in terms of the rank correlations with the test BLEU score when multiple hyperparameters are varied. Metrics are ranked by the median rank correlations.
    }
    \label{fig:rank_correlations_all_hyperparameters}
\end{figure}

\begin{figure}[ht!]
{\small\textbf{Correlations with generalization gap when different hyperparameters are varied.}}
    \centering
    \begin{subfigure}{0.49\textwidth}
         \includegraphics[width=0.95\textwidth]{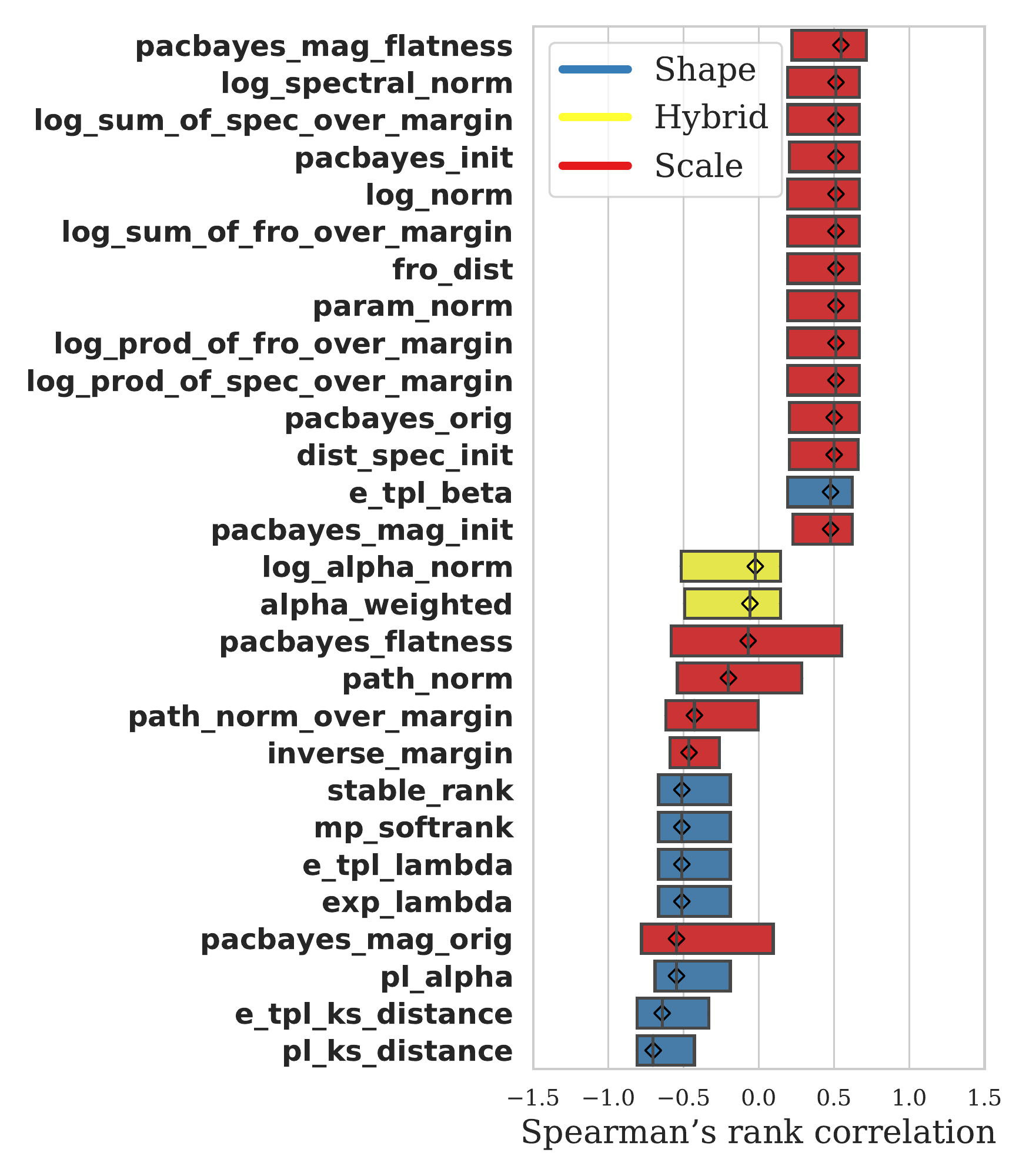}
        \caption{Varying learning rate.}
        \label{fig:rank_correlations_varying_lr_generalization_gap}
    \end{subfigure}\hfill
    \begin{subfigure}{0.49\textwidth}
        \includegraphics[width=0.95\textwidth]{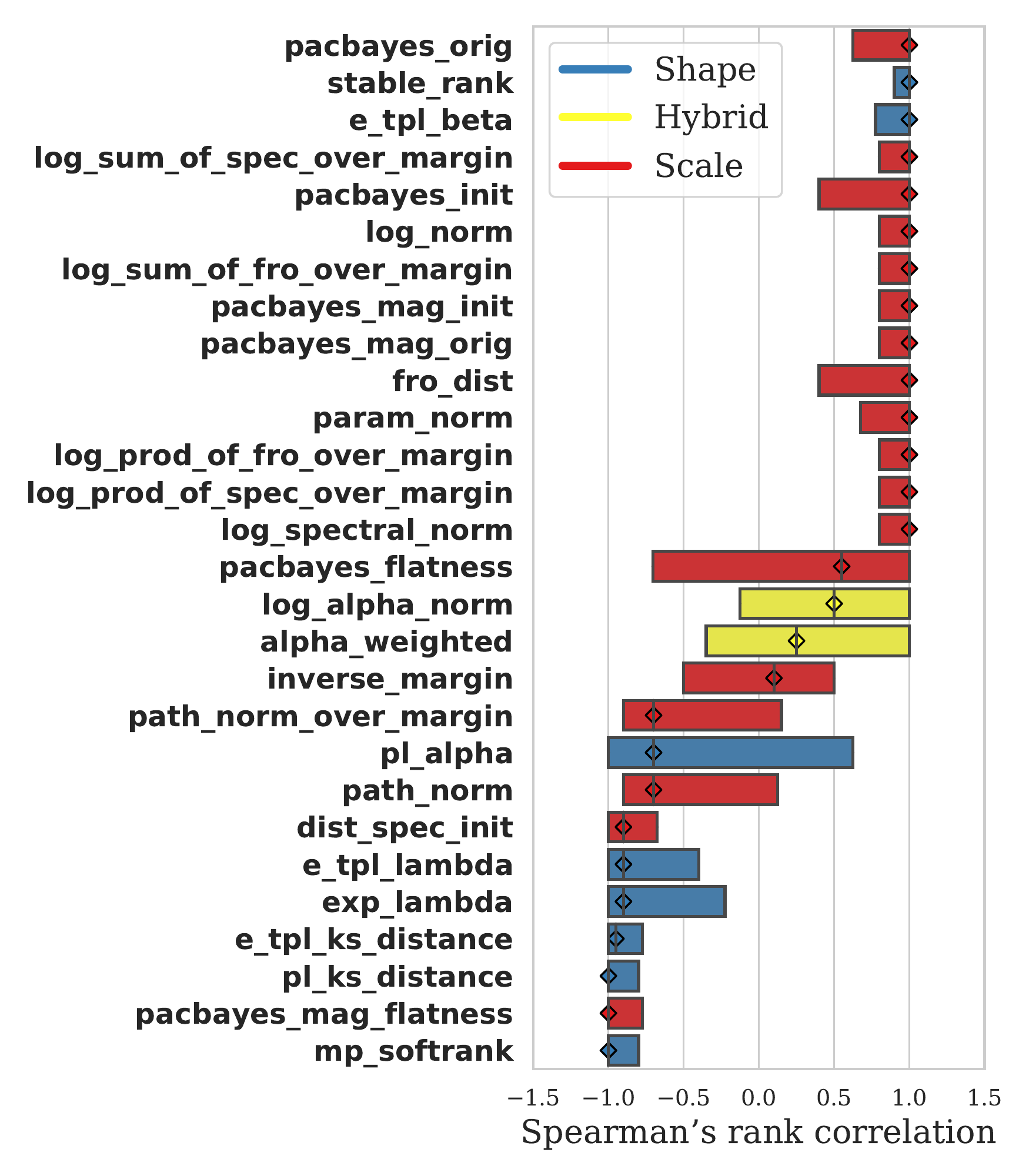}
        \caption{Varying network width.}
        \label{fig:rank_correlations_varying_width_generalization_gap}
    \end{subfigure}
    \begin{subfigure}{0.49\textwidth}
        \includegraphics[width=0.95\textwidth]{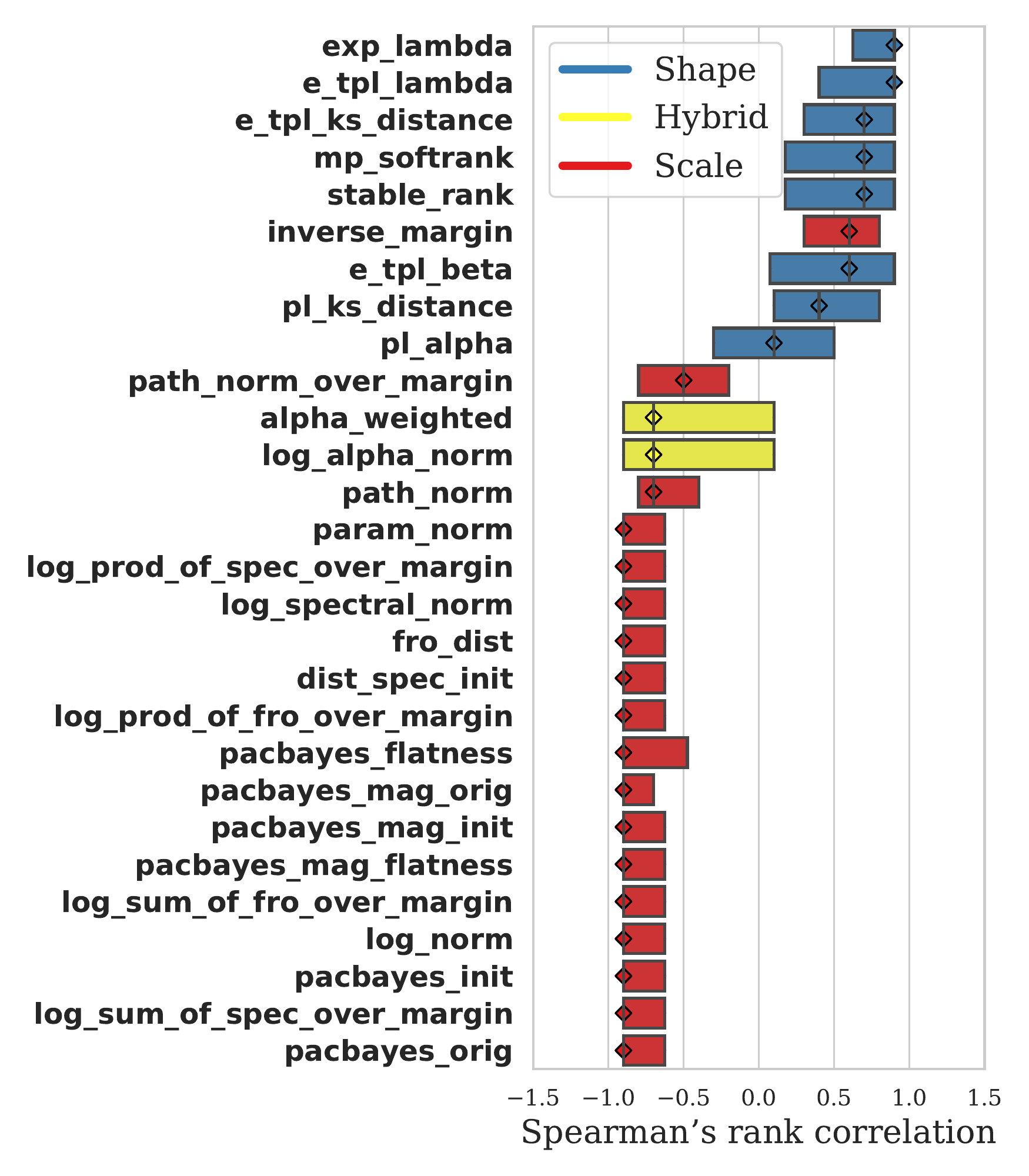}
        \caption{Varying number of training samples. In this subplot, metrics from \citet{jiang2019fantastic} are \emph{not} normalized by the number of training samples.}
        \label{fig:rank_correlations_varying_sample_generalization_gap}
    \end{subfigure}\hfill
    \begin{subfigure}{0.49\textwidth}
        \includegraphics[width=0.95\textwidth]{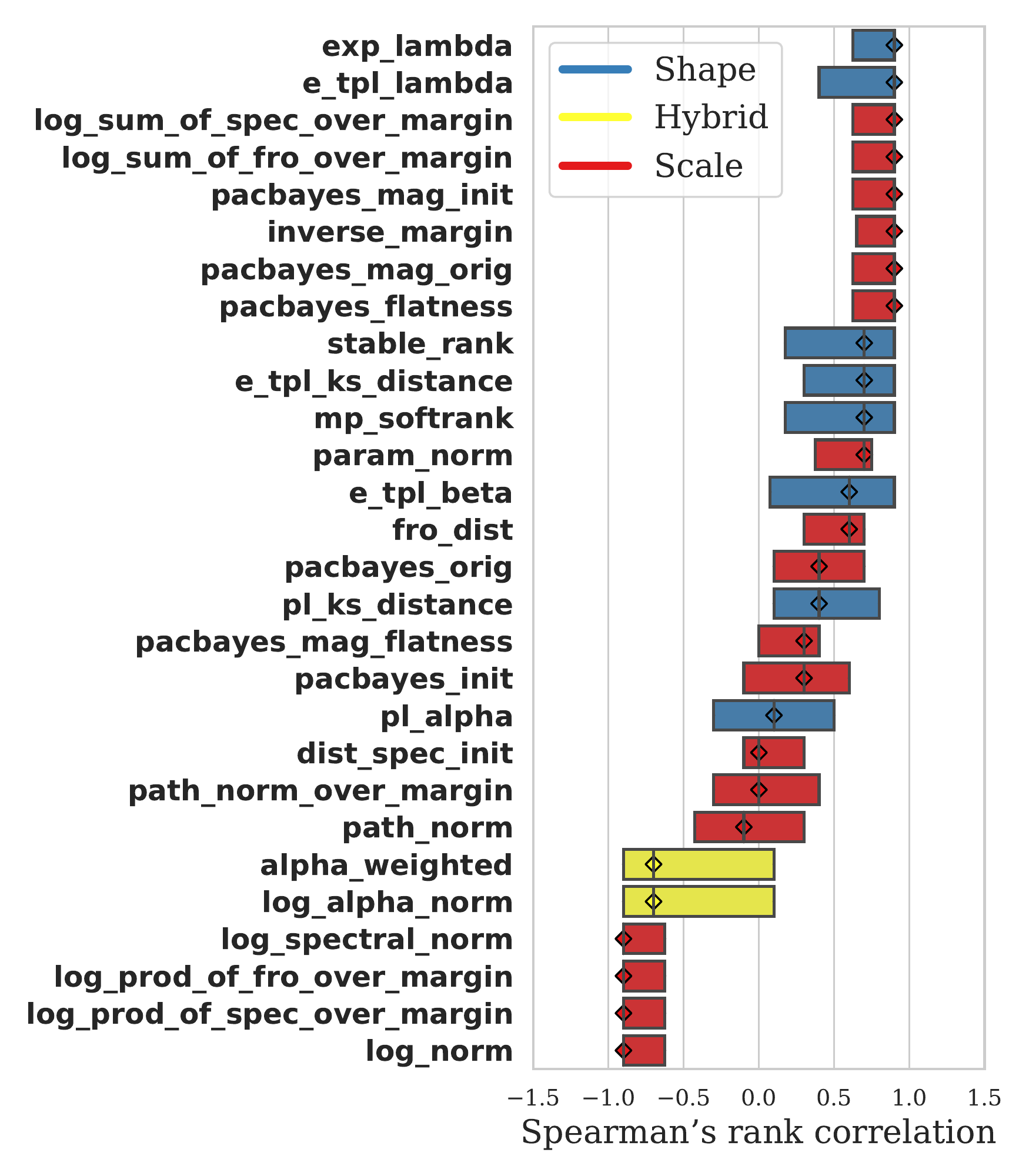}
        \caption{Varying number of training samples. In this subplot, metrics from \citet{jiang2019fantastic} are normalized by the number of training samples.}
        \label{fig:rank_correlations_varying_sample_normalized_generalization_gap}
    \end{subfigure}
    \caption{Comparing multiple generalization metrics in terms of the rank correlations with the \emph{generalization gap} when multiple hyperparameters are varied. Metrics are ranked by the median rank correlations.}
    \label{fig:rank_correlations_all_hyperparameters_generalization_gap}
\end{figure}

\noindent\textbf{Rank correlations.} 
To systematically evaluate the various metrics considered in this paper, we study the rank correlation
between these metrics and the BLEU score.
For Task three, we consider each one-dimensional slice of the hyperparameter space $\mathbf{\Theta} = \{(\theta_1, \ldots, \theta_K): \theta_1\in \mathbf{\Theta}_1, \ldots, \theta_K\in \mathbf{\Theta}_K\}$, i.e., slices of the form \[\{(\theta_1, \ldots, \theta_K): \theta_i \in \Theta_i \text{ while other parameters $\theta_j, j\neq i$ are fixed}\},\] 
and we calculate the rank correlation using the models in each such slice.
Then, we aggregate the rank correlations from all the one-dimensional slices and plot the distributions of the rank correlations.
See Figure~\ref{fig:rank_correlations_all_hyperparameters}.
As we have shown in Figure~\ref{fig:group_by_depth}, the rank correlations calculated with varying network depths may not be informative due to the insignificant change in the BLEU score.
Therefore, we focus on the other three hyperparameters, namely learning rate, network width and number of samples.
Also, similar to Task two, we provide the rank correlation results on both the test BLEU scores and the generalization gap.
Results on the generalization gap are shown in Figure~\ref{fig:rank_correlations_all_hyperparameters_generalization_gap}.

Before we analyze the results of Figure ~\ref{fig:rank_correlations_all_hyperparameters} and~\ref{fig:rank_correlations_all_hyperparameters_generalization_gap}, we discuss a subtle issue in calculating the generalization metrics. We note that, in \citet{jiang2019fantastic}, generalization metrics are ``normalized'' by dividing by the (square root of the) number of training samples, in correspondence with how they appear in uniform generalization bounds.
However, normalizing the generalization metrics from \citet{jiang2019fantastic} by the number of samples unavoidably complicates the correlation when varying the number of samples. 
This makes the comparison between scale metrics from \citet{jiang2019fantastic} and the shape metrics challenging, as there is no natural way to normalize the shape metrics with respect to the number of training samples.
Therefore, in Figure~\ref{fig:rank_correlations_all_hyperparameters}, we include the results for both with and without dividing the generalization metrics from \citet{jiang2019fantastic} by the (square root of the) number of samples.

Here are our observations from Figure~\ref{fig:rank_correlations_all_hyperparameters} and \ref{fig:rank_correlations_all_hyperparameters_generalization_gap}.
\begin{itemize}
    \item From Figure~\ref{fig:rank_correlations_varying_lr} and \ref{fig:rank_correlations_varying_sample}, shape metrics perform particularly well when varying the learning rate and the number of samples. 
    Specifically, in Figure~\ref{fig:rank_correlations_varying_sample}, several shape metrics achieve perfect rank correlations, which are close to 1.
    From Figure~\ref{fig:rank_correlations_varying_width}, shape metrics also perform well for varying network widths except for \STABLERANK and \ETPLBETA \footnote{The insufficiency of \STABLERANK is caused by the influence of the matrix size when the model width is varied, i.e., wider models tend to have a larger \STABLERANK simply because of the increased matrix size instead of the model quality.
    The insufficiency of \ETPLBETA is likely due to the \emph{fix-finger method} in E-TPL fittings, which fixes $x_{\min}$ as the peak of the ESD without searching for the optimal value.
    However, optimal E-TPL fitting requires simultaneous searching of $x_{\min}$, \ETPLBETA and \EXPONENT, which is computationally demanding. Thus, further investigation is necessary for achieving a balance between the quality of fitting and the computational cost.
    }.
    \item From Figure~\ref{fig:rank_correlations_varying_sample_normalized}, normalizing the scale metrics from \citet{jiang2019fantastic} by the number of samples significantly improves their correlational predictions.
    However, shape metrics can achieve a similar performance without the help of normalizing the number of samples.
    \item By comparing Figure~\ref{fig:rank_correlations_all_hyperparameters} and \ref{fig:rank_correlations_all_hyperparameters_generalization_gap}, one can see that the scale metrics are much better correlated with the generalization gap than the test BLEU scores. 
    For Figure~\ref{fig:rank_correlations_varying_lr_generalization_gap} and~\ref{fig:rank_correlations_varying_width_generalization_gap}, this conclusion is obvious from the plots.
    For Figure~\ref{fig:rank_correlations_varying_sample_generalization_gap} and \ref{fig:rank_correlations_varying_sample_normalized_generalization_gap}, the scale metrics need to be divided by (the square root of) the number of samples to achieve a good correlation when the number of samples is varied.
\end{itemize}

%% file: sections/Experiments_Huggingface.tex
\subsection{Selecting Huggingface Transformers}
\label{sec:Huggingface}

\ifisKDD
\else
\begin{figure}
    \centering
    \begin{subfigure}{0.6\textwidth}
    \centering
        \includegraphics[width=\textwidth]{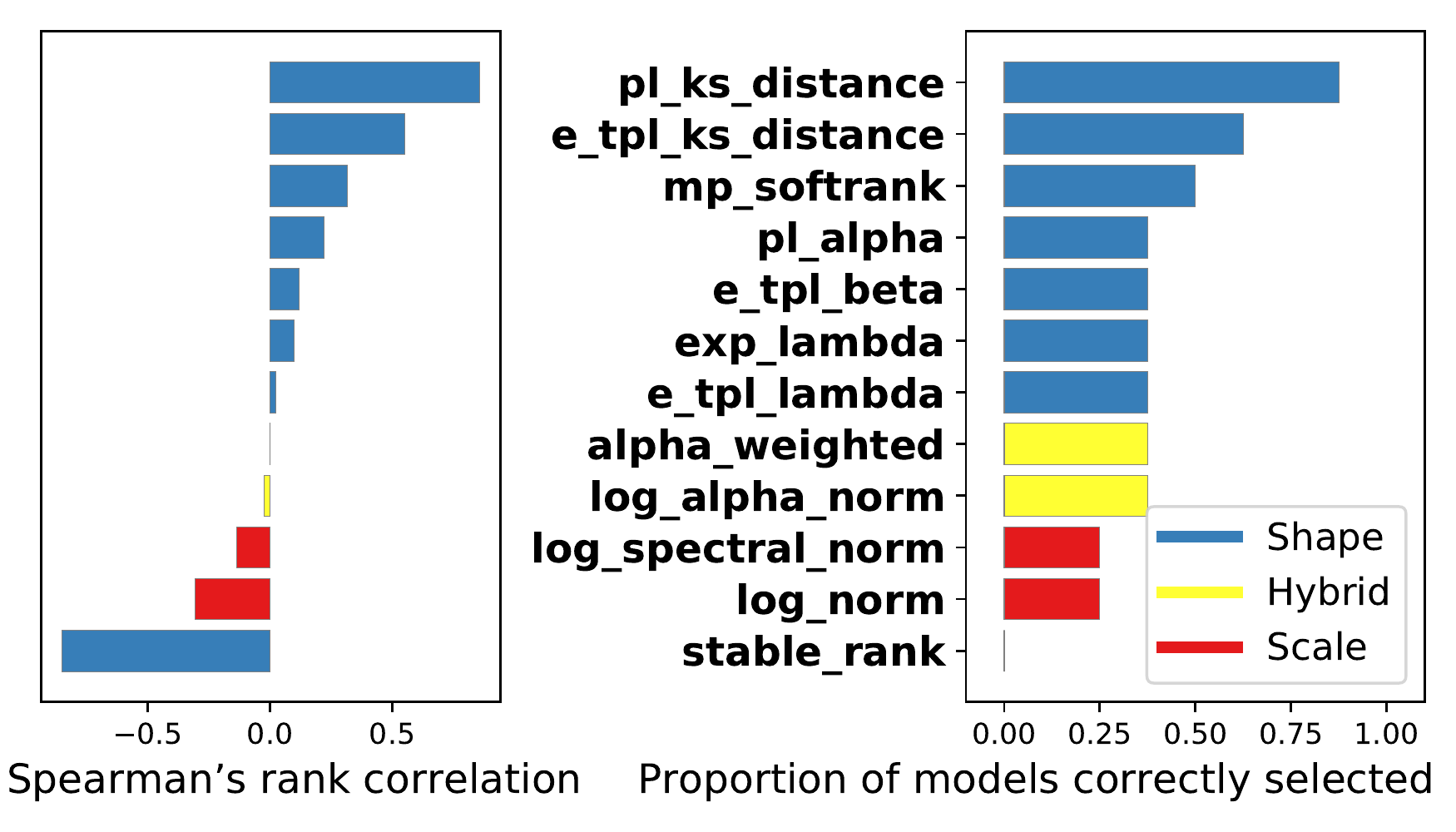}
        \caption{Model selection on Huggingface Transformers. Metrics on the left and right are aligned.
        \label{fig:Huggingface_ave_rank_correlation}}
    \end{subfigure}\\
    \begin{subfigure}{0.45\textwidth}
        \includegraphics[width=\textwidth]{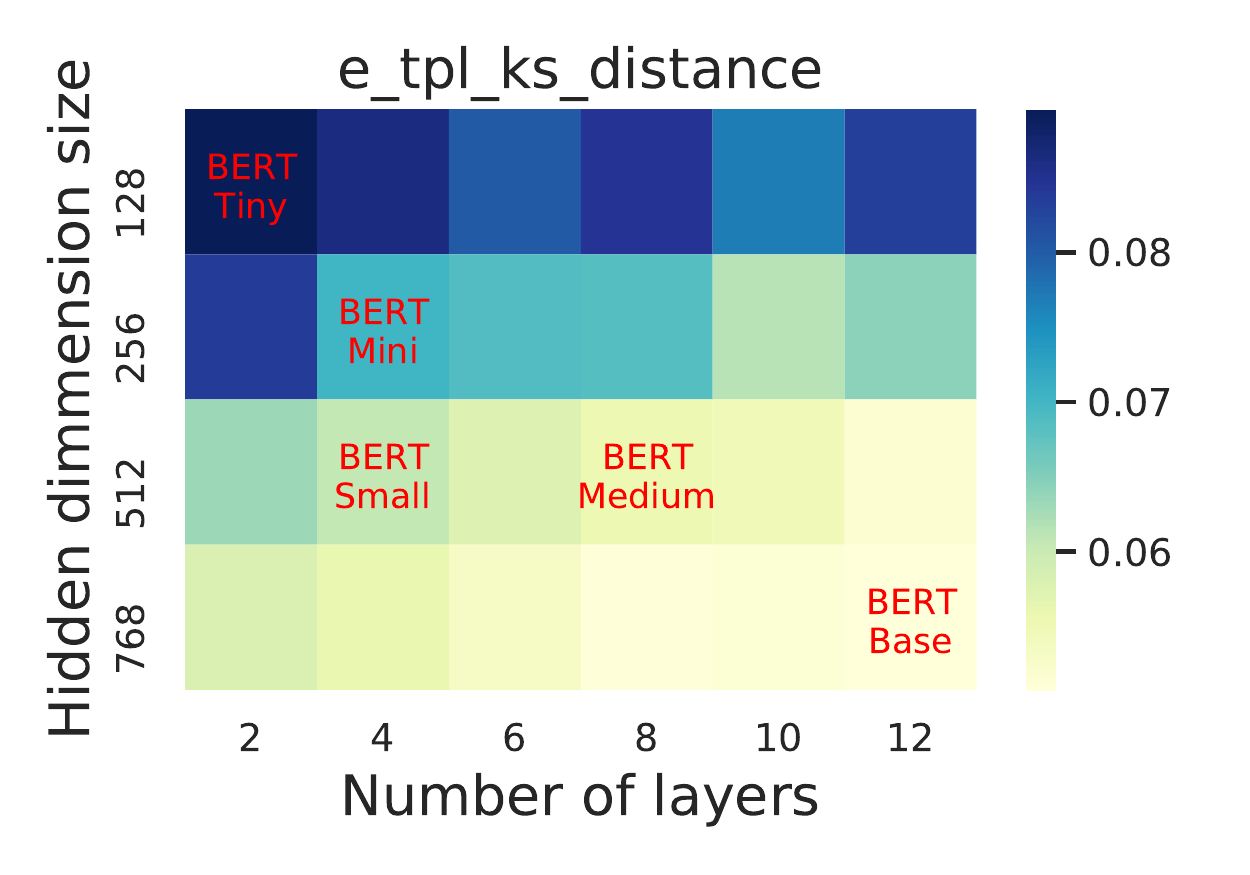}
        \caption{\ETPLKSDISTANCE evaluated on BERT models of different size. \label{fig:BERT_tpl_ks}}
    \end{subfigure}\hfill
    \begin{subfigure}{0.45\textwidth}
        \includegraphics[width=\textwidth]{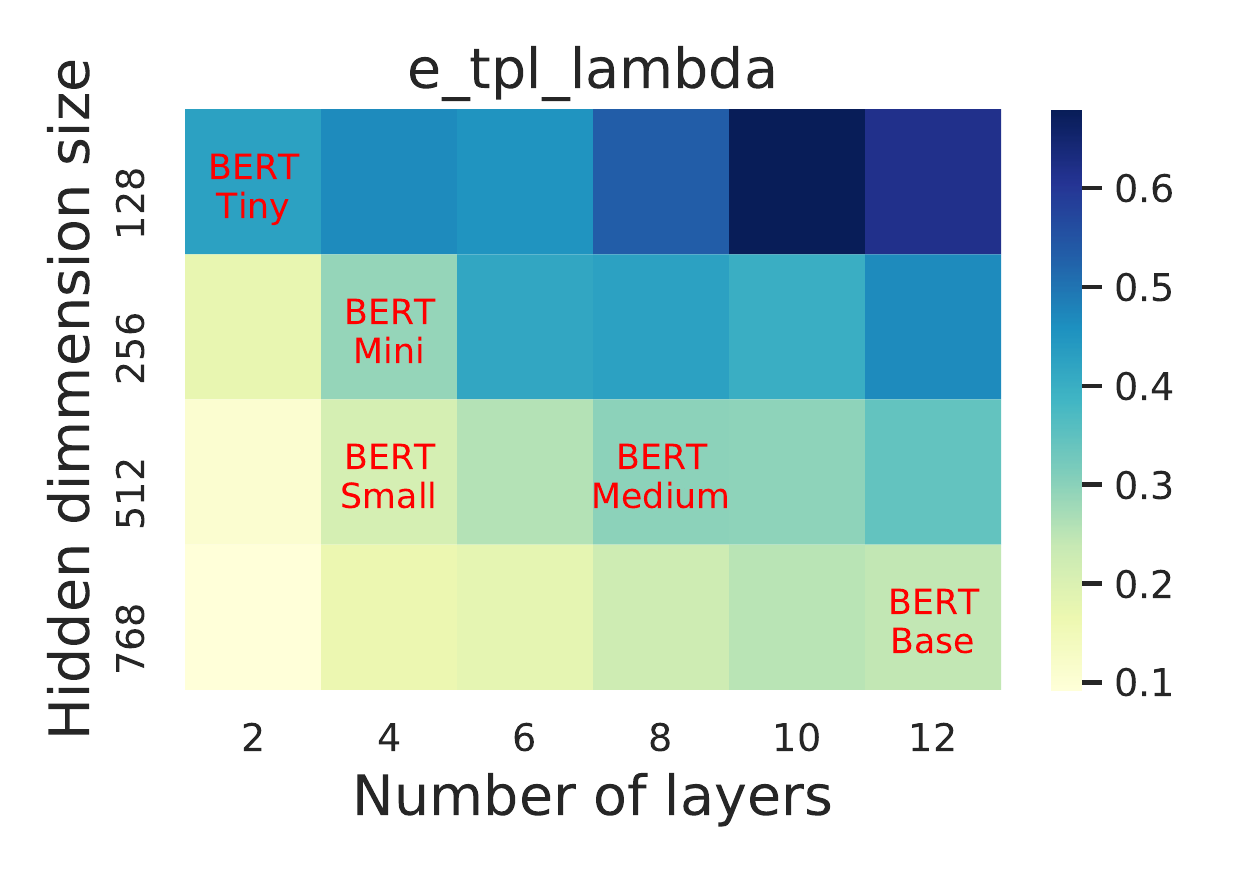}
        \caption{\EXPONENT evaluated on BERT models of different size. \label{fig:BERT_tpl_lambda}}
    \end{subfigure}
    \caption{Generalization metrics evaluated on pretrained Transformers. (a) Model selection results on eight Huggingface Transformer model series: BERT, GPT2, ALBERTv1, ALBERTv2, T5, DialoGPT, FlauBERT, Funnel Transformer. 
    Left shows the rank correlation averaged over different Transformers. Right shows the proportion of the best Transformers correctly selected using different metrics.
    Shape metrics outperform scale metric only except \STABLERANK which is strongly influenced by the matrix size.
    (b and c) Evaluating two metrics on the ``Smaller BERT'' series. While \ETPLKSDISTANCE predicts the correct trends, \EXPONENT shows the reversed trends with depth.
    } 
    \label{fig:Huggingface}
\end{figure}
\fi

Finally, we evaluate generalization metrics on the model selection task of pretrained Transformers. This section presents the first systematic study of applying generalization metrics to the model selection of Transformers without any training/validation/testing data. In our study, eight series of models downloaded from Huggingface~\citep{wolf-etal-2020-transformers} are considered---see Table~\ref{tab:pretrained_transformers}.
We also include 24 BERT models from the ``Smaller BERT'' series~\citep{turc2019well} produced from a ``pretrained distillation'' pipeline that combines masked language modeling pretraining \citep{devlin2018bert} and knowledge distillation from a single BERT teacher model.
In total, there are 51 pretrained Transformers.

\begin{table}
{\small
    \centering
    \begin{tabular}{c|c}
    \hline
       \gb Model series & Models \\
    \hline
    \hline
         BERT \citep{devlin2018bert} & BERT \{Tiny, Mini, Small, Base, Large\}\\
         \gb Smaller BERT \citep{turc2019well} & 24 smaller BERT models (English, uncased, \\ 
         \gb & trained with WordPiece masking) \\
         GPT2 \citep{radford2019language} & GPT2 \{Original, Medium, Large, XL\}\\
        \gb ALBERTv1 \citep{lan2019albert} & ALBERT-v1 \{base, large, xlarge, xxlarge\}\\
        ALBERTv2 \citep{lan2019albert} & ALBERT-v2 \{base, large, xlarge, xxlarge\}\\
        \gb T5 \citep{raffel2020exploring} & T5 \{small, base, large\} \\
        DialoGPT \citep{zhang2019dialogpt} & DialoGPT \{small, medium, large\} \\
        \gb FlauBERT \citep{le2019flaubert} & FlauBERT \{small, base, large\} \\
        Funnel Transformer \citep{dai2020funnel} & FunnelModel \{small, medium, \\
         & intermediate, large, xlarge\} \\
    \hline
    \end{tabular}
    \vspace{.2cm}
    \caption{Pretrained Transformers considered in this paper. \label{tab:pretrained_transformers}\vspace{-.2cm}}
}
\end{table}

We report rank-correlations averaged over these 8 model series in Figure~\ref{fig:Huggingface_ave_rank_correlation} (left subplot), i.e., larger/deeper models should have smaller generalization metric values. Again, we find that the shape metrics outperform scale metrics (except for \STABLERANK, which is strongly influenced by the size of the weight matrix).
The hybrid models achieve performance in-between the shape and scale metrics.
In Figure~\ref{fig:Huggingface_ave_rank_correlation} (right subplot), we compare different metrics in their ability to select the \emph{best model}. That is, we report for each metric the proportion that the best model is selected from one model series when this metric is used as the model selection criterion.
Note that the rankings of metrics on the two subplots in Figure~\ref{fig:Huggingface_ave_rank_correlation} are the same.

From Figure~\ref{fig:Huggingface_ave_rank_correlation}, we can see that, while the shape metrics perform better than scale metrics, none show a particularly strong rank correlation. To understand this, we examine the ``Smaller BERT'' series~\citep{turc2019well}, which contains a more fine-grained structure of different model sizes.
Specifically, these models are arranged in a 4-by-6 grid, where 6 represents \{2,4,6,8,10,12\} transformer layers and 4 means different hidden embedding sizes \{128,256,512,768\}.
From Figure~\ref{fig:BERT_tpl_ks}, we see that the \ETPLKSDISTANCE correctly predicts the trend that wider and deeper models perform better.
On the other hand, from Figure~\ref{fig:BERT_tpl_lambda}, \EXPONENT correctly predicts that wider models are better, but incorrectly predicts that shallower models are better (yet another form of Simpson's paradox in a data set of neural network model quality; see also \citet{MM21a_simpsons_TR}).
\ifisKDD\else
We note that the BERT series \{BERT-Tiny, BERT-Mini, BERT-Small, BERT-Medium, BERT-Base, BERT-Large\} overlaps with the 2D grid (as shown in Figure~\ref{fig:BERT_tpl_ks} and~\ref{fig:BERT_tpl_lambda}). Consequently, the rank correlations for the BERT series (which we include as one of the eight series in making Figure~\ref{fig:Huggingface_ave_rank_correlation}, is a ``noisy subsample'' of the results in Figure~\ref{fig:BERT_tpl_ks} and~\ref{fig:BERT_tpl_lambda}.
\fi

Another curious observation from Figure~\ref{fig:Huggingface_ave_rank_correlation} is that, for the pretrained transformers, PL metrics, such as \PLKSDISTANCE and \ALPHA, outperform E-TPL metrics, such as \ETPLKSDISTANCE, \EXPONENT, and \ETPLBETA.
This phenomenon may seem surprising as one may expect E-TPL fits to be more flexible than PL fits.
These pretrained models are likely trained with much larger datasets and over many more epochs than the models we have otherwise considered. Here, PLs appear to provide a more natural fit.
This is further evidence that HT-SR theory is particularly well-suited for evaluating the quality of relatively high-quality models.

%% file: sections/Conclusion.tex
\section{Conclusion}

Poor correlations between existing generalization metrics and test-time performance have been reported in prior work \citep{nagarajan2019uniform,jiang2019fantastic,dziugaite2020search}.
Rather than providing a ``lump sum'' to rank existing and novel generalization metrics (Figure~\ref{fig:test_acc_vs_generalization_gap}), we evaluated these metrics in several ways: 
quantifying correlations only on optimally-trained models (Figure~\ref{fig:Optimal_models_shape_metric}); examining the time-wise correlation during training (Figure~\ref{fig:generalization});
differentiating between the correlation with test accuracy versus generalization gap (Figure~\ref{fig:test_acc_vs_generalization_gap});
\ifisKDD
\else thoroughly investigating the rich correlational structures when different hyperparameters are varied (Figures~\ref{fig:Shape_metrics} to~\ref{fig:rank_correlations_all_hyperparameters_generalization_gap});
and evaluating these metrics on pretrained Transformer models where we do not have any control over the training process (Figure~\ref{fig:Huggingface}).
\fi
Our large-scale empirical analyses suggest that popular generalization metrics still exhibit excellent correlations with generalization gap on NLP tasks. However, metrics derived from HT-SR theory appear to be most valuable to large language model practitioners, allowing one to assess pretrained NLP models without requiring training or testing data. Due to their apparent utility and current niche status, we recommend further investigations into these metrics, in particular, to address some of their remaining weaknesses (e.g. for suboptimally-trained models). 

%% file: sections/Metrics.tex
\newpage
\onecolumn
\section{Generalization metrics}\label{sec:metrics}

In this section, we provide definitions and details on the various metrics considered in our analysis.
We begin with scale metrics, and then consider shape metrics obtained from the ESDs of the weight matrices.
Although our focus is on generalization metrics that do not need data to evaluate, we also define generalization metrics based on margin \citep{bartlett2017spectrally,pitas2017pac} and PAC-Bayesian bounds \citep{mcallester1999pac,neyshabur2017pac}.

\subsection{Notation and preliminaries}

{\bf General notation.}
As before, we consider a NN with $d$ layers and corresponding weight matrices $\Wb_1$, $\Wb_2$,..., $\Wb_d$. We use $\Wb$ to denote the collection of all the weights and denote the vector that consists of all the model weights as $\vect(\Wb)$.
The neural network (as a function) is denoted by $f_\Wb$, taking a single input sample $\xb$ and outputs a vector $f_\Wb(\xb)$.
The superscript ${}^\text{init}$ on a weight matrix, e.g. $\Wb_1^\text{init}$, denotes the initial weights from which the model is trained.
The notation $\mathbf{1}$ means an all-one vector, and $\Ib$ means the identity~matrix.

{\bf Norms and distances.}
We use different types of norms defined on vectors and matrices. $\|\cdot\|_2$ and $\|\cdot\|_1$ used on vectors respectively means the $\ell_2$ norm and the $\ell_1$ norm.
$\|\cdot\|_F$ and $\|\cdot\|_2$ used on matrices respectively denotes the Frobenius norm and the spectral norm (which is the induced $\ell_2$ norm).

\subsection{Scale metrics}

{\bf Norm-based and distance-based metrics.} In the following, we discuss multiple metrics obtained from the norms of the weights or the distance between the final weights and those at initialization. While some metrics are averaged over layers, and others are not,
this inconsistency is not in error. We follow definitions of metrics from several prior papers \emph{verbatim}. Results in the second task (comparing model performance across a single training run) are independent of these factors. However, to compare networks with different sizes, proper normalization is necessary.
Some metrics across the literature are also linearly dependent on others, and are therefore redundant for comparison. For example, $\LOGPRODOFSPEC$ and $\LOGSUMOFSPEC$ from \citet{jiang2019fantastic} overlap with $\LOGSPECTRALNORM$ from \citet{MM21a_simpsons_TR}, and $\LOGSUMOFFRO$ and $\LOGPRODOFFRO$ from \citet{jiang2019fantastic} overlap with $\LOGNORM$ from \citet{martin2018implicit_JMLRversion}. These metrics are not considered.

\begin{itemize}[noitemsep,topsep=0pt,leftmargin=*,after=,before=]
    \item (\PARAMNORM). The squared Frobenius norm summed over all weight matrices.
    \begin{equation}\label{eqn:PARAM_NORM}
        \mu_\text{\PARAMNORM} = \sumid \|\Wb_i\|_F^2.
    \end{equation}
    \item (\FRODIST). The distance between a weight matrix and its initilized value, calculated using the Frobenius norm and summed over all layers.
    \begin{equation}\label{eqn:FRO_DIST}
        \mu_\text{\FRODIST} = \sumid \|\Wb_i-\Wb_i^\text{init}\|_F^2.
    \end{equation}
    \item (\LOGNORM).
    \begin{equation}\label{eqn:LOG_NORM}
        \mu_\text{\LOGNORM} = \aveid \log \|\Wb_i\|_F^2.
    \end{equation}
    \item (\LOGSPECTRALNORM).
    \begin{equation}\label{eqn:LOG_SPECTRAL_NORM}
        \mu_\text{\LOGSPECTRALNORM} = \aveid \log \|\Wb_i\|_2^2.
    \end{equation}
    \item (\DISTSPECINIT).
    \begin{equation}\label{eqn:DIST_SPEC_INIT}
        \mu_\text{\DISTSPECINIT} = \sumid \|\Wb_i-\Wb_i^\text{init}\|_2^2.
    \end{equation}
    \item (\PATHNORM). The metric is introduced in \citet{neyshabur2015norm}. To calculate the metric, we square the parameters of the network, perform a forward pass on an all-ones input and then compute the sum of the network outputs.
    \begin{equation}\label{eqn:PATH_NORM}
        \mu_\text{\PATHNORM} = \left\| f_{\Wb^2}(\mathbf{1})\right\|_1.
    \end{equation}
\end{itemize}

{\bf Scale metrics that require more shape information from the ESDs.} The following metrics require more than just a single type of norm, instead involving a combination of a norm with other factors.

\begin{itemize}[noitemsep,topsep=0pt,leftmargin=*,after=,before=]
    \item (\MPSOFTRANK). This metric is introduced in \citet{martin2018implicit_JMLRversion}. To calculate this metric, we fit the MP distribution on the ESD, obtain the bulk max of the MP distribution and then divide by the maximum eigenvalue.
    \begin{equation}\label{eqn:MP_SOFTRANK}
        \mu_\text{\MPSOFTRANK} = \aveid \lambda_{i,\text{MP}}/\lambda_{i,\max}.
    \end{equation}
    \item (\STABLERANK). The metric is a norm-adjusted measure of the scale of the ESD.
    \begin{equation}\label{eqn:STABLE_RANK}
        \mu_\text{\STABLERANK} =  \aveid  \|\Wb_i\|_F^2 / \|\Wb_i\|_2^2
    \end{equation}
\end{itemize}

\subsection{Shape metrics}

{\bf Tail-exponent fitting.} The following metrics are derived from heavy or light-tailed fits to the ESD.

\begin{itemize}[noitemsep,topsep=0pt,leftmargin=*,after=,before=]
    \item (\ALPHA). The slope of the tail of the ESD, on a log-log scale. We use MLE from \citet{alstott2014powerlaw} to estimate \ALPHA. The distribution of eigenvalues is assumed to have the form of \eqref{eqn:ALPHA}.
    \item (\EXPONENT). The tail exponent of the E-TPL fit to the ESD. This is a novel generalization metric introduced in this work.
    \item (\EXPDISTEXPONENT). The tail exponent of the EXP fit to the ESD, under the assumption that the ESD follows an exponential distribution~\eqref{eqn:EXP_DIST_EXPONENT}. This is also a novel generalization metric introduced in this work.
    \item (\PLKSDISTANCE). The Kolmogorov-Smirnoff (KS) goodness-of-fit test statistic for the PL fit:
    \begin{equation}\label{eqn:KS_DISTANCE}
        \mu_\text{\KSDISTANCE} = \aveid \sup_x |F_i^*(x) - S_i(x)|,
    \end{equation}
    where $F_i^*(x)$ is the distribution of the estimated PL fit to the ESD, and $S_i(x)$ is the ESD itself.
    \item (\ETPLKSDISTANCE). The KS test statistic for the E-TPL fit, defined in the same way as \eqref{eqn:KS_DISTANCE}, except that $F_i^*(x)$ is the distribution of the estimated E-TPL fit to the ESD.
\end{itemize}

\subsection{Hybrid metrics}

The following metrics are scaled versions of \ALPHA, involving both shape information from \ALPHA and scale information from other weighted norms.
Let $\alpha_i$ denote the estimated PL coefficient of the ESD of the $i$-th weight matrix $\Wb_i$. Recall that $\lambda_{i,\max}$ is the largest eigenvalue of $\Wb_i$.
\begin{itemize}[noitemsep,topsep=0pt,leftmargin=*,after=,before=]
    \item (\ALPHAWEIGHTED). A scale-adjusted form of \ALPHA. This metric is denoted as $\hat{\alpha}$ in \citet{martin2018implicit_JMLRversion,martin2020predicting_NatComm,MM21a_simpsons_TR}.
    \begin{equation}\label{eqn:ALPHA_WEIGHTED}
        \mu_\text{\ALPHAWEIGHTED} = \aveid \alpha_i \log \lambda_{i,\max}.
    \end{equation}
    \item (\LOGALPHANORM). This metric is another scale-adjusted \ALPHA metric in the form of a Schatten norm.
    Recall that we let $\{\lambda_j\}_{j=1}^M$ denote the set of eigenvalues of the correlation matrix $\Xb_i=\Wb_i^\top \Wb_i$, where $\Wb_i$ is the $N$-by-$M$ weight matrix that satisfies $N\ge M$. Then, we can define the Schatten $p$-norm as
    \begin{equation}\label{eqn:LOG_ALPHA_NORM}
        \|\Xb_i\|_p = \left(\sum_{j=1}^M \lambda_j^p \right)^{\frac{1}{p}}.
    \end{equation}
    The metric \LOGALPHANORM is given by
    \begin{equation}
        \mu_\text{\LOGALPHANORM} = \aveid \log \|\Xb_i\|_{\alpha_i}^{\alpha_i}.
    \end{equation}
\end{itemize}

\subsection{Margin-based metrics}

Finally, we discuss generalization metrics derived from margins. Recalling that $f_\Wb$ denotes a neural network with weights $\Wb$, for a multi-class classification problem with sample-label pair $(\xb, y)$, we define the \emph{margin}~as
\begin{equation}
    \gamma(\xb,y,f_\Wb) = (f_\Wb(\xb))[y]-\max_{i\neq y}f_\Wb(\xb)_i.
\end{equation}
For machine translation, we consider the margin of each output token. We note that the number of classes, or the number of possible tokens, is often particularly large (in the order of thousands) for machine translation. Note that margins can be defined in any layer \citep{elsayed2018large,yang2020boundary,wei2019improved}.
Following \citet{jiang2019fantastic}, we consider output margins only, and use the 10$^\text{th}$ percentile of the margin distribution calculated from the entire training set as a robust surrogate for the minimum margin.
Using the margin $\gamma$ defined as the 10$^\text{th}$ percentile, we define several generalization metrics.
\begin{itemize}[noitemsep,topsep=0pt,leftmargin=*,after=,before=]
    \item (\INVERSEMARGIN).
    \begin{equation}\label{eqn:INVERSE_MARGIN}
        \mu_\text{\INVERSEMARGIN} = \frac{1}{\gamma^2}.
    \end{equation}
    \item (\LOGPRODOFSPECOVERMARGIN).
    \begin{equation}\label{eqn:LOG_PROD_OF_SPEC_OVER_MARGIN}
        \begin{split}
            \mu_\text{\LOGPRODOFSPECOVERMARGIN}
            = \log \frac{\prod_{i=1}^d \|\Wb_i\|_2^2}{\gamma^2} 
            = \mu_\text{\LOGPRODOFSPEC} - 2 \log \gamma.
        \end{split}
    \end{equation}
    Note that \LOGPRODOFSPEC is not used in this paper due to overlap with \LOGSPECTRALNORM.
    \item (\LOGSUMOFSPECOVERMARGIN).
    \begin{equation}\label{eqn:LOG_SUM_OF_SPEC_OVER_MARGIN}
        \begin{split}
            \mu_\text{\LOGSUMOFSPECOVERMARGIN}
            = & \log d \left(\frac{\prod_{i=1}^d \|\Wb_i\|_2^2}{\gamma^2}\right)^{1/d}\\
             = & \log d + \frac{1}{d} \left(\mu_\text{\LOGPRODOFSPEC} - 2\log \gamma \right).
        \end{split}
    \end{equation}
    \item (\LOGPRODOFFROOVERMARGIN).
    \begin{equation}\label{eqn:LOG_PROD_OF_FRO_OVER_MARGIN}
        \begin{split}
            \mu_\text{\LOGPRODOFFROOVERMARGIN}
            =  \log \frac{\prod_{i=1}^d \|\Wb_i\|_F^2}{\gamma^2} 
            =  \mu_\text{\LOGPRODOFFRO} - 2 \log \gamma.
        \end{split}
    \end{equation}
    Note that \LOGPRODOFFRO is not used in this paper due to overlap with \LOGNORM.
    \item (\LOGSUMOFFROOVERMARGIN).
    \begin{equation}\label{eqn:LOG_SUM_OF_FRO_OVER_MARGIN}
        \begin{split}
               \mu_\text{\LOGSUMOFFROOVERMARGIN}
            = \log d \left(\frac{\prod_{i=1}^d \|\Wb_i\|_F^2}{\gamma^2}\right)^{1/d} 
             = \log d + \frac{1}{d} \left(\mu_\text{\LOGPRODOFFRO} - 2\log \gamma \right).
        \end{split}
    \end{equation}
    \item (\PATHNORMOVERMARGIN).
    \begin{equation}\label{eqn:PATH_NORM_OVER_MARGIN}
        \begin{split}
            \mu_\text{\PATHNORMOVERMARGIN} = \frac{\mu_\text{\PATHNORM}}{\gamma^2}.
        \end{split}
    \end{equation}
\end{itemize}

\subsection{Metrics derived from PAC-Bayesian bounds}

Several well-known generalization bounds are derived using the PAC-Bayesian framework, which bounds the generalization gap using the KL-divergence between a predefined prior distribution (usually chosen as Gaussian) and the posterior distribution of the trained models.
A key component of the PAC-Bayesian bounds used in most existing implementations is the procedure of searching for the largest magnitude of Gaussian perturbation, denoted as $\sigma$, such that the perturbed weights of the neural network achieve a bounded increase in the training loss.
More specifically, $\sigma$ is defined such that
\begin{equation}
    \mathbb{E}_{\Ub\sim \mathcal{N}(\mathbf{0},\sigma^2\Ib)}[\text{TrainLoss}(f_{\Wb+\Ub})]\le \text{TrainLoss}(f_{\Wb}) +\delta,
\end{equation}
where $\delta$ is a predetermined threshold, and is chosen as $\delta=\frac{1}{2}$ in our machine translation experiments.
Similarly, one can define a ``magnitude-aware'' perturbation $\sigma'$ satisfying
\begin{equation}
    \mathbb{E}_{\Ub}[\text{TrainLoss}(f_{\Wb+\Ub})]\le \text{TrainLoss}(f_{\Wb}) +\delta,
\end{equation}
where each weight entry $u_i$ in $\Ub$ is distributed as $\mathcal{N}(0,\sigma'^2 |w_i|^2+\epsilon^2)$, and $\epsilon$ is chosen as 1e-3 \citep{dziugaite2020search}.
Given the perturbation magnitude $\sigma$, the magnitude-aware perturbation $\sigma'$ and the number of samples $m$, one can define the following generalization metrics.
\begin{itemize}
    \item (\PACBAYESINIT).
    \begin{equation}\label{eqn:PACBAYES_INIT}
        \mu_\text{\PACBAYESINIT} = \frac{\mu_\text{\LTWODIST}^2}{4\sigma^2}+\log\frac{m}{\sigma}+10.
    \end{equation}
    \item (\PACBAYESORIG).
    \begin{equation}\label{eqn:PACBAYES_ORIG}
        \mu_\text{\PACBAYESORIG} = \frac{\mu_\text{\LTWO}^2}{4\sigma^2}+\log\frac{m}{\sigma}+10.
    \end{equation}
    \item (\PACBAYESFLATNESS).
    \begin{equation}\label{eqn:PACBAYES_FLATNESS}
        \mu_\text{\PACBAYESFLATNESS} = \frac{1}{\sigma^2}.
    \end{equation}
    \item (\PACBAYESMAGINIT).
    \begin{equation}\label{eqn:PACBAYES_MAG_INIT}
        \begin{split}
            &\mu_\text{\PACBAYESMAGINIT} 
            = \frac{1}{4}\sum_{i=1}^\omega\log \left(\frac{\epsilon^2+(\sigma'^2+1)\mu_\text{\LTWODIST}^2/\omega}{\epsilon^2+\sigma'^2|w_i-w_i^\text{init}|^2}\right) 
            +\log\frac{m}{\sigma}+10.
        \end{split}
    \end{equation}
    \item (\PACBAYESMAGORIG).
    \begin{equation}\label{eqn:PACBAYES_MAG_ORIG}
        \begin{split}
            &\mu_\text{\PACBAYESMAGORIG} 
            = \frac{1}{4}\sum_{i=1}^\omega\log \left(\frac{\epsilon^2+(\sigma'^2+1)\mu_\text{\LTWO}^2/\omega}{\epsilon^2+\sigma'^2|w_i-w_i^\text{init}|^2}\right) 
            +\log\frac{m}{\sigma}+10.
        \end{split}
    \end{equation}
    \item (\PACBAYESMAGFLATNESS).
    \begin{equation}\label{eqn:PACBAYES_MAG_FLATNESS}
        \mu_\text{\PACBAYESMAGFLATNESS} = \frac{1}{\sigma'^2}.
    \end{equation}
    
\end{itemize}

%% file: sections/Additional_setup.tex
\section{Additional details on the experiment setup}
\label{app:experiment_setup}

For training Transformers, we follow exactly the setup in \citet{vaswani2017attention}, and we develop our implementation based on an online repository\footnote{\url{https://github.com/gordicaleksa/pytorch-original-transformer}} which reproduces the results from \citet{vaswani2017attention} with more easily configurable Transformer architectures.
The architecture puts the LayerNorm before the residual connection, which has been shown to provide more stabilized training \citep{xiong2020layer,liu2020understanding_admin}.
As we have mentioned earlier, we vary the hyperparameters of training to evaluate the correlations between the generalization metrics and model quality.
In the ``standard setting'', we train with Transformer-base, which has six layers, eight attention heads and embedding dimension 512.
Then, we vary the number of Transformer layers from 4 to 8, and we vary the embedding dimension from 256 to 1024.
When varying the embedding dimension, we let the number of attention heads vary proportionally.

We train with dropout 0.1 and 10\% label smoothing.
Note that in one experiment shown in Figure~\ref{fig:generalization}, we remove dropout to observe the effect of overfitting.
For all of our experiments, we train with the inverse square-root learning rate.
Given the embedding dimension $d_e$, step number $t$, number of warm-up steps $t_w$, the formula for the inverse square-root learning rate schedule \citep{vaswani2017attention} is the following.
\begin{equation}\label{eq:inverse_sqrt}
    \text{Learning Rate} = d_e^{-0.5}\cdot\min(t^{-0.5}, t\cdot t_w^{-1.5}).
\end{equation}
For results trained with a particular learning rate $lr$, such as the results shown in Figure~\ref{fig:group_by_lr}, $lr$ is the constant factor multiplied with the standard learning rate schedule \eqref{eq:inverse_sqrt}.
For each experiment, we train the model for 20 epochs.
When calculating the ESDs of the weight matrices, we treat the query, key and value matrices as separate weight~matrices.

%% file: sections/Margin_issue.tex
\section{Additional analysis on scale metrics}
\label{app:scale_metrics}

\begin{figure}
    \centering
    \includegraphics[width=0.45\textwidth]{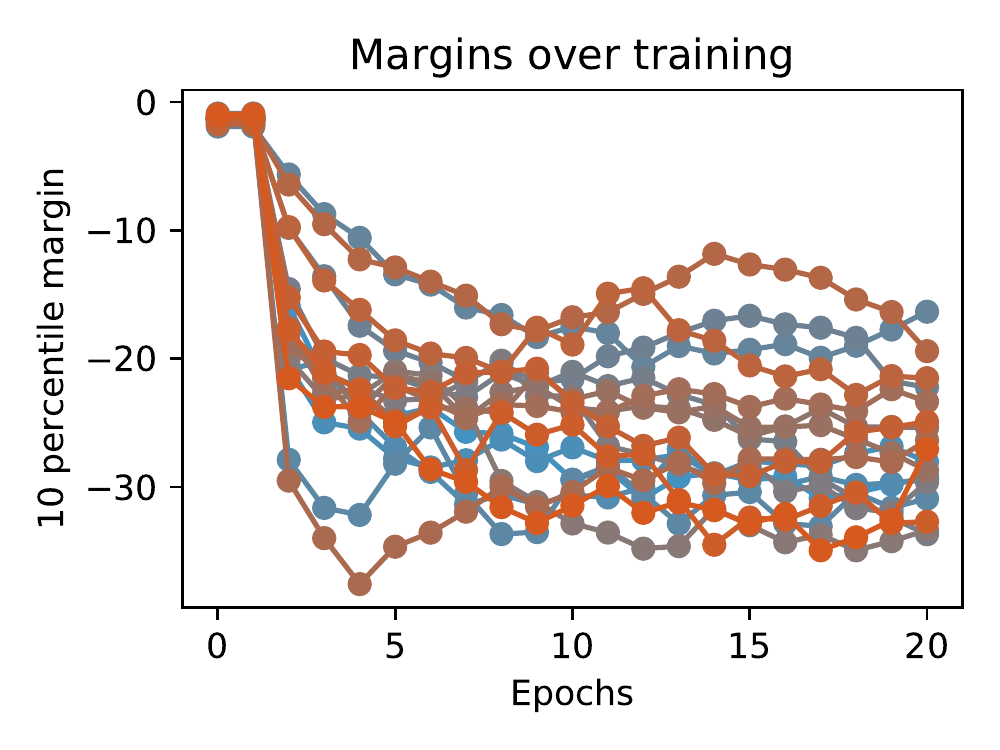}
    \caption{The margins remain negative in the experiments on machine translation due to the large alphabet~size.}
    \label{fig:margin}
\end{figure}

In this section, we discuss an issue of computing margin-based generalization metrics. Generically, these bounds are of the form
$$
L(f) \leq \hat{L}_\gamma(f) + C/\gamma 
$$
where $L(f)$ is the population error, $\hat{L}_\gamma$ is the training margin loss at margin $\gamma$, typically 
$$
\sum_{(x,y)\in S} \mathbf{1}\{\max_j f(x)_j \leq \gamma + f(x)_y\},
$$ 
and $C$ is some complexity term. First, note that this construction requires the margin $\gamma$ to be positive. Moreover, the training margin loss is an increasing function of $\gamma$, while the complexity term $C/\gamma$ is decreasing in $\gamma$, thus the conventional way of using the margin bound is to optimize over the margin to balance two terms in the margin bound \citep{bartlett2017spectrally}, rather than pre-specifying the value of the margin dependent on the data.
However, we choose to follow the related papers \citet{dziugaite2020search,jiang2019fantastic}, and we use the 10th percentile margin as a robust estimate of the minimum margin in the dataset.
We use this margin in all of the margin-normalized generalization metrics.
However, in all of the experiments on machine translation, the 10th percentile margin remains negative throughout the whole training, violating the requirement that the bound is evaluated at a positive value of margin. 
See Figure~\ref{fig:margin}.
This problem results from the large Alphabet for machine translation, which makes it difficult to fully interpolate the data, and hence makes the margin-normalized generalization metrics in \citet{dziugaite2020search,jiang2019fantastic} hard to be applicable to the present setting.

%% file: sections/Experiments_ODR.tex
\section{Fitting regression lines using orthogonal distance regression}\label{sec:odr}

\begin{figure*}
    \centering
    \includegraphics[width=0.32\textwidth]{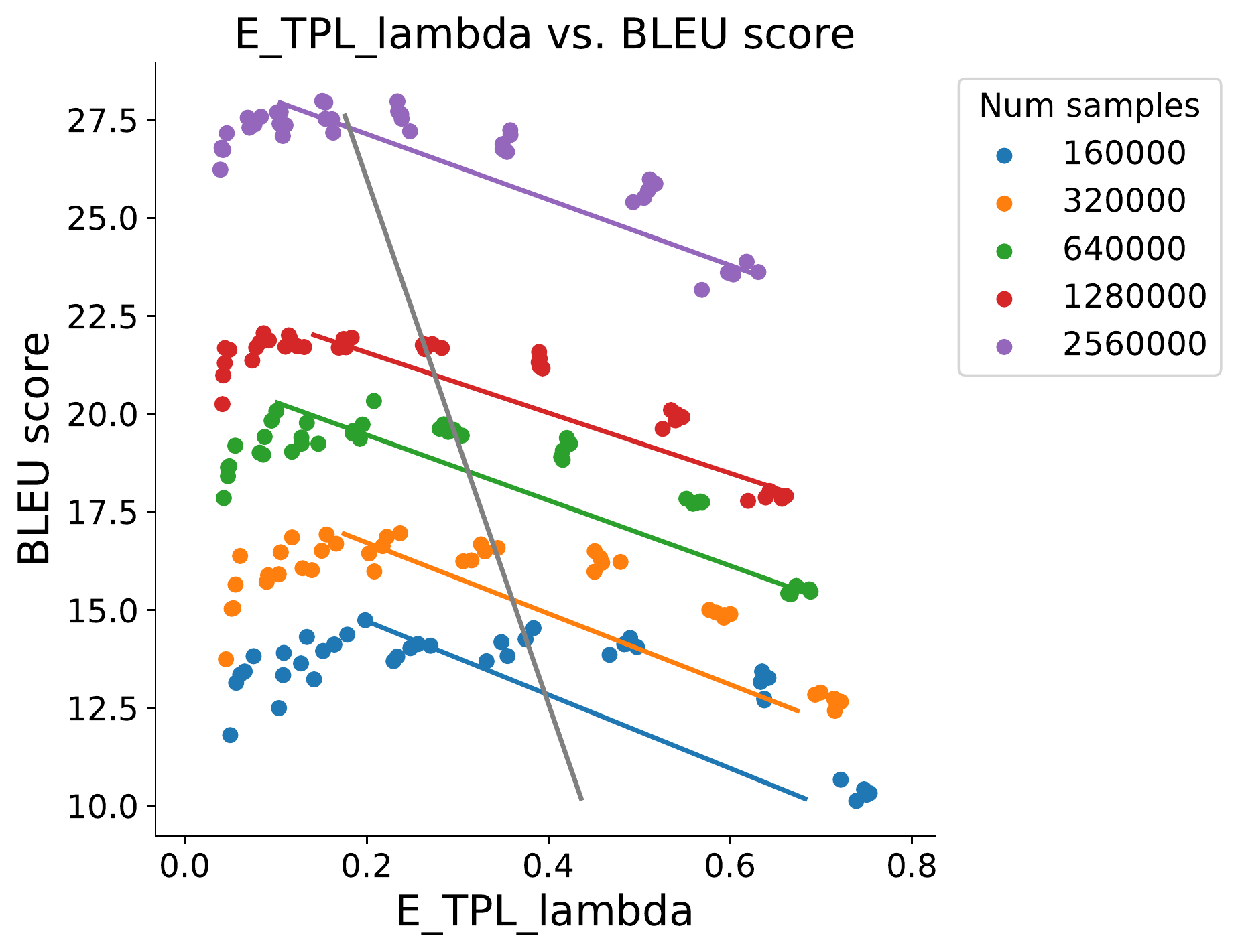}
    \includegraphics[width=0.32\textwidth]{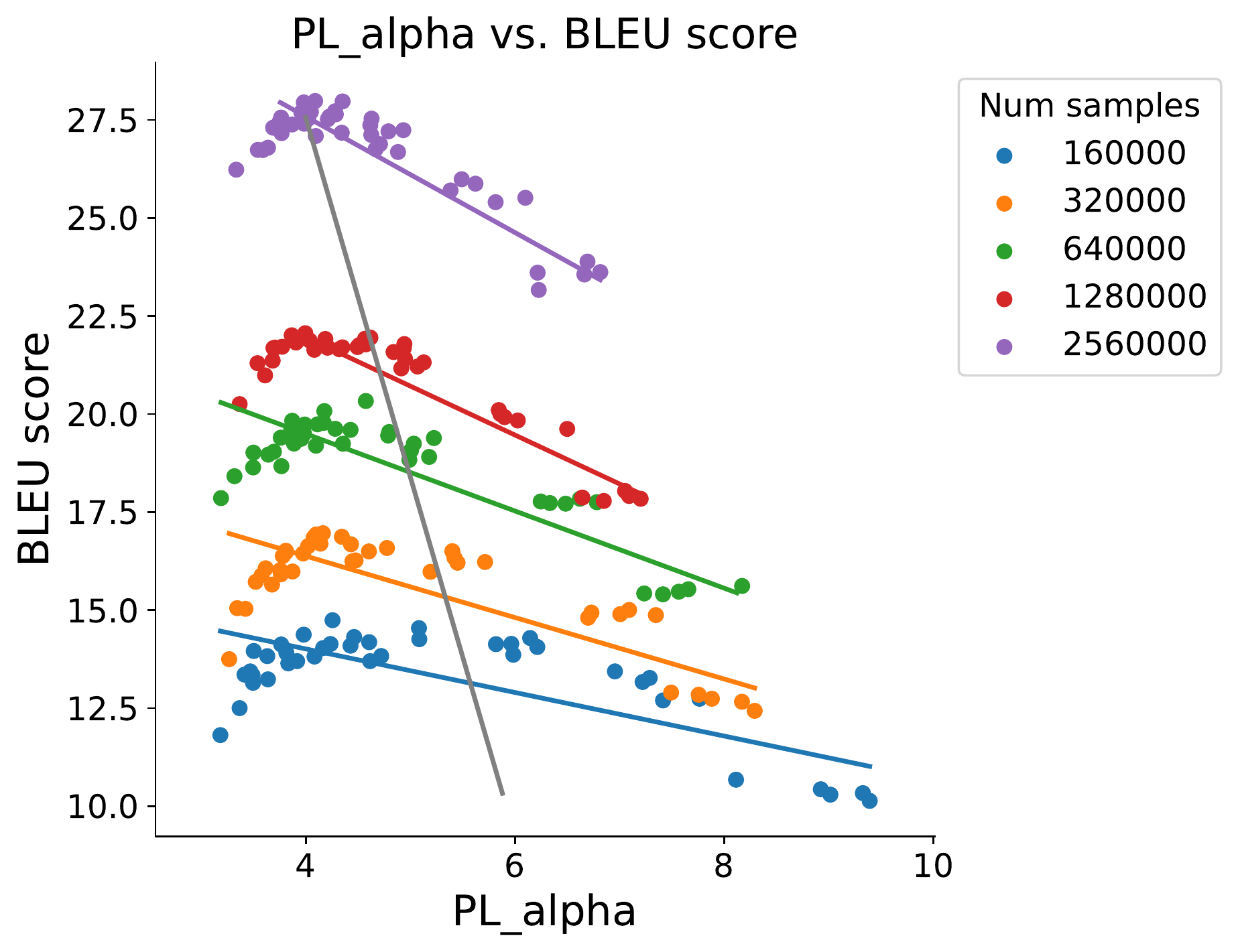}
    \includegraphics[width=0.32\textwidth]{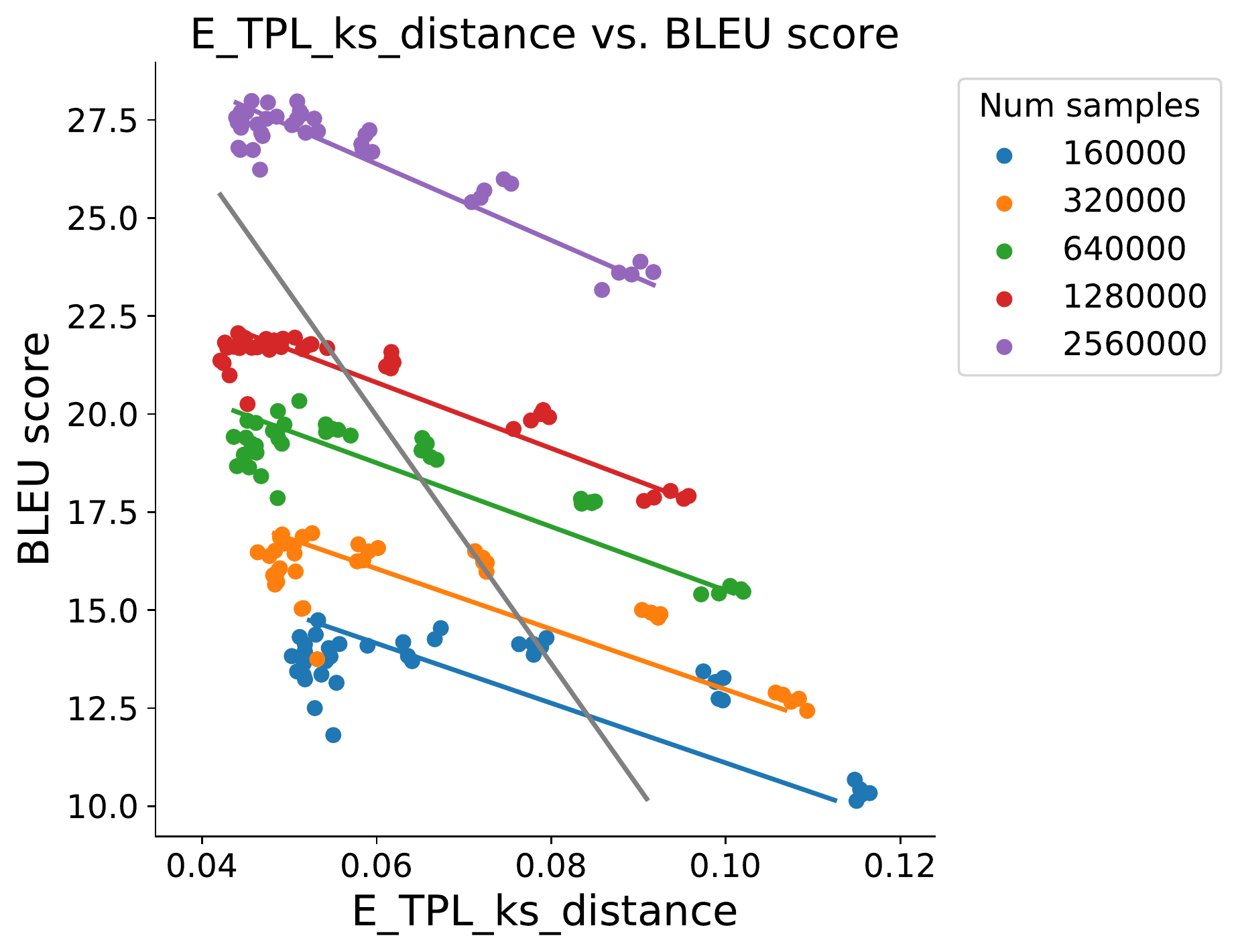}
    \includegraphics[width=0.32\textwidth]{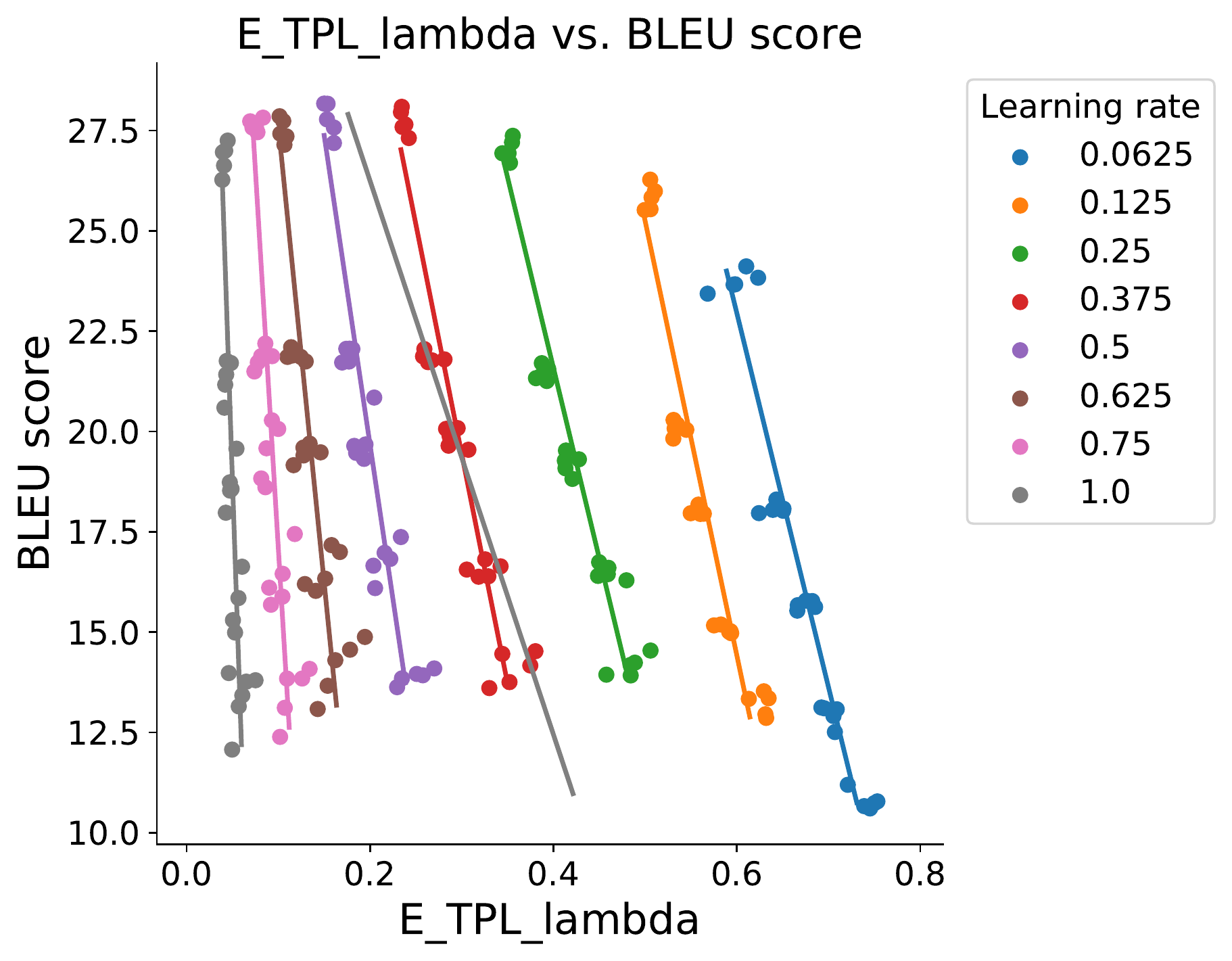}
    \includegraphics[width=0.32\textwidth]{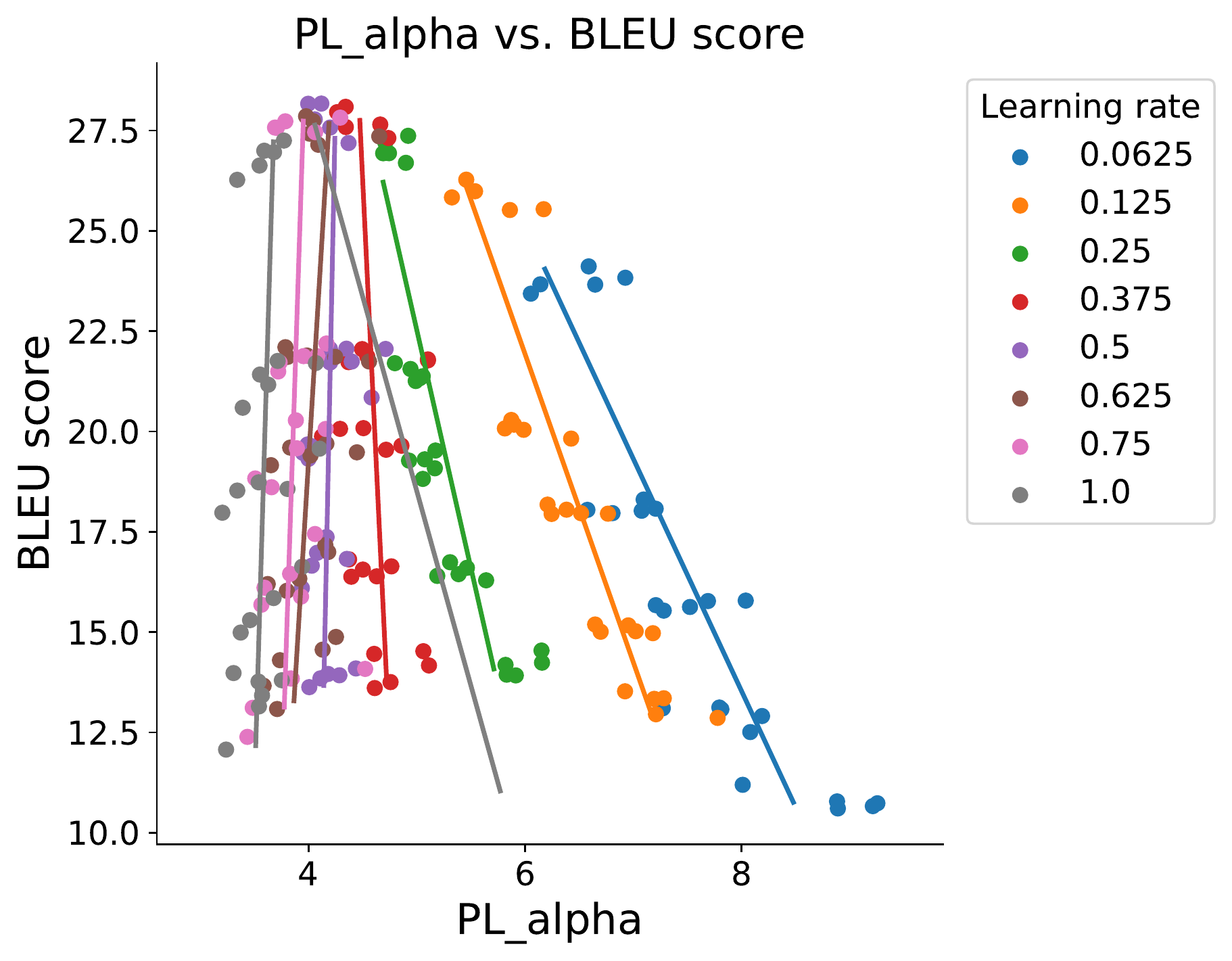}
    \includegraphics[width=0.32\textwidth]{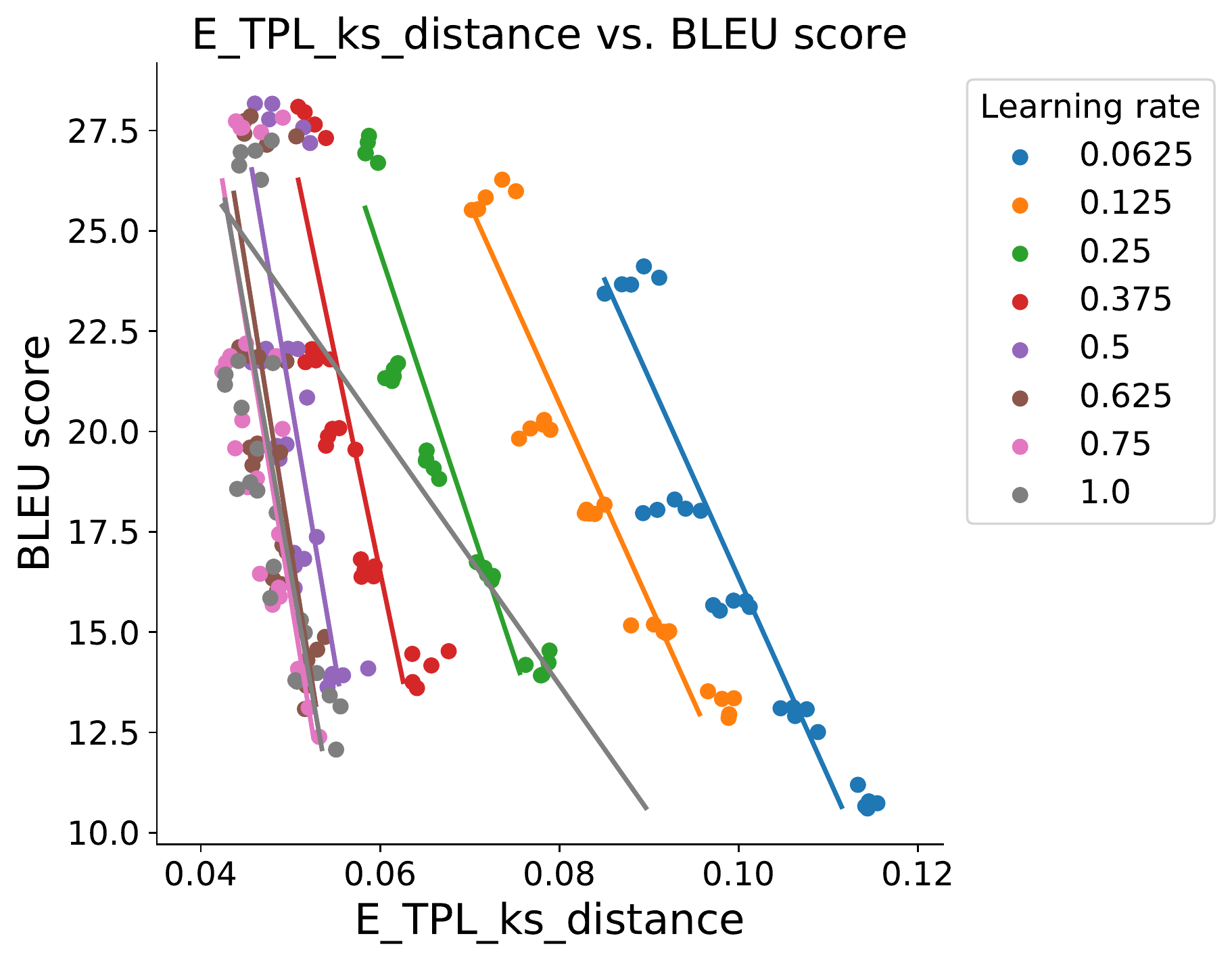}
    \caption{BLEU-score versus shape metrics. {\bf(First row).} Trained models grouped by the learning rate. {\bf(Second row).} Trained models grouped by the number of samples. The regression lines are fitted using orthogonal distance regression.}
    \label{fig:Shape_metrics_odr}
\end{figure*}

\begin{figure*}
    \centering
    \includegraphics[width=0.32\textwidth]{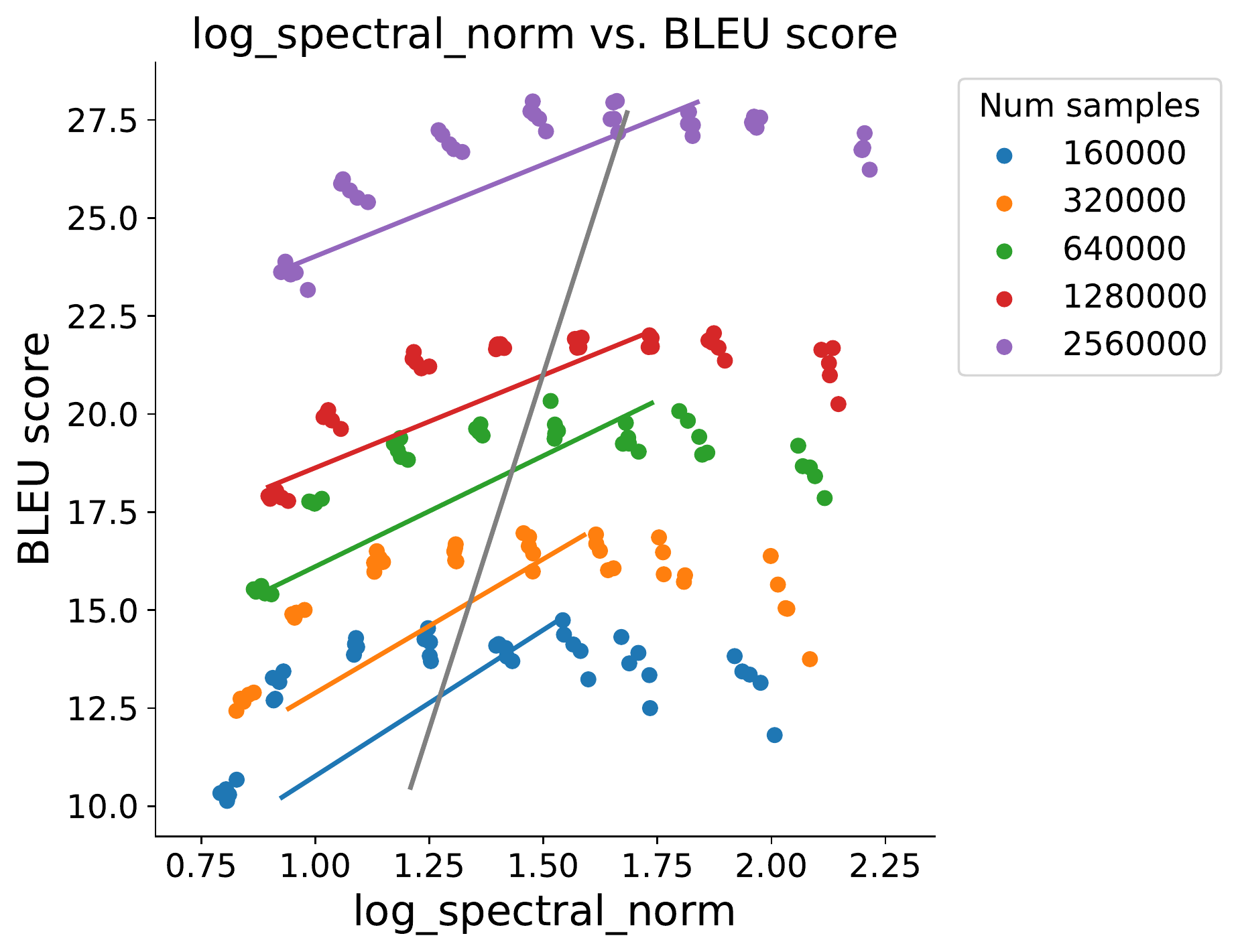}
    \includegraphics[width=0.32\textwidth]{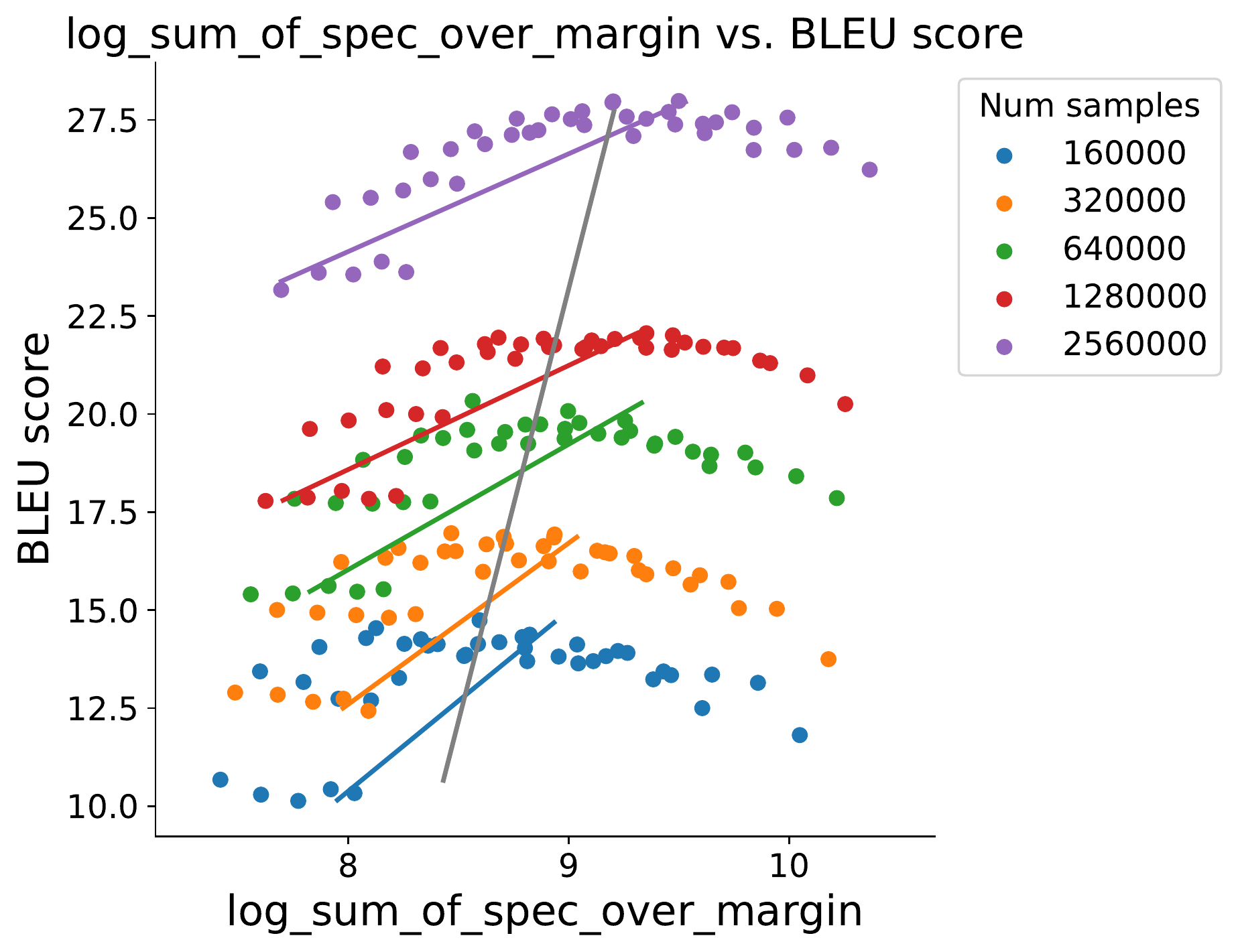}
    \includegraphics[width=0.32\textwidth]{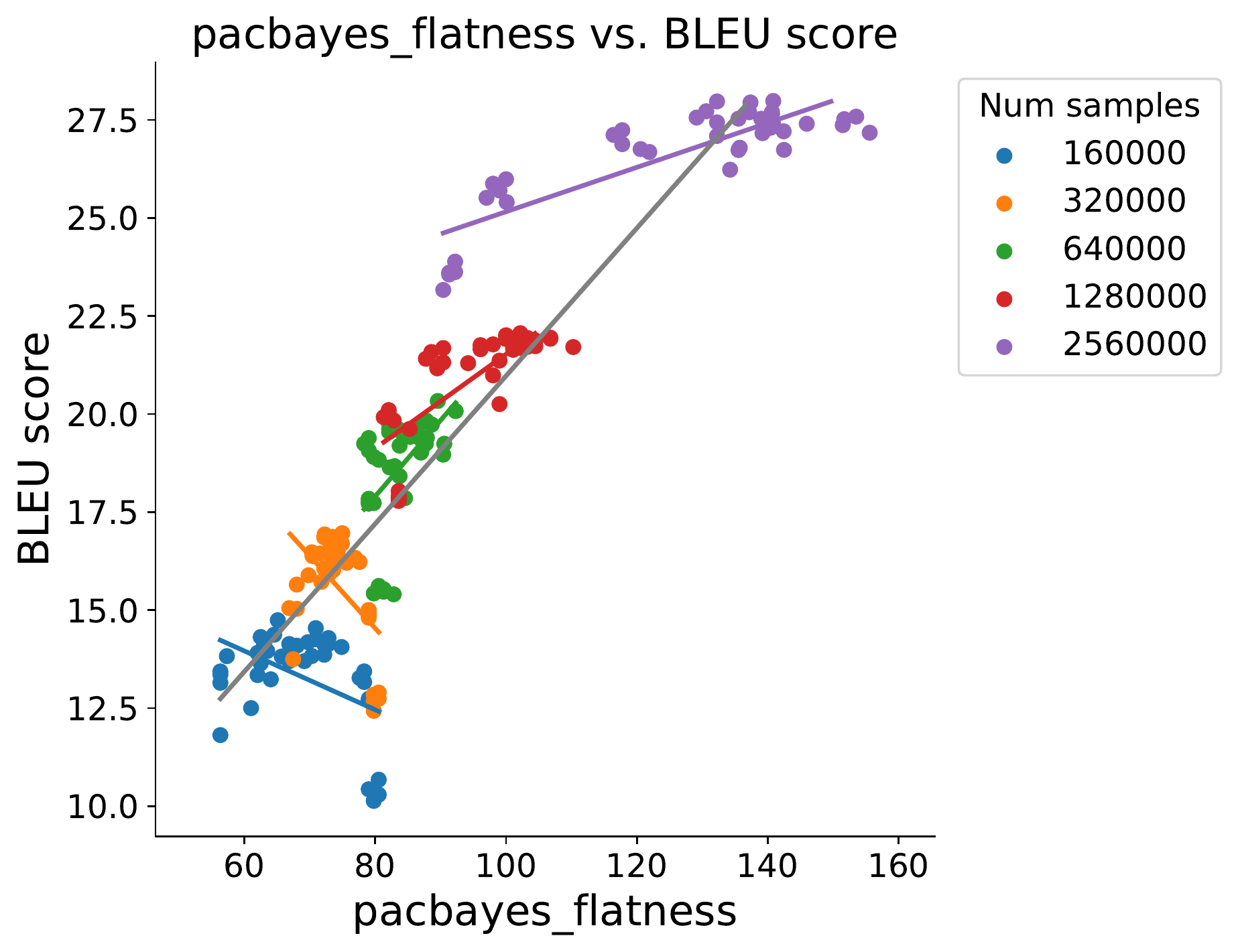}
    \includegraphics[width=0.32\textwidth]{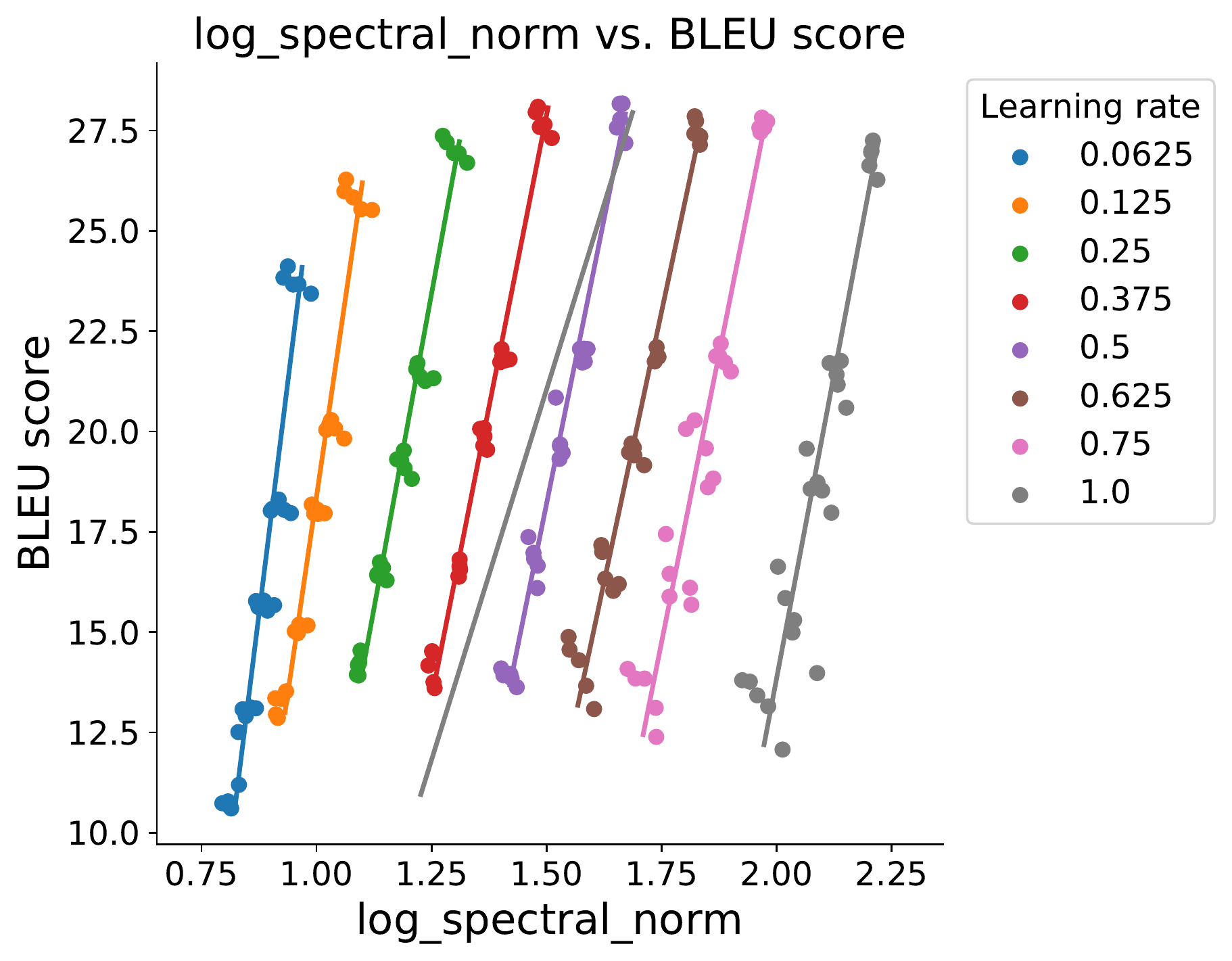}
    \includegraphics[width=0.32\textwidth]{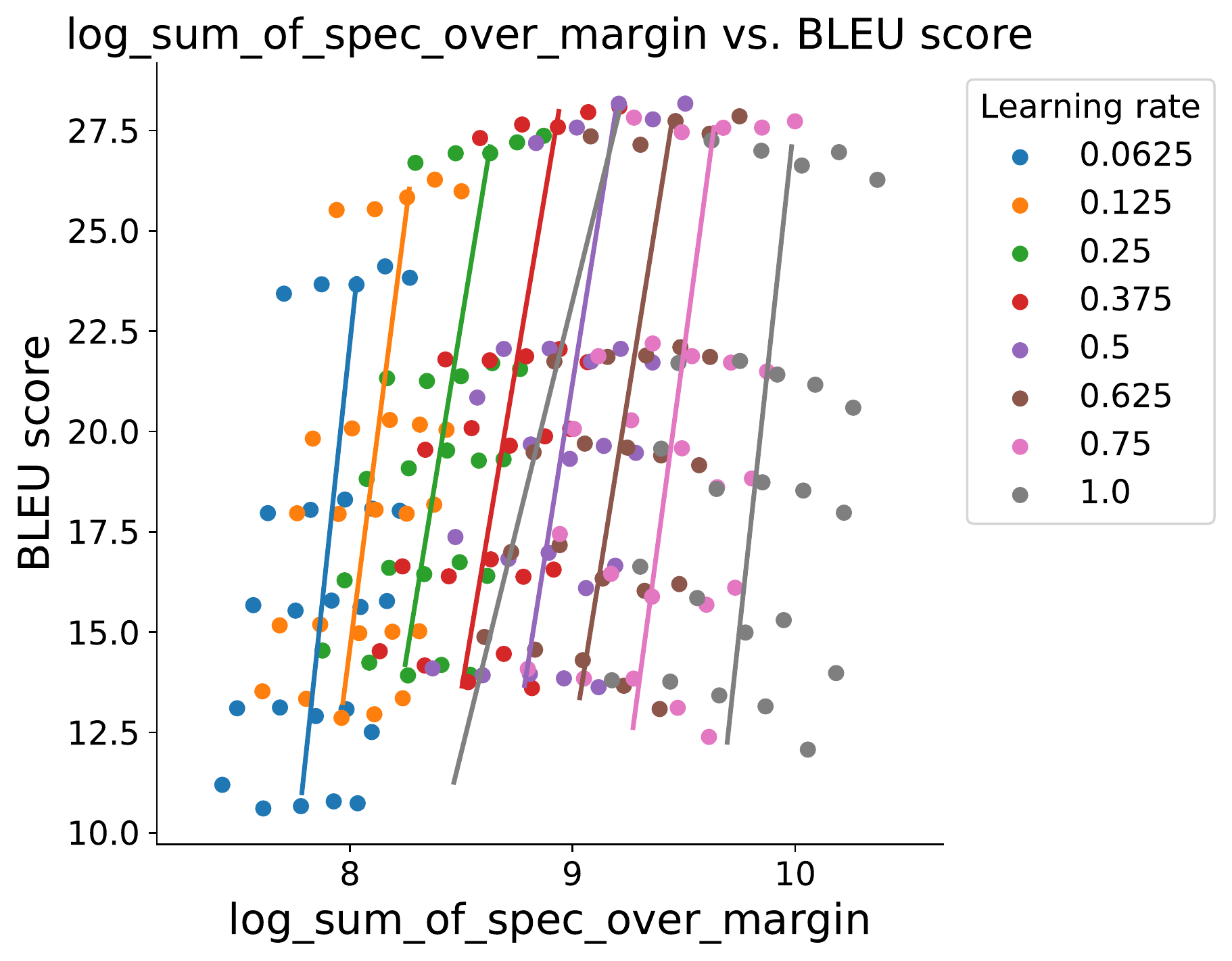}
    \includegraphics[width=0.32\textwidth]{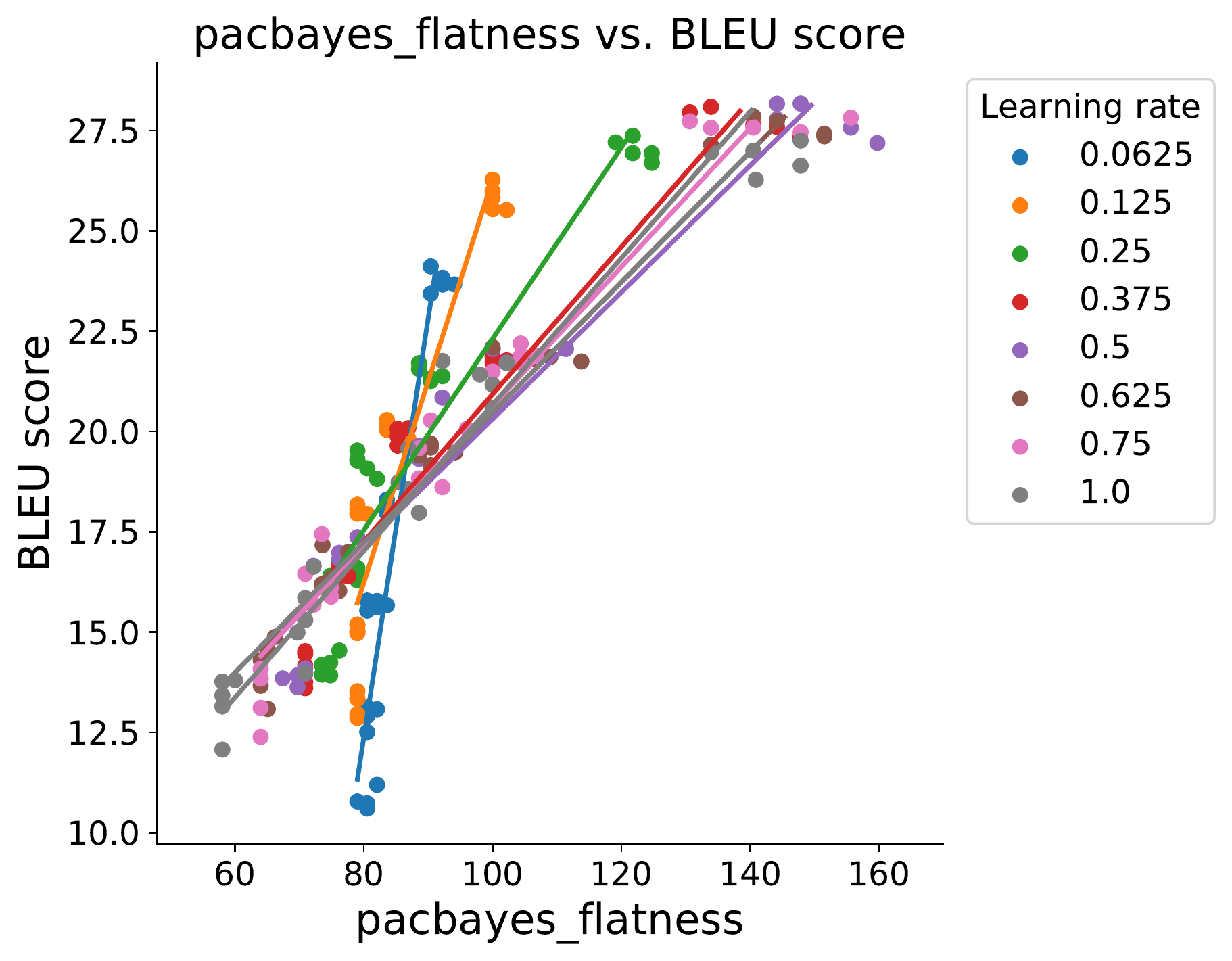}
    \caption{BLEU-score versus scale metrics. {\bf(First row).} Trained models grouped by the learning rate. {\bf(Second row).} Trained models grouped by the number of samples. The regression lines are fitted using orthogonal distance regression.}
    \label{fig:Scale_metrics_odr}
\end{figure*}

In this section, we fit the regression plots from Figure~\ref{fig:Shape_metrics} to~\ref{fig:Scale_metrics} using the orthogonal distance regression \citep{boggs1990orthogonal}.
See Figure~\ref{fig:Shape_metrics_odr} to Figure~\ref{fig:Scale_metrics_odr}.

%% file: sections/Experiments_Additional.tex
\section{Corroborating results}\label{app:additional_results}

In this subsection, we consider corroborating results, extending the setup of the main paper to more datasets and different evaluation methods.

\subsection{Additional results on natural language processing tasks}\label{sec:additional_experiments}

We consider three other NLP tasks:
\begin{itemize}
    \item Roberta \citep{liu2019roberta} trained on the masked language modeling task using the Wikitext-103 dataset, and then finetuned on the MNLI dataset \citep{williams2018broad}.
    \item Six-layer base Transformers trained on the language modeling task using the Wikitext-103 dataset \citep{merity2016pointer};
    \item Six-layer base Transformers trained on the next-word prediction task using the Reddit dataset,
    following the implementation in \citet{bagdasaryan2020backdoor};
\end{itemize}

\begin{figure}
    \centering
    \begin{subfigure}{0.245\textwidth}
        \includegraphics[width=\textwidth]{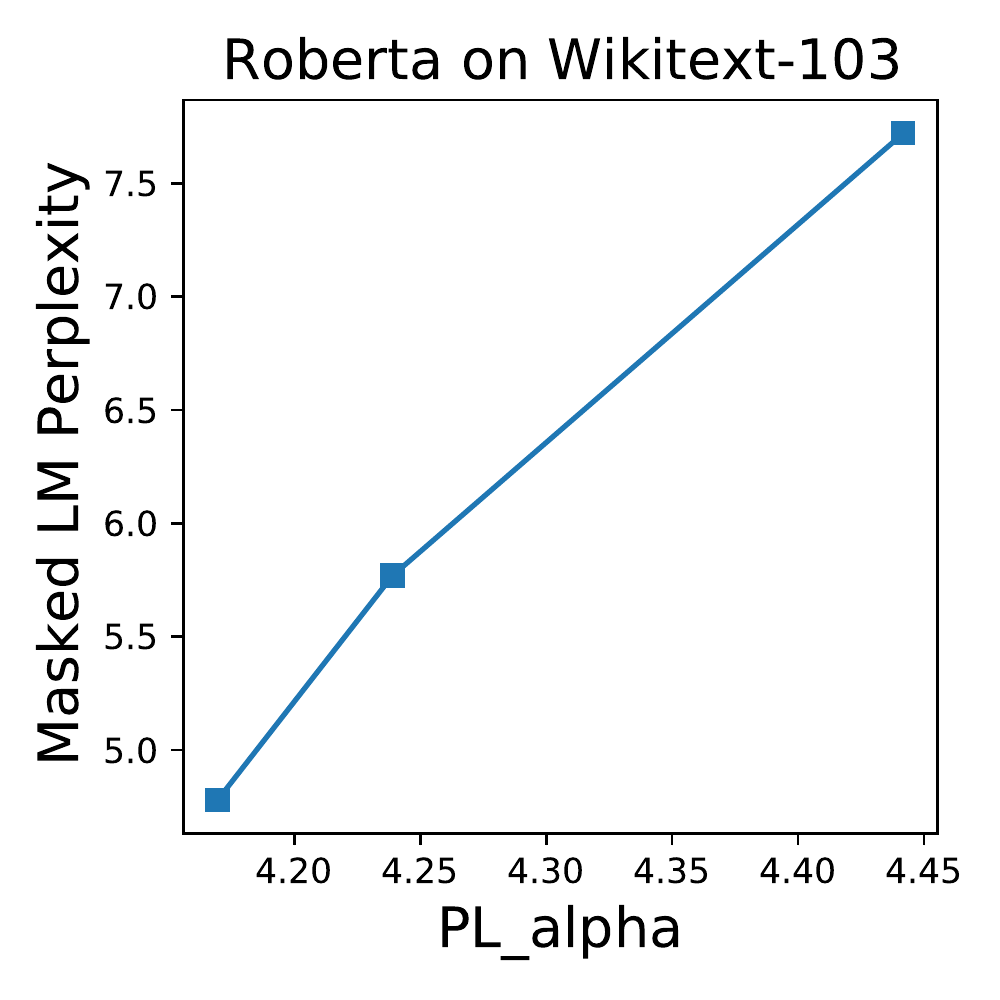}
        \caption{\scriptsize Lower perplexity is better.}
    \end{subfigure}\hfill
    \begin{subfigure}{0.245\textwidth}
        \includegraphics[width=\textwidth]{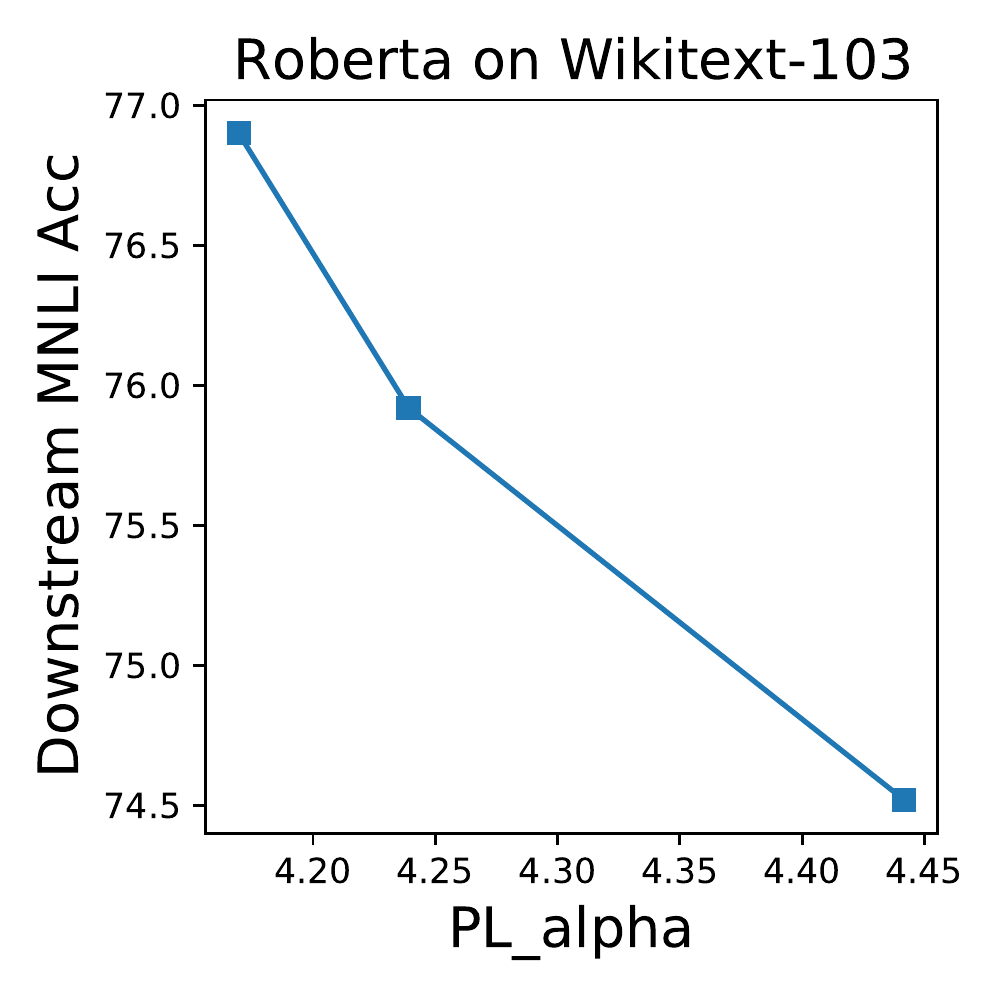}
        \caption{\scriptsize Higher accuracy is better.}
    \end{subfigure}\hfill
    \begin{subfigure}{0.245\textwidth}
        \includegraphics[width=\textwidth]{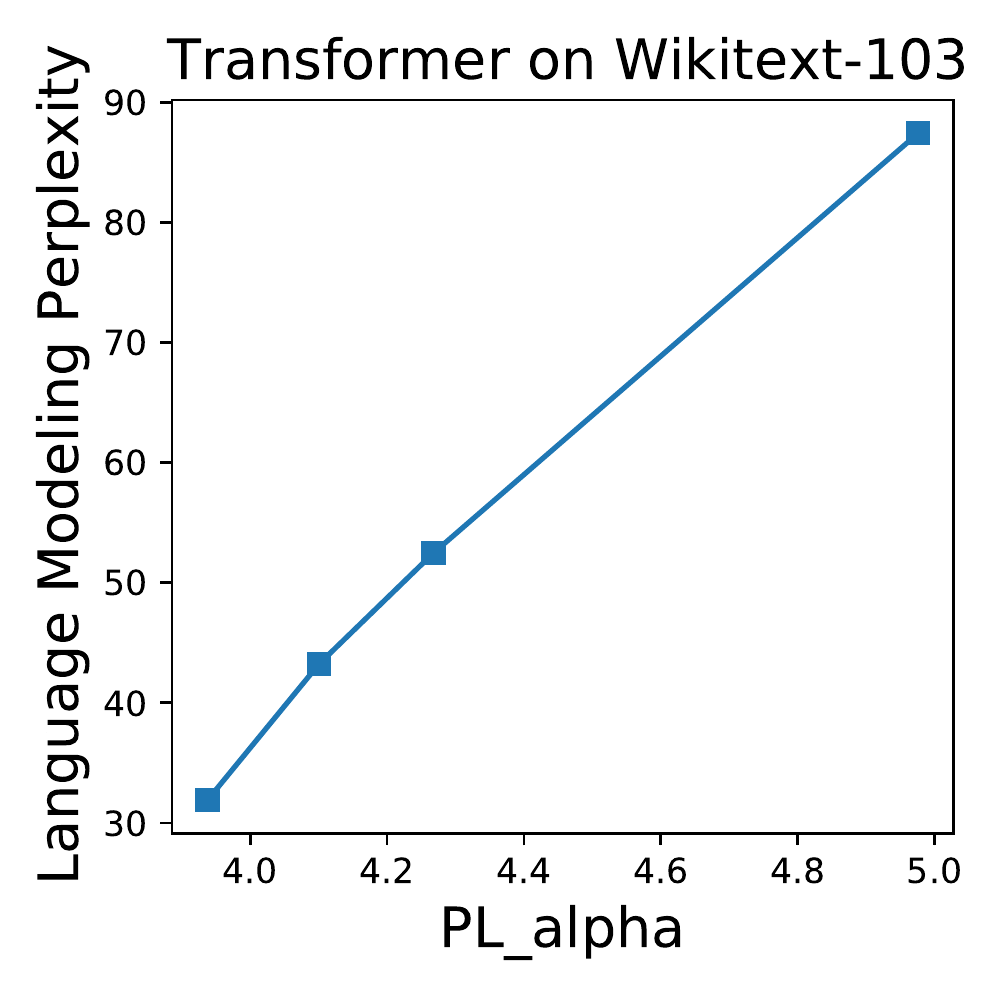}
        \caption{\scriptsize Lower perplexity is better.}
    \end{subfigure}\hfill
    \begin{subfigure}{0.245\textwidth}
        \includegraphics[width=\textwidth]{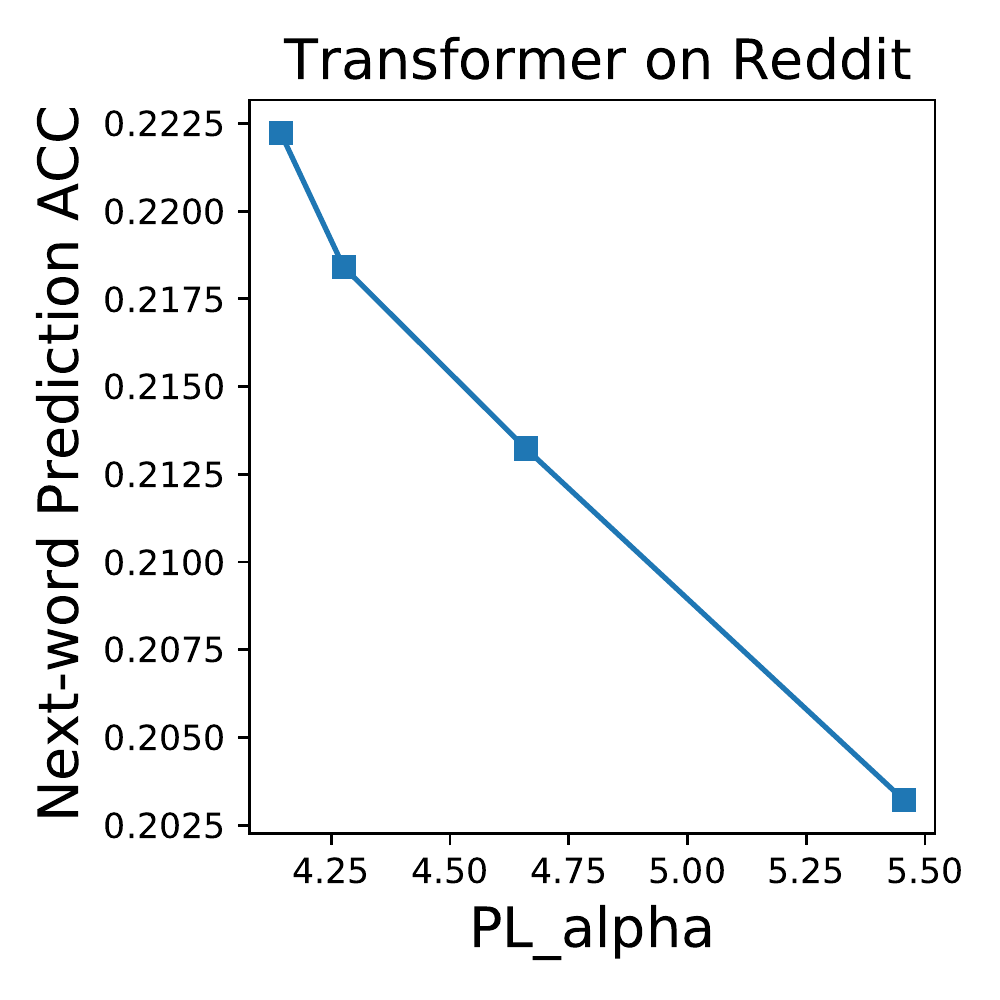}
        \caption{\scriptsize Higher accuracy is better.}
    \end{subfigure}
    \caption{Multiple language processing tasks evaluated using the \ALPHA metric on models trained with different data sizes.
    The \ALPHA metric correctly predicts the trend in all tasks.
    }
    \label{fig:additional_results}
\end{figure}

For each task, we train models on different data sizes. Then, we measure the \ALPHA metric and report the correlation with the ground-truth quality metric. See Figure~\ref{fig:additional_results}.
In these experiments, the \ALPHA metric predicts the correct trend, i.e., a lower value of \ALPHA corresponds to a better model.

\subsection{Evaluating rank correlations using Kendall's tau metric}\label{sec:kendalltau}

\begin{figure}
    \centering
    \begin{subfigure}{0.48\textwidth}\centering
        {\small\textbf{Correlations with model quality}} \\
         \includegraphics[width=0.95\textwidth]{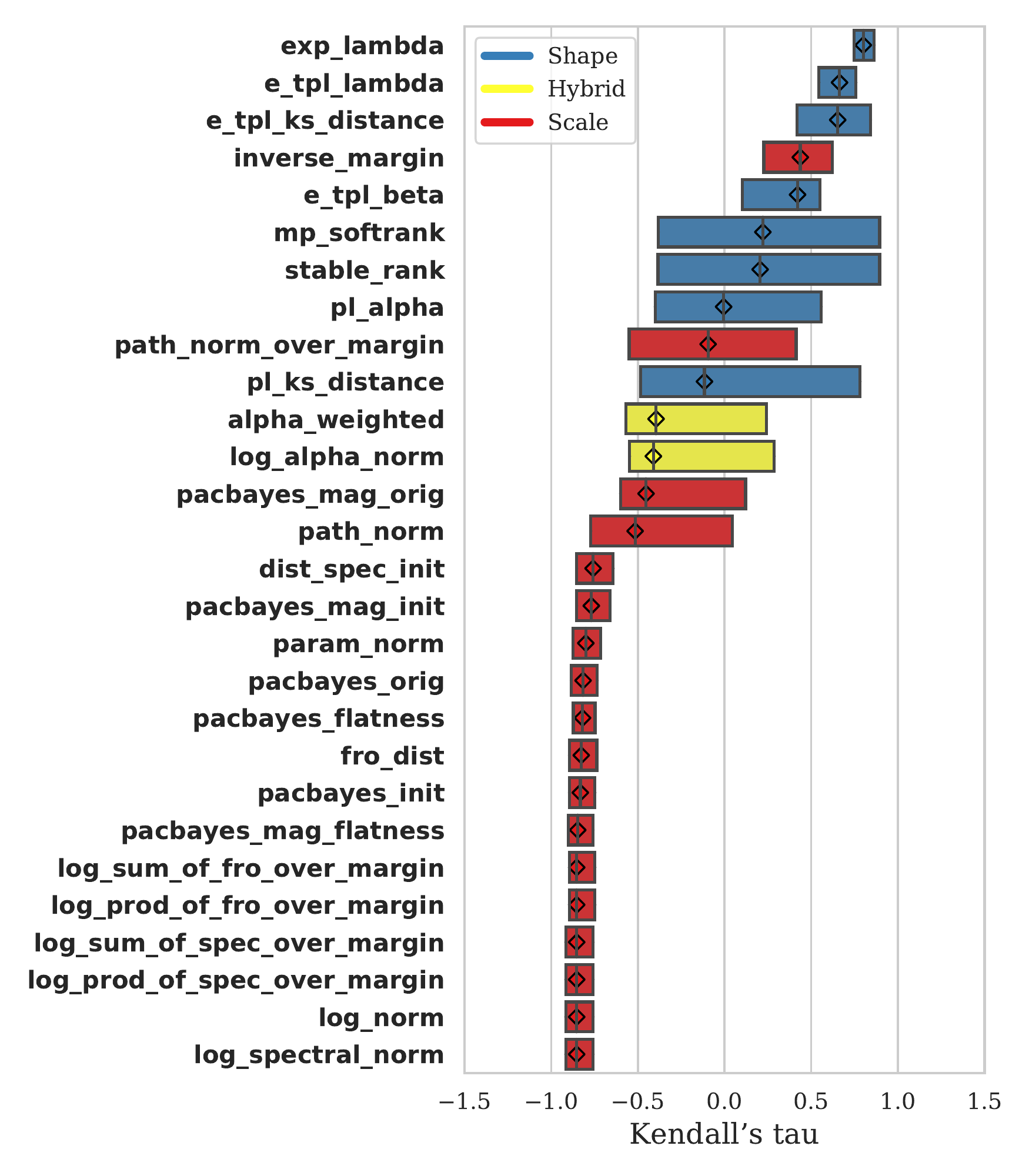}
        \caption{\textbf{Correlations with model quality.} Kendall's tau between various generalization metrics and BLEU.}
        \label{fig:ave_worst_rank_correlations_kendalltau}
    \end{subfigure}\hfill
    \begin{subfigure}{0.48\textwidth}\centering
        {\small\textbf{Correlations with generalization gap}} \\
        \includegraphics[width=0.95\textwidth]{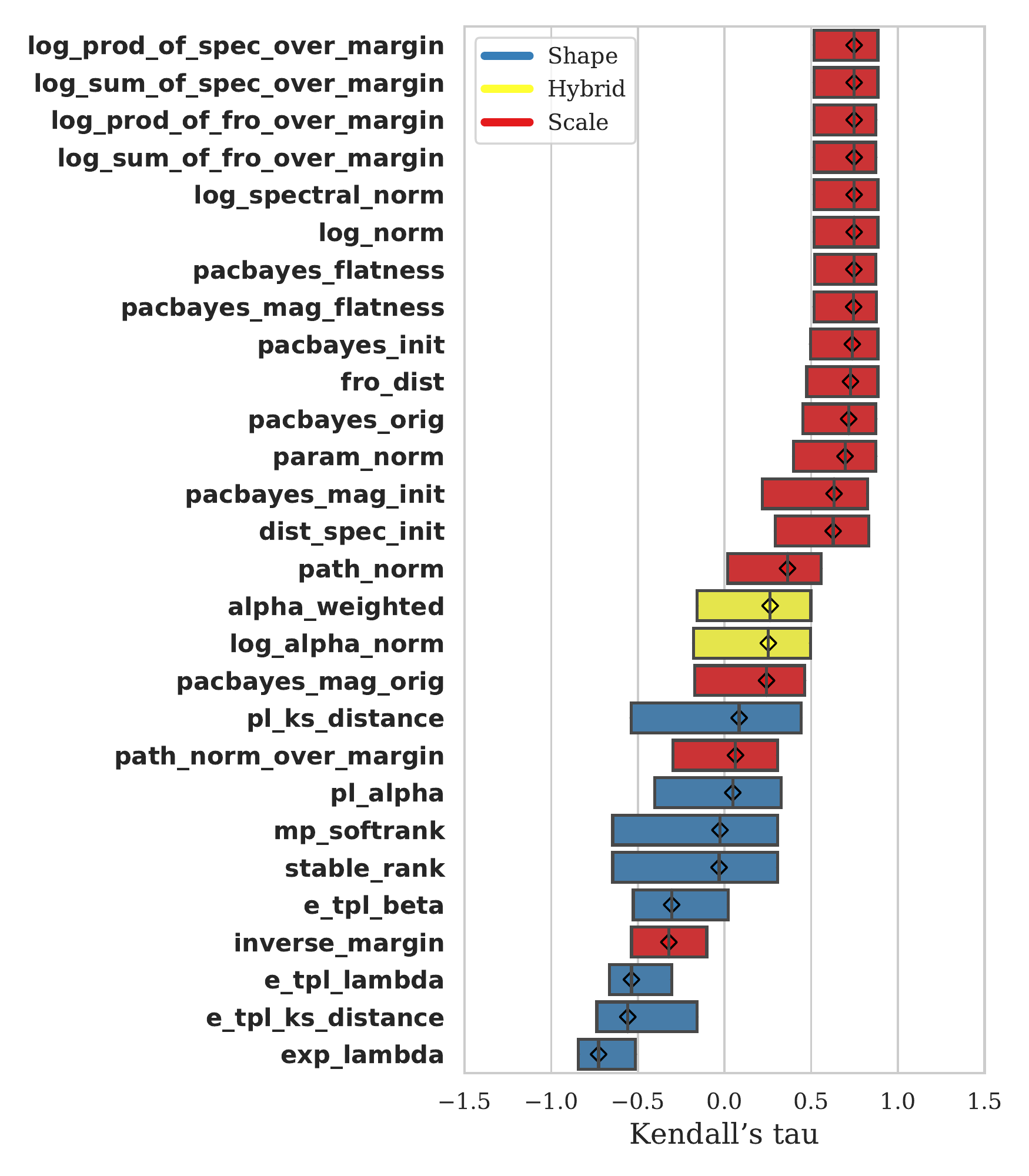}
        \caption{\textbf{Correlations with generalization gap.} Kendall's tau between various generalization metrics and the generalization gap. 
        }
        \label{fig:ave_worst_rank_correlations_gap_kendalltau}
    \end{subfigure}
    \caption{Evaluating Task Two using Kendall's tau. Results are similar to those in Figure~\ref{fig:test_acc_vs_generalization_gap}. \vspace{-.05cm}
    }
    \label{fig:test_acc_vs_generalization_gap_kendalltau}
\end{figure}

Next, we reimplement Task two using Kendall's tau to calculate the rank correlations. The results in Figure~\ref{fig:test_acc_vs_generalization_gap_kendalltau} are very similar to those in Figure~\ref{fig:test_acc_vs_generalization_gap}.

